\documentclass[lettersize,journal]{IEEEtran} 
\usepackage{cite}
\usepackage{amsmath,amssymb,amsfonts}
\usepackage{algorithmic}
\usepackage{graphicx}
\usepackage{textcomp}
\usepackage{algorithm2e}
\usepackage[dvipsnames]{xcolor}
\usepackage{array}
\usepackage{arydshln}

\usepackage{tikz}
\makeatletter
\let\NAT@parse\undefined
\makeatother
\usepackage{url}
\makeatletter
\g@addto@macro\UrlSpecials{\do\!{\newline}}
\usepackage[colorlinks=true, allcolors=blue, pdfborderstyle={/S/U/W 0}]{hyperref}
\newcommand{\blockcomment}[1]{}
\newcommand{\cmmntfew}[1]{}

\newcommand{\clinicalcomments}[1]{{\color{black}#1}}
\newcommand{\clinicalcommentr}[1]{{\color{black}#1}}

\newcommand{\revfresponse}[1]{{\color{black}#1}} 
\newcommand{\tip}[1]{{\color{black}#1}} 
\newcommand{\tobereviewd}[1]{{\color{black}#1}} 
\newcommand{\infoclear}[1]{{\color{black}#1}} 
\def\BibTeX{{\rm B\kern-.05em{\sc i\kern-.025em b}\kern-.08em
    T\kern-.1667em\lower.7ex\hbox{E}\kern-.125emX}}

\begin{document}
\title{sFRC for assessing hallucinations in medical image restoration}
\author{Prabhat Kc, Rongping Zeng, Nirmal Soni, and Aldo Badano
\thanks{\blockcomment{Manuscript received March 27, 2024. }This work was conducted at the U.S. Food and Drug Administration by employees of the Federal Government in the course of their official duties. (Corresponding Author: Prabhat Kc) }
\thanks{ P.~Kc, R.~Zeng, N.~Soni, and A.~Badano are with the Center for Devices and Radiological Health in FDA,
Silver Spring, MD, 20993, USA (e-mail: \href{mailto:prabhat.kc@fda.hhs.gov}{prabhat.kc@fda.hhs.gov}; \href{mailto:rongping.zeng@fda.hhs.gov}{rongping.zeng@fda.hhs.gov}; \href{mailto:nirmal.soni@fda.hhs.gov}{nirmal.soni@fda.hhs.gov}; \href{mailto:aldo.badano@fda.hhs.gov}{aldo.badano@fda.hhs.gov}).}
\thanks{This article has supplementary (\textbf{SI}) downloadable material.}
\thanks{\tip{\textbf{DISCLAIMER:} This article reflects the views of the authors and does not represent the views or policy of the U.S. Food and Drug Administration, the Department of Health and Human Services, or the U.S. Government.  The mention of commercial products, their sources, or their use in connection with material reported herein is not to be construed as either an actual or implied endorsement of such products by the Department of Health and Human Services.}}
}

\maketitle
\begin{abstract}
Deep learning (DL) methods are currently being explored to restore images from sparse-view-, limited-data-, and undersampled-based acquisitions in medical applications. Although outputs from DL may appear visually appealing based on likability/subjective criteria (such as less noise, smooth features), they may also suffer from hallucinations. This issue is further exacerbated by a lack of easy-to-use techniques and robust metrics for the identification of hallucinations in DL outputs. In this work, we propose performing Fourier Ring Correlation (\underline{FRC}) analysis over small patches and concomitantly (\underline{s})canning across DL outputs and their reference counterparts to detect hallucinations \tip{(termed as sFRC).} We describe the rationale behind sFRC and provide its mathematical \tip{formulation}. The parameters \tip{essential to} sFRC \infoclear{may} be set using predefined hallucinated features annotated by subject matter experts or using imaging theory-based hallucination maps. We use sFRC to detect hallucinations for \tip{three} undersampled medical imaging problems: CT super-resolution, \tip{CT sparse view}, and MRI subsampled restoration. \revfresponse{In the testing phase,} we demonstrate sFRC's effectiveness in detecting hallucinated features for the \revfresponse{CT problem and sFRC's agreement with imaging theory-based outputs on hallucinated feature maps for the MR problem}. Finally, we quantify the \tip{hallucination rates} of DL methods on in-distribution versus out-of-distribution data and under increasing subsampling rates to characterize the robustness of DL methods. \revfresponse{Beyond DL-based methods, sFRC’s effectiveness in detecting hallucinations for a conventional regularization-based restoration method and \tip{a state-of-the-art unrolled method} is also shown.}
\end{abstract}

\section{Introduction}
\begin{figure}[!b]
\centering
\includegraphics[width=0.95\linewidth]{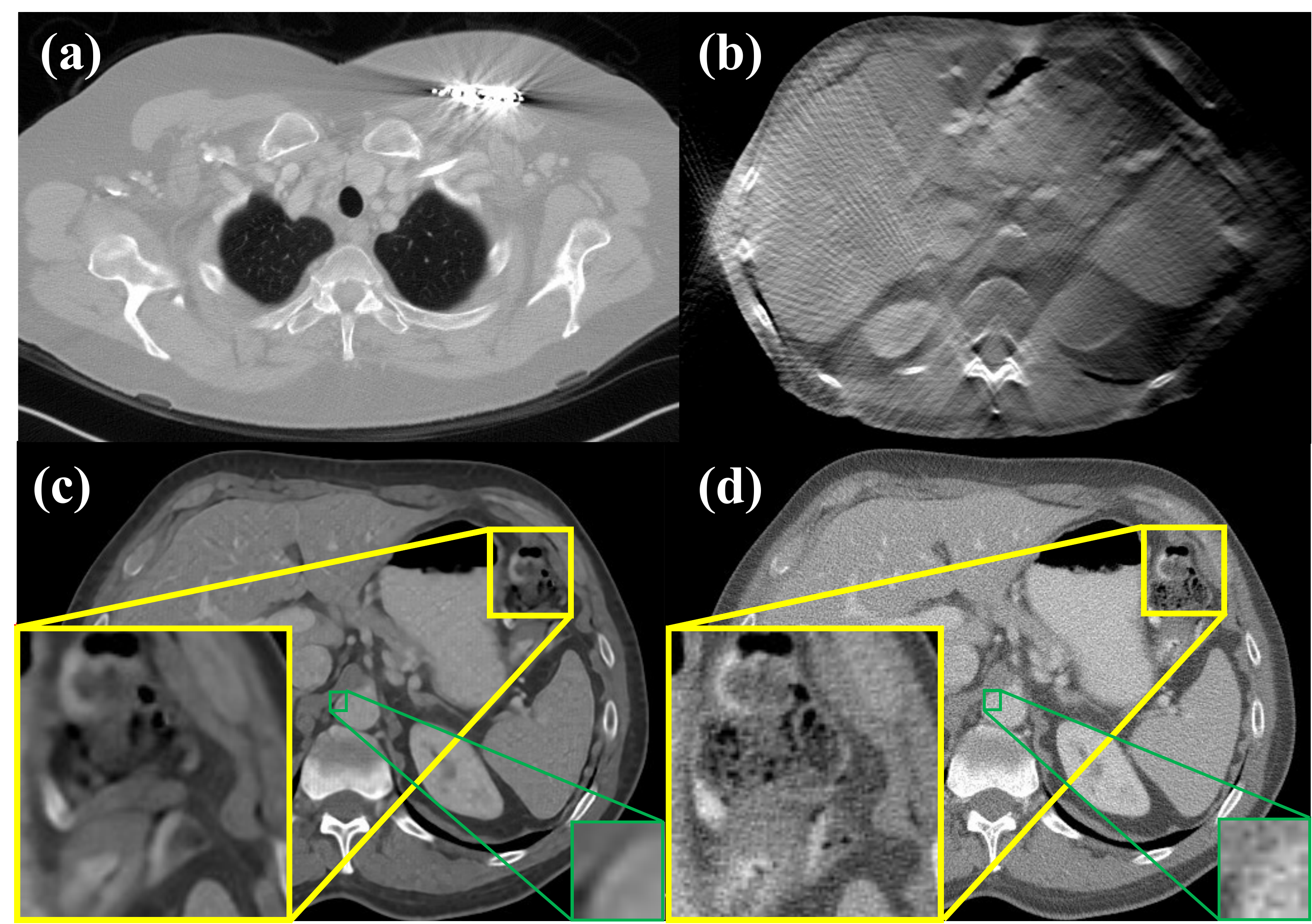}
\caption{A range of artifacts, including (a) patient implant-based artifact \cite{ldct_data_2016}), (b) missing wedge-based acquisition artifacts (described in SI section S2), and (c) hallucinations due to the application of an AI-assisted SRGAN model (described in \ref{sec:srgan_model}) to upsample the low-resolution counterpart of (d) by a factor of $4$. As opposed to artifacts in (a) \infoclear{and} (b), \clinicalcomments{hallucinations in (c) in the yellow box (with the two loops of bowel instead of one contiguous loop) and in the green box (an unwarranted plaque-like structure), only become concretely apparent after a thorough comparison against its normal-resolution reference in (d)}. Display windows for (a) is (W:$\,1402$ L:$\,-1358$), (b) is (W:$\,780$ L:$\,260$, and (c-d) is (W:$\,700$ L:$\,50$).}
\label{img:intro_plot_on_artifacts}
\end{figure}

\begin{table*}[!htb]
\tip{
\caption{Performance testing methodologies, their primary purposes, and their ability to assess hallucinations in image restoration problems}
\label{tab:iq_table_summary}
\centering
\renewcommand{\arraystretch}{1.2}
\begin{tabular}{p{3.5cm} p{3.2cm} p{4.2cm} p{5.2cm}}
\hline
\textbf{} & \textbf{Data Fidelity Analysis} &  \textbf{Physical Quality Analysis} & \textbf{Task-based Image Quality Approach} \\
\hline 
Examples & Image fidelity, visual perceptual quality, distance, and distribution metrics (e.g., PSNR, SSIM, RMSE, FID, KL divergence, etc.). & Traditional Bench testing approaches (e.g., NPS, MTF, HU accuracy, geometric accuracy, signal ghosting, position accuracy, etc.) \cite{cythia_acr_paper,acr_large_mri_phantom}. & Numerical model observers (e.g., Channelized Hotelling Observer, Non-Prewhitening Matched Filter); forced-choice experiments (e.g., 2AFC, 4AFC); multireader multicase (MRMC) studies \cite{rpz_mo_paper,obuchowski_mrmc_paper}. \\
\hline
Task dependent & No & No & Yes \\
\hline
Designed for or effective in analyzing non-linear methods & No & No & Yes \\
\hline
Purpose & Inference of generic, task-agnostic performance; commonly used for system appraisal \cite{icru_report}. & Inference of generic, task-agnostic performance; commonly used for system appraisal \cite{icru_report}. & Inference of human or model performance for specific tasks (e.g., detection or discrimination of subtle lesions or signals) \cite{cynthia_2afc_paper,yu_dlmo_4r_sr}. \\
\hline
\textbf{Hallucinatory inference} & No & No & No or not explored \\
\hline
\end{tabular}
}
\end{table*}

\IEEEPARstart {A}{rtifacts} in medical imaging are known to stem from imaging physics (such as beam hardening, streaking, and shading artifacts), patients that are being imaged (such as patient implants, motion, and truncation artifacts), and algorithms integrated within a scanning system that transforms raw acquisition data to human-readable output images (such as aliasing, helical interpolation/re-binning, over-regularized/cartoonish reconstruction artifacts) \cite{barrett_ct_artifacts,thorsten_book_artifact}. Medical professionals can often discern (or visually distinguish) that such artifacts are imaging errors and do not represent structures in the patient's internal organs. \blockcomment{Nonetheless, such imaging artifacts may reduce the diagnostic accuracy of the reconstructed images.}

In the current age of Artificial Intelligence (AI), we are encountering a new form of artifact colloquially called ''hallucination''. These artifacts are sometimes also referred to as fakes or pseudo/false structures. Deep learning methods are employed fully or partially - in combination with physics-based methods - to resolve severely undersampled \cite{mri_dc_paper} and severely noisy (i.e., low-dose) \cite{residual_cnn_4r_lowdoseCT} acquisitions. \textit{In such AI-restored\footnote{We use "restoration" as a generic term to describe post-processing in the image domain (such as denoising, deblurring, etc.) or reconstruction of images from their corresponding raw measurements (such as sinogram, k-space data, etc.).} solutions, we occasionally observe \revfresponse{hallucinations (either additive with respect to the addition of false structures or subtractive with respect to the removal of true structures)} that may be confounded with proper anatomy even though, in reality, such structures are not present in the patient being scanned. Such hallucinations are not readily discernible by human eyes, \infoclear{may} easily be misconstrued as real anatomy when analyzed without ground truth, and \infoclear{may} adversely impact clinical performance}. This phenomenon is depicted in fig.\,\ref{img:intro_plot_on_artifacts}. The implant and image-acquisition artifacts are easily distinguishable in fig.\,\ref{img:intro_plot_on_artifacts}(a,b) \tip{even in absence of} their corresponding artifact-free ground truth images. \clinicalcomments{However, the hallucinations depicted in the zoomed box in fig.\,\ref{img:intro_plot_on_artifacts}(c), with the two loops of bowel, only becomes evident after a thorough comparison against its true counterpart in fig.\,\ref{img:intro_plot_on_artifacts}(d), that encompasses only one contiguous loop of the bowel}. 

\tip{
Although the issue of deep learning (DL) induced hallucinations has been widely recognized, \infoclear{current performance testing methodologies may not explicitly evaluate hallucinations as defined in this work.} Full-image based data fidelity metrics like peak-signal-to-noise ratio (PNSR), Structural Similarity Index Measure (SSIM), root-mean-squared error (RMSE) have been shown to not correlate with the preservation of important and subtle features (e.g., lesions, plaques, abscesses) for DL-based methods \cite{kaiyan_breast_cnnio_paper,kayan_bi_sd_on_denoiser}, as illustrated in \tobereviewd{figs.\,S14, S15 in \textbf{Supplementary Information (SI)}}. Physical image quality (IQ) metrics, such as resolution based on modulation transfer function (MTF) analysis, noise texture based on noise power spectrum (NPS), and contrast to noise ratio (CNR), etc., were originally designed to characterize linear systems. Moreover, such  physical IQ metrics are usually measured using uniform phantoms embedded with simple geometric test objects like disk, wires or bar patterns (e.g., ACR CT accreditation phantom \cite{cythia_acr_paper}, ACR MRI accreditation program phantom \cite{acr_large_mri_phantom}). It is obvious that these phantom-based bench-testing performance metrics do not allow us to infer a DL method’s hallucinatory behavior on patient images. 

To address both the inadequacy of data fidelity metrics in measuring the preservation of subtle features and the limitations of linear physical IQ metrics, task-based performance assessment methodologies have been developed. Human observer studies, such as two-alternative forced choice (2AFC) \cite{cynthia_2afc_paper} and multireader multicase (MRMC) studies \cite{obuchowski_mrmc_paper}, as well as numerical model observer studies, including channelized Hotelling and non-prewhitening matched-filter observers \cite{rpz_mo_paper}, have been applied to evaluate the ability of novel restoration techniques to reconstruct low-contrast, small-sized signals in uniform and even real patient backgrounds \cite{kaiyan_breast_cnnio_paper}. However, the outcomes of such observer-based performance testing are limited to the scope of specific signal-detection or signal-discrimination tasks, whereas hallucinations in DL methods are variable and often unpredictable. Hence, it remains difficult to comprehensively characterize the hallucination behavior of a novel DL method using these task-specific approaches. These shortcomings of existing medical imaging performance testing methodologies \infoclear{in the published literature} are summarized in Table \ref{tab:iq_table_summary}. \infoclear{To bridge this gap and enable detection of hallucinations in AI-restored medical images, we propose \textbf{scanning-Fourier Ring Correlation (sFRC)}.}
\blockcomment{Given that medical images form the basis for many advanced downstream applications (such as segmentation, quantitative measurements, 3D visualization, biomarking, and treatment planning), it is not difficult to realize clinical repercussions if DL-based outputs are verified using traditional bench-testing approaches that may not appropriately probe the occurrence of hallucinations.
} 

sFRC performs a local region-of-interest (ROI) or patch-based comparison between images from a novel \textbf{non-linear (such as deep learning or regularization)} method and their reference counterparts at different spatial frequencies to objectively and automatically detect hallucinations. Reference images are obtained using a fully sampled acquisition - restored using analytical methods (like the Filtered Backprojection (FBP) or inverse Fast Fourier Transform (iFFT)). Likewise, sFRC utilizes an internal parameter, the hallucination threshold, which is set using ROIs confirmed by experts to be hallucinated or by imaging-theory-based limiting conditions, to discriminate between hallucinated and faithfully restored ROIs. Our contribution is summarized as follows:
 \begin{enumerate}
    \item [{(1)}] sFRC explicitly outputs candidate hallucinated ROIs that can be readily verifiable by experts and AI developers as hallucinatory or not.
    \item [{(2)}] Due to its local ROI-based construct, sFRC is not skewed by other parts of the image that may have been faithfully restored. Hence, sFRC is consistent with data-processing inequality and hallucinatory results when analyzing non-linear methods, as compared to image-fidelity- or distribution-based metrics.
    \item [{(3)}] sFRC reapplies its tuned knowledge on spatial frequencies that may not have been properly restored---causing hallucinations---to the rest of the testing datasets to detect new types of hallucinations that may be different in terms of intensity, shape, and size.
    \item [{(4)}] sFRC's construct allows one to set its hallucination threshold parameter in the range of relaxed (decreasing hallucination rate, as well as false positives) to aggressive (increasing hallucination rate, as well as false positives). As such, sFRC \infoclear{may} be used to construct a hallucination operating characteristic curve similar to the receiver operating characteristic curve, which summarizes diagnostic performance at different operating thresholds.
\end{enumerate}
}
\tip{\section{Related work on hallucination in image restoration}}\label{sec:related_work}
The issue related to hallucinated structures in deep learning (DL) restored images has also been highlighted in other contributions. Bhadra et al.~\cite{varun_hallu} demonstrated that such false structure may arise due to the reconstruction method incorrectly estimating parts of the object that either did not contribute to observed measurement data or cannot be restored in a stable manner. Similarly, Gottschling et al.~\cite{troublesome_kernel} observed that DL methods for inverse problems are typically unstable and \infoclear{may} cause misdiagnosis (in terms of false positives as well as false negatives of medically important features like tumors). Beyond these references that systematically attempt to define and identify hallucinations in medical imaging, the summary report on the 2020 fastMRI challenge \cite{fast_mri_2020_result} discussed hallucinated artifacts in terms of the addition of false brain structures observed in some of the challenge submissions that used DL methods when reconstructing \revfresponse{undersampled} acquisition. Even in the field of natural image-based applications, one can find results contaminated with hallucinations from DL methods for tasks such as super-resolution (see fig.\,$2$ in \cite{srgan_paper}), domain transfer (see fig.\,$17$ in \cite{domain_transfer_paper}), etc. 

Despite significant progress in mathematical formalism and vast empirical evidence of hallucinations from DL methods, accurately finding such artifacts in a DL output remains challenging. For instance, although Bhadra et al.~\cite{varun_hallu} provided a formal linear operator theory-based mathematical \tip{model} to deduce hallucination maps in medical images restored using AI, their method \tip{involves} the singular value decomposition (SVD) of the forward system matrix \cite{ct_forward_mode} used in a given imaging problem. In real-world applications, developers may not have access to the system matrix used to invert raw measurements or have complete information on acquisition parameters---including details on the imaging scanner---to construct their own system matrix. Notably, the singular value decomposition of system matrices associated with different imaging applications remains a challenging problem---especially memory issues---and an active area of research. Additionally, the initial hallucination maps produced by Bhadra et al.’s method \tip{necessitate} further thresholding to yield specific maps that delineate image features that may have been hallucinated.  The authors indicate that there is no universal rule for selecting such thresholds; consequently, a thresholding scheme \tip{optimized for one restoration problem} may not efficiently identify hallucinated features in a \tip{different problem}. 

Antun et al.~\cite{antun_instabilities} have mathematically demonstrated that instabilities in DL reconstruction methods are not a rare event. Certain tiny imperceptible perturbations in the raw measurements \infoclear{may} lead to myriads of different artifacts (including hallucinations) in the reconstructed medical image. Estimating such perturbation for a given imaging system and a DL solver is a non-trivial problem. Even when such perturbation maps are estimated, it is not trivial to infer what specific signal processing events during the image formation process caused the perturbation. Also, there is no direct route to use such perturbation maps to detect hallucinated features in \tip{restored images}. 

\tip{Distribution-based metrics have also been proposed to quantify hallucinations, such as Tivnan et al.’s Hellinger distance-based metric called Hallucination Index \cite{tivan_hi}. However, distribution- or full image-based metrics may fail to capture locally confined hallucinations. In contrast, our sFRC methodology more effectively detects hallucinations and infers the overall accuracy of restoration methods against hallucinations. A detailed evaluation is provided in Section \ref{sec:compare}.}
\blockcomment{
This manuscript describes an image-processing-based solution referred to as the ''sFRC metric'' or ''sFRC analysis'' to detect hallucinations from DL-assisted imaging methods that seek to mitigate issues related to an undersampled acquisition. sFRC \tip{relies on} reference data from a fully sampled acquisition - restored using analytical methods (like the Filtered Backprojection (FBP) or inverse Fast Fourier Transform (iFFT)) - and a predefined imaging theory-based limiting criteria on what constitutes as hallucinations or clinically validated regions-of-interest (ROIs) as hallucinations. 
}
\blockcomment{
This manuscript is arranged as follows: In section \ref{sec:rationale}, we provide the rationale behind our metric, sFRC, that identifies hallucinations. Section \ref{sec:method} provides the mathematical formulation of sFRC and on the two ill-conditioned \cite{per_just_enough} imaging problems we have considered in this paper, i.e., CT super-resolution and MRI subsampled restoration. In section \ref{sec:results}, we provide a thorough report on hallucinations identified using sFRC for both the ill-conditioned imaging problems.}
\vspace{3mm}
\section{Rationale behind sFRC and its understanding}\label{sec:rationale} 
\begin{figure}[!t]
\centering
\includegraphics[width=1.0\linewidth]{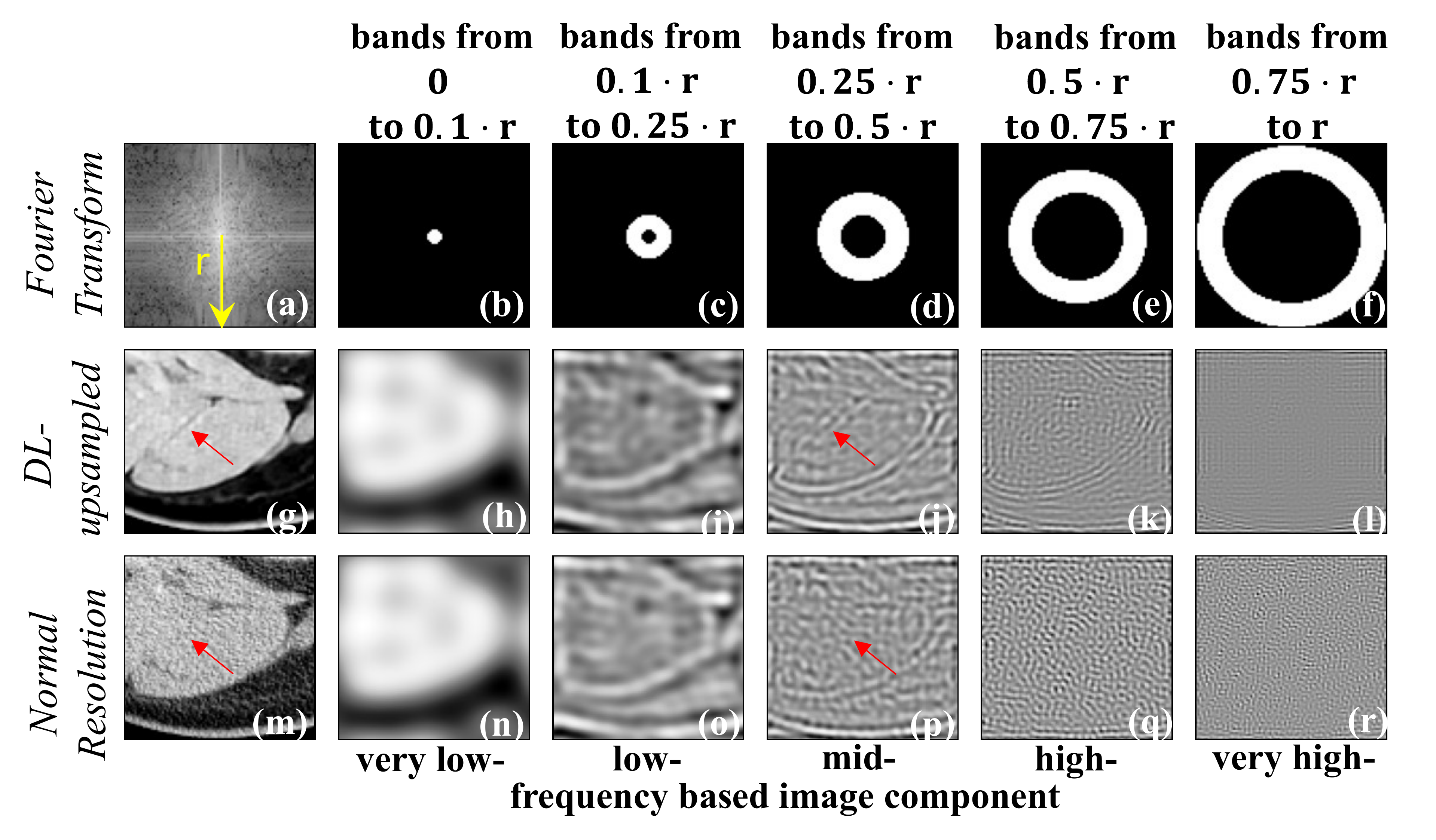}
\caption{Decomposition of a patch containing a hallucinated structure in (g) - indicated by a red arrow - into its frequency-based components in (h-l). (m) is a normal-resolution CT patch. (g) is obtained by \revfresponse{upsampling} (using SRGAN described in section \ref{sec:srgan_model}) the low-resolution (\revfresponse{downsampled} four times) counterpart of (m). (a) is the Fourier transform of (g). (b-f) are different \revfresponse{bandpass filters}. (h-l) are the image components of (g) obtained by \revfresponse{convolving the bandpass filters, as indicated in (b-f), with (g). Similarly, (n-r) are the image components of (m) obtained by convolving the bandpass filters, as shown in (b-f), with (m).}}  
\label{img:fourier_decomposition}
\end{figure}

The following two observations are the basis for the construction of sFRC:
\begin{enumerate}
    \item [{(a)}] Readily non-discernible hallucinations to human eyes are usually limited to small ROIs rather than the entire image. 
    \item [{(b)}] The \revfresponse{hallucinations (either additive or subtractive) from} a DL model may be isolated into certain spatial frequency bands in the Fourier space.
\end{enumerate}

The first observation goes hand-in-hand with how we have defined hallucinations in the introductory section and is depicted in fig.\,\ref{img:intro_plot_on_artifacts}(c). Likewise, a visual depiction of point (b) is provided in fig.\,\ref{img:fourier_decomposition}. The figure illustrates the decomposition of an indentation-like hallucination in a DL solution (indicated by the red arrow) in \revfresponse{fig.\,\ref{img:fourier_decomposition}(g) into its mid-frequency counterpart in fig.\,\ref{img:fourier_decomposition}(j)}. Fig.\,\ref{img:fourier_decomposition}(j) is obtained by \revfresponse{convolving} a bandpass filter in fig.\,\ref{img:fourier_decomposition}(d) with fig.\,\ref{img:fourier_decomposition}(g). The DL output in fig.\,\ref{img:fourier_decomposition}(g) is obtained from a GAN-based CT super-resolution solver detailed below in section \ref{sec:srgan_model}.

The above observations led us to construct a ROI- or patch-, and Fourier component-based evaluation technique. Broadly, \textbf{we seek to compare a DL restored and reference pair at the patch level (and not at the full-image level) to increase our chances of capturing a locally limited hallucination.} Unlike full-image-based analysis, patch-based ensures that when a local hallucination gets captured within a patch, a frequency-based comparison against its reference counterpart does not get overweighted by the overall feature (or the rest of the non-hallucinated features that have been restored faithfully in the entire image). 

Next, different frequency bands incorporate different information on a patch. For instance, column 2 in fig.\,\ref{img:fourier_decomposition}, corresponding to very low-frequency bands, encapsulates information about the blurred version of the original patch (in fig.\,\ref{img:fourier_decomposition}(g,m)). Likewise, for a less-noisy or simulated noise-less acquisition, very high-frequency bands (i.e., column 6 in fig.\,\ref{img:fourier_decomposition}) contain information related to edges (or fine details) present on the original patch. For a typical medical acquisition, a very high-frequency image component will be dominated by noise. As such, we expect image components corresponding to very low-frequency bands from a reference and DL pair to be very similar, irrespective of the presence or absence of hallucinations in the DL patch. We expect patch components corresponding to very high-frequency bands from a reference and DL pair to be very dissimilar, even if the DL patch does not contain hallucinations. This is because even the noise terms in two full image-based scans---from a repeated fully sampled acquisition of an object (keeping all other acquisition parameters and the restoration algorithm constant)---become more dominant than the signal contents at very high-frequency bands. Correspondingly, their very high-frequency-based image components will be very dissimilar \cite{vincent_paper}. Let alone patch-based very high-frequency image components from two different methods. Therefore, using very high-frequency bands to compare reference and DL patch pairs will label almost every DL patch as a hallucinated ROI. 

For the two imaging problems considered in this paper---CT image downsampling by factor
$4$ and MRI subsampled acquisition by factor $3$---we expect the downsampled CT images and subsampled MRI acquisition to contain at least $25\%$ and $33.33\%$ of the Fourier information (i.e., $0.25$ and $0.33$ of the Nyquist frequency ($r$\tip{$=\frac{1}{2\times \text{pixel size}}$}). Combining these two pieces of information---on the Fourier contents along very-high and very-low to low-frequency bands---we expect isolating hallucinations along mid-frequency bands (as shown in the figs.\,\ref{img:fourier_decomposition}(j,p)). In this mid-frequency region, if the mismatch between complementary pairs of reference and DL patches falls below a pre-set hallucination threshold (either set by imaging scientists based on their tolerance for hallucinations to make a clinically accurate decision or by using an imaging theory-based criteria on hallucinations), the DL patches will be labeled as hallucinations. Importantly, once a hallucination threshold is set, it is automatically and repeatedly applied to detect hallucinations across the stack of testing images. The patches to be compared are obtained using bounding boxes that are scanned across the DL and its reference counterpart. Hence, we name the metric \textbf{scanning-FRC (sFRC)}. Mathematical details on sFRC are provided in \ref{sec:sFRC}, and how to set its parameters is discussed in \ref{sec:tune_sfrc_4r_ct}, \ref{sec:sfc_tuning_4r_mr}, and \ref{sec:ct_sparse_case_study} with additional details in section \tobereviewd{S1, SI}.

Note that this high-level explanation for determining the frequency bands/regions to isolate hallucinations \infoclear{is expected} to be appropriately re-evaluated considering the imaging problem at hand (such as a more aggressive undersampling). It might be the case that due to the ill-conditioning of a given problem, very-low and low-frequency regions (columns 2 and 3 in fig.\,\ref{img:fourier_decomposition}) might be relevant to isolate hallucinations (\tobereviewd{see fig.\,S12, SI}). Similarly, it might also be more useful to consider high-frequency region (column 5 in fig.\,\ref{img:fourier_decomposition}) to isolate hallucinations if one were to consider a less ill-conditioned problem or use a scanning device that is capable of resolving features at high spatial resolution (such as micro-CT) than that considered in this paper.
\vspace{-0.3cm}
\section{Method}
\label{sec:method} 
This section describes the mathematical formulation of our metric, sFRC, and the three medical applications (i.e., CT super-resolution, MRI subsampled restoration, and \tip{CT sparse view}) in which sFRC \infoclear{was} employed. 
\vspace{-0.5cm}
\subsection{sFRC}\label{sec:sFRC}
\vspace{-0.5cm}
\begin{figure}[!thb]
\centering
\includegraphics[width=0.7\linewidth]{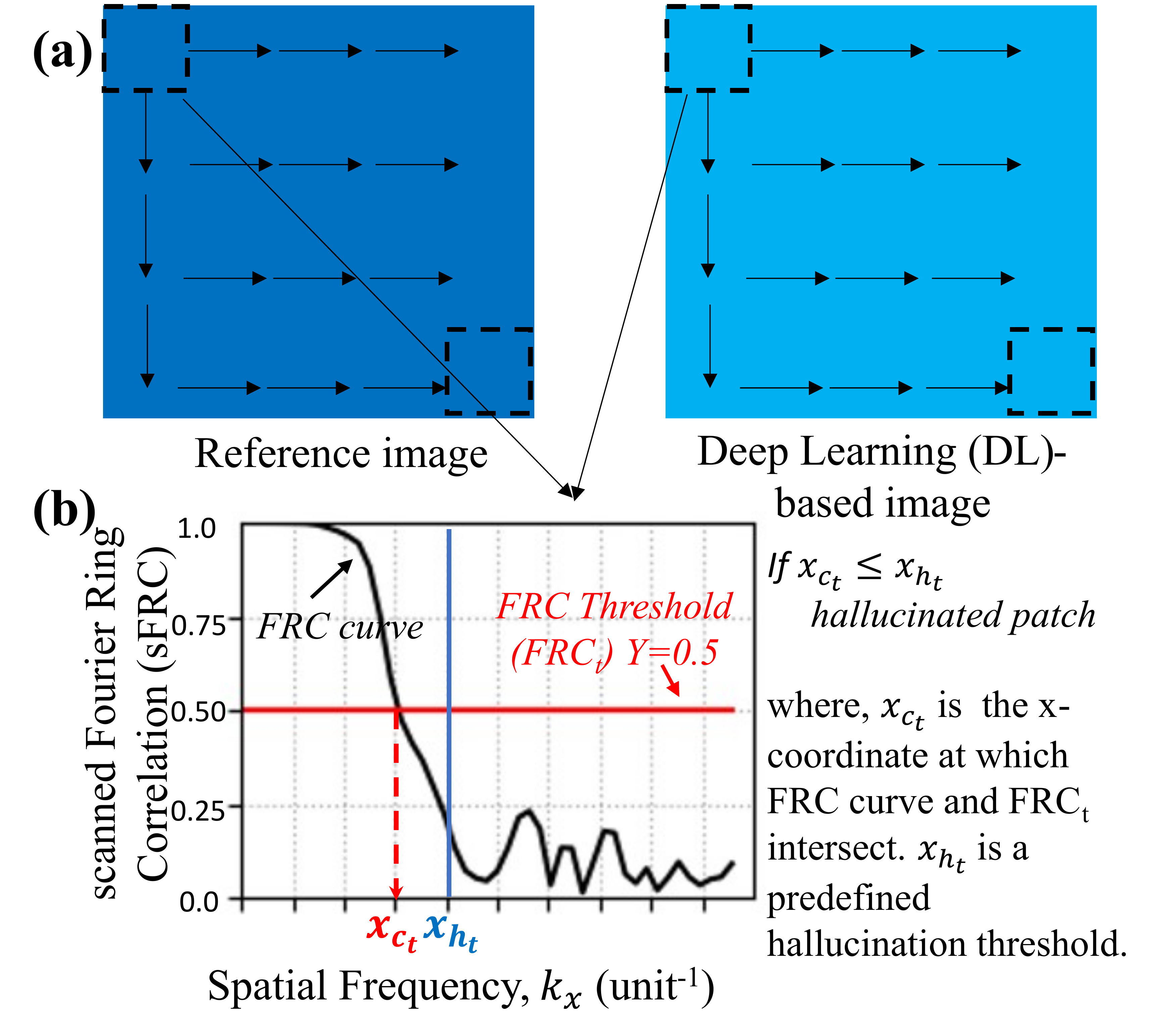}
\caption{A visual depiction of our sFRC analysis to detect hallucinated regions. Complimentary pairs of patches (i.e., over the same x-y coordinate) of the same dimension are scanned across the image pairs from the two methods (reference and deep learning images) in (a). FRC is calculated in (b) across all patch pairs from (a). In addition to the FRC threshold used in a typical FRC calculation, we introduce a new threshold called $x_{h_{t}}$, a line parallel to $y-$axis. Patches corresponding to the deep learning method whose FRC comparison against their reference counterparts lead to $x_{c_{t}} \leq x_{h_{t}}$ are labeled as candidates to exhibit hallucinations. As shown in fig.\,\ref{img:set_xht_general}, $x_{h_{t}}$ is set a priori using hallucinated ROIs/maps estimated by an imaging theory or \revfresponse{annotated by subject matter experts} using tuning or developmental set.}
\label{img:sfrc_visualization}
\end{figure}

Fourier ring correlation provides a measure of similarity between two images as a function of spatial frequency \cite{saxton_82}. The Fourier-based similarity analysis is calculated in a ring-wise fashion and is mathematically expressed as \cite{preFRC_86}:
\revfresponse{
\begin{equation}
\text{FRC}(q_{i})=\frac{\underset{q\,\epsilon\,\text{ring }q_{i}}{\sum}\hat{f}_{1}(q)\cdot\hat{f}_{2}^{*}(q)}{\sqrt{\underset{q\,\epsilon\,\text{ring }q_{i}}{\sum}|\hat{f}_{1}(q)|^{2}\underset{q\,\epsilon\,\text{ring }q_{i}}{\sum}|\hat{f}_{2}(q)|^{2}}}\label{eq:frc_eq},
\end{equation}
}
where the numerator indicates the dot product of the 2D Fourier components \revfresponse{of all pixels, $q$, that are contained in the ring, $q_{i}$,} of the image pairs, $f_{1}$, and $f_{2}$. $\hat{f}_{1}$ and $\hat{f}_{2}$ are the Fourier transforms of $f_{1}$ and $f_{2}$. $\hat{f}_{2}^{*}$ indicates conjugate of $\hat{f}_{2}$. The square root of the sum of the power spectrum corresponding to the rings of the two transforms in the denominator normalizes the FRC values to the range $[0,1]$.

FRC is typically calculated using images with independent noise realizations, i.e., images acquired from the same object but from different, usually consecutive scans or measurements. Likewise, the FRC curve, in combination with a pre-set empirical threshold curve (commonly referred as FRC threshold), is used to estimate resolution for imaging techniques such as electron microscopy \cite{malhotra_em_app}, optical nanoscopy \cite{opt_nano_app},  nanoscale X-ray computed tomography\cite{vincent_paper}. However, this manuscript uses FRC to compare outputs/images obtained from two methods (i.e., reference standard against DL method). A visual depiction of extracting complementary pairs of patches (or ROIs) from the two images (corresponding to outputs retrieved from two different methods), scanning the complimentary ROIs across the image pairs, and calculating FRC on each patched pair is provided in fig.\,\ref{img:sfrc_visualization}. 

The sFRC analysis encompasses two important  variables, $x_{c_{t}}$ and $x_{h_{t}}$. We define $x_{c_{t}}$ as the $x-$coordinate at which the FRC curve---generated from a pair of patches---intersects with an FRC threshold (such as the $0.5$ threshold). Next, hallucination threshold ($x_{h_{t}}$), a line parallel to $y-$axis, bounds all the candidate patches with hallucinations in DL outputs.  
\begin{figure}[!thb]
\centering
\includegraphics[width=1.0\linewidth]{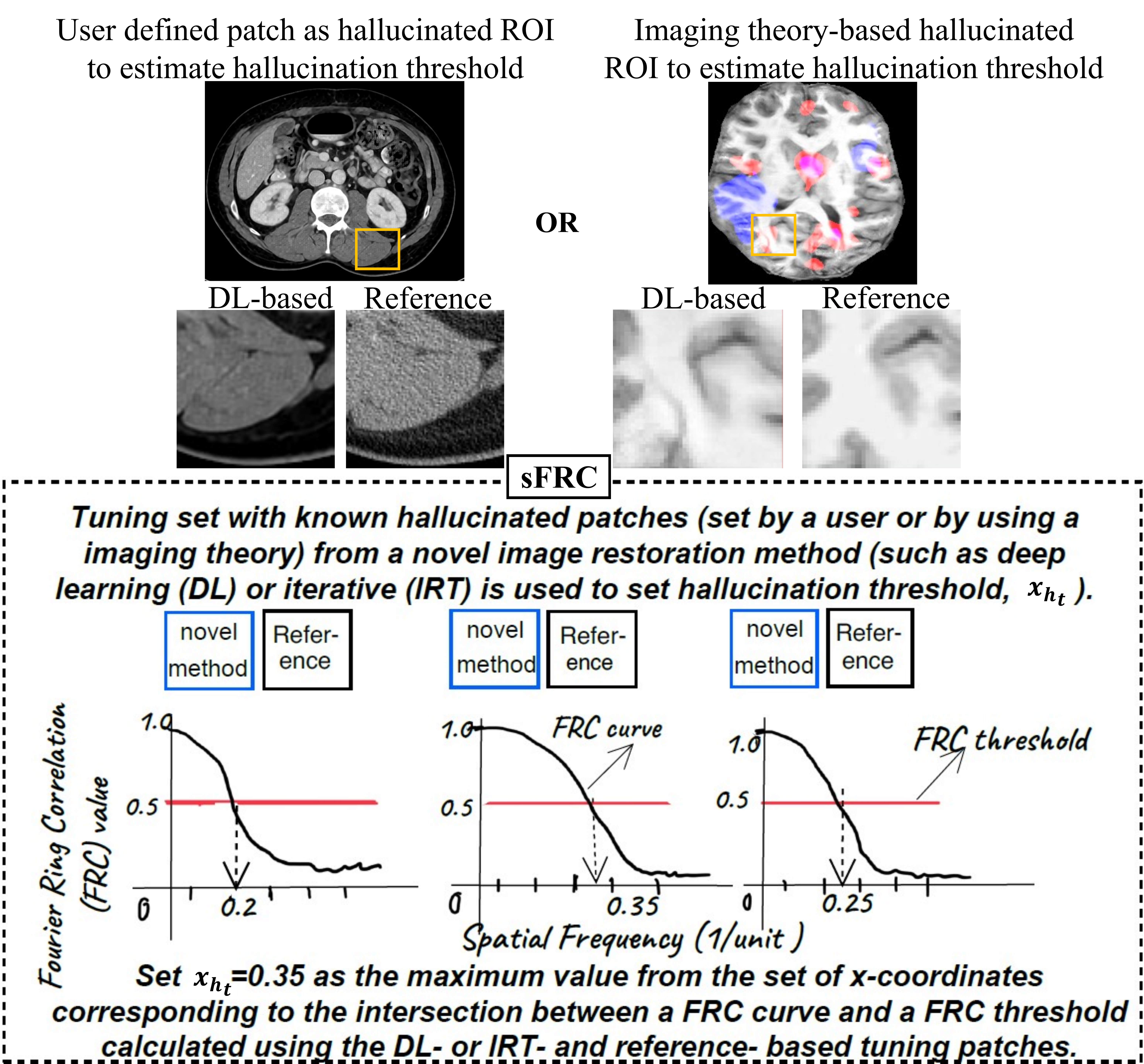}
\caption{An illustration for setting $x_{h_{t}}$ as an upper bound of $x_{c_{t}}$s using patches that are labeled as hallucinations by subject matter experts or using an imaging theory.}
\label{img:set_xht_general}
\end{figure}
\RestyleAlgo{ruled}
\begin{algorithm}[t]
\tip{
\caption{Workflow for tuning sFRC parameters and applying tuned sFRC on a test set}
\label{alg:sfrc}
\SetAlgoLined

\KwIn{Tuning set (with annotated hallucinations), test set}
\KwOut{Hallucinated patches indicated as bounding boxes in test set images}

\SetKwFunction{Tune}{tune\_sFRC}
\SetKwFunction{Apply}{apply\_sFRC}
\SetKwProg{Fn}{Function}{}{end}
\SetKwComment{AlgComment}{//}{}

Choose patch size for tuning and test sets \AlgComment*[r]{ e.g., $32 \times 32$, $48 \times 48$, $64 \times 64$}
Choose FRC threshold ($Y$) 
\AlgComment*[r]{$Y=k_{y}$, where $k_{y}\ \epsilon  \ [0,1]$. Select $Y$ based on the imaging system’s resolving capacity (e.g., $Y \sim 0.75$ for MR, $Y \sim 0.5$ for CT (\tobereviewd{See fig.\,S7, SI})).}

\Fn{\Tune{$Y$, tuning set}}{
    Initialize an empty list $\mathcal{X}_{c_{t}}$\;
    \ForEach{annotated patch pair in tuning set}{
        Compute FRC between patches restored using reference and Deep Learning (DL) (or any non-linear) methods per eq.\ \ref{eq:frc_eq}\;
        $x_{c_{t}} \leftarrow$ $x$-coordinate of intersection between the FRC curve and the threshold $Y$\;
        Append $x_{c_{t}}$ to $\mathcal{X}_{c_{t}}$\;
    }
    $x_{h_{t}} \leftarrow \max(\mathcal{X}_{c_{t}})+\epsilon$ per eq.~\ref{eq:xht_bound_eq}\;
    \Return{$x_{h_{t}}$}
}
\BlankLine
\Fn{\Apply{test set, $x_{h_{t}}$, $Y$, patch size}}{
    Initialize an empty set $\mathcal{H}$\ \AlgComment*[r]{hallucinated patches}
    \ForEach{patch pair in test set}{
        Compute FRC between patches restored using reference and Deep Learning (DL) (or any non-linear) methods per eq.\ \ref{eq:frc_eq}\;
        $x_{c_{t}} \leftarrow$ $x$-coordinate of intersection between the FRC curve and the threshold $Y$\;
        \If{$x_{c_{t}} < x_{h_{t}}$}{
            Mark the DL patch as \textbf{hallucinated}\;
            Add the DL patch to $\mathcal{H}$\;
        }
    }
    \Return{the test-set images with red bounding boxes marking using the hallucinated patches in $\mathcal{H}$}
}}
\end{algorithm}

For a given imaging modality and acquisition condition, one \infoclear{may} use a tuning/developmental dataset to set $x_{h_{t}}$. For instance, using a tuning set, identify ROIs that are clinically or conclusively labeled as hallucinated ROIs (as depicted in figs.\,\ref{img:intro_plot_on_artifacts}(c) \infoclear{and} \ref{img:fourier_decomposition}(g)). Such hallucinations may be identified manually by experts like imaging scientists and clinicians (as shown in our first case study, \ref{sec:expert_annotation}), or by employing imaging theory-based principles (as shown in our second case study, \ref{sec:hall_map_n_unet}). Suppose there are $k$ such ROIs, then set $x_{h_{t}}$ such that: 
\begin{equation}
    x_{h_{t}}=\max(\{x^{(1)}_{c_{t}}, x^{(2)}_{c_{t}}, ..., x^{(k)}_{c_{t}}\})+\epsilon,\label{eq:xht_bound_eq}
\end{equation}
where $\epsilon > 0$ is an arbitrary, very small positive number. So intuitively, $x_{h_{t}}$ acts as a threshold such that if the mismatch between a pair of patches (here between reference and DL methods) falls at or below (or to the left of) this threshold, then the DL patch will be labeled as a candidate to exhibit hallucinated structure. 

\tip{
The vertical line $x_{h_{t}}$ in the sFRC plot has a range $[0, r]$. \textbf{The closer $x_{h_{t}}$ is to Nyquist frequency ($r$), the stricter the sFRC analysis becomes (i.e., a larger number of patches are labeled as hallucinated).} This occurs because sFRC includes an increasingly broader frequency range—spanning from low to high frequencies (i.e., from image blurred components to edges and fine details)—in its correlation analysis between reference and DL image pairs. Accordingly, \textbf{as $x_{h_{t}} \rightarrow 0$, the sFRC analysis becomes more relaxed.} This behavior of $x_{h_{t}}$ is visually illustrated in \tobereviewd{fig.\,S6 in  SI}.

When tuning ($x_{h_{t}}$) and applying sFRC, the FRC threshold ($Y$) also plays a critical role. In our sFRC construction, $Y$ is defined as a line parallel to the $x-$axis and is set within the range $[0, 1]$. So, the behavior of $Y$ aligns with the FRC-based correlation itself (i.e., high coorelation is expected as $Y \rightarrow 1$). \textbf{This means as $Y \rightarrow 1$, sFRC becomes stricter and as $Y \rightarrow 0$, sFRC is more relaxed.} Further, $Y$ is set depending on the resolving capacity of the imaging system. As $Y \rightarrow 0$, sFRC targets higher frequency bands (aligning with imaging modalities that have high resolving power), whereas as $Y \rightarrow 1$, sFRC targets lower frequency bands (aligning with imaging modalities that have lower resolving power). This behavior of $Y$ is visually illustrated in \tobereviewd{fig.\,S7 in SI}. 

The pseudo-code as a workflow for tuning the sFRC parameters and applying the method to the test set is provided in algorithm \ref{alg:sfrc}.
}
\tip{\subsection{CT super-resolution case study}\label{sec:ct_case_study}}
\subsubsection{\tip{Super-resolution GAN (SRGAN) model}}\label{sec:srgan_model}
We \tip{employed deep architectures within a generative adversarial network (GAN) model to complete a CT super-resolution task \cite{srgan_paper}}. We simulated low-resolution CT images, $\mathbf{X}_{\text{L}}$, using their normal-resolution CT counterparts, $\mathbf{X}_{\text{N}}$, in the following manner \cite{srcnn_prior}:
\begin{equation}
\textbf{X}_{\text{L}}=(\mathbf{B} \otimes \textbf{X}_{\text{N}})\downarrow_{s},\label{eq:LR-NR-model}
\end{equation}
where $\mathbf{B}$ is a blurring kernel and $\downarrow_{s}$ is a downsampling operator with a factor $s$. We considered a simplified version of the super-resolution model by not including noise terms in eq.\,\ref{eq:LR-NR-model}.  

The upsampled counterpart, $\Tilde{\textbf{X}}_{\text{N}}$, of $\mathbf{X}_{\text{L}}$ was restored after minimizing the discriminator and generator losses---shown in eqs.\,\ref{eq:discriminator_loss} \infoclear{and} \ref{eq:generator_loss}---in an alternating manner:
\begin{equation}
\begin{array}{cc}
\theta_{D}\leftarrow & \underset{\theta_{D}}{\text{arg min}}\left\{ -\frac{1}{m}\overset{m}{\underset{i=1}{\sum}}\log D_{\theta_{D}}(X_{N}^{(i)})\right.\\
 & \left.-\frac{1}{m}\overset{m}{\underset{i=1}{\sum}}\log(1-D_{\theta_{D}}(G_{\theta_{G}}(X_{L}^{(i)})))\right\} 
\end{array}\label{eq:discriminator_loss}
\end{equation}

\begin{equation}
\begin{array}{cc}
\theta_{G}\leftarrow & \underset{\theta_{G}}{\text{arg min}}\left\{ \frac{1}{m}\overset{m}{\underset{i=1}{\sum}}||X_{N}^{(i)}-G_{\theta_{G}}(X_{L}^{(i)})||_{2}^{2}\right.\\
 & \left.-\frac{10^{-3}}{m}\overset{m}{\underset{i=1}{\sum}}\log(D_{\theta_{D}}(G_{\theta_{G}}(X_{L}^{(i)})))\right\} 
\end{array}\label{eq:generator_loss}
\end{equation}

The discriminator network and SRResNet (with $16$ residual blocks)---detailed in \cite{srgan_paper}---were used to set our GAN-based CT super-resolution model’s  discriminator, $D_{\theta_{D}}$, and generator,  $G_{\theta_{G}}$, networks. Their corresponding network weights, $\theta_{D}$ \infoclear{and} $\theta_{G}$, in eqs.\,\ref{eq:discriminator_loss} \infoclear{and} \ref{eq:generator_loss}, were updated using the  $\{\mathbf{X}_{\text{N}}, \textbf{X}_{\text{L}}\}$ training pairs. $\mathbf{X}_{\text{N}}$ was set using the $3$ mm (thick) \infoclear{and} B$30$ (smooth kernel) CT scans provided in the Low-Dose Grand Challenge (LDGC) dataset \cite{ldct_data_2016}. $\mathbf{X}_{\text{L}}$ was obtained from $\mathbf{X}_{\text{N}}$ by using eq.\,\ref{eq:LR-NR-model}, where $\mathbf{B}$ was set as an identity (or do-nothing) kernel and $s$ was set as $4$. This CT super-resolution model will be hereafter referred to as SRGAN.

\subsubsection{\tip{SRGAN model training and testing datasets}}
\label{sec:train_srgan_model}
The $\{\mathbf{X}_{\text{N}}, \textbf{X}_{\text{L}}\}$ training pairs were set using six patients data (L$096$, L$143$, L$192$, L$291$, L$286$, L$333$) from the LDGC dataset. We employed patch-based training over $96\times96$-sized patches that were normalized (to a $[0,1]$ data range) and augmented (with scaling-, rotation-, and flipping-based augmentations). The discriminator and generator losses in eqs.\,\ref{eq:discriminator_loss} \infoclear{and} \ref{eq:generator_loss}, were minimized using stochastic gradient descent and Adam optimizers with $10^{-6}$ and $10^{-4}$ as learning rates respectively. For a methodological overview of how to set various data-, optimization-, or hyperparameter-based options for deep learning-based training, refer to \cite{pkc_deep_ct}. 

\tip{To evaluate training data sufficiency, we trained SRGAN using data from eight patients (by adding patient L$109$ and L$310$ from the LDGC \cite{ldct_data_2016} in the set $\{\mathbf{X}_{\text{N}}, \textbf{X}_{\text{L}}\}$). No PSNR or SSIM improvement was observed on the sharp CT test set (\tobereviewd{See S4, SI and Table I, SI}). Therefore, all SRGAN results reported in this paper are based on the six-patient training data.}

We used CT scans covering the region from the upper abdomen to the lower leg of patient L$067$ as the test set. They contain images reconstructed using the filtered backprojection (FBP) algorithm with the smooth kernel (B$30$) and a sharp kernel (D$45$) kernel at a $3$ mm thickness \tip{to formulate in- and out-of-distribution-based test sets} for our SRGAN model (trained using smooth $3$mm CT images). \tip{The sharp kernel-based test set was deliberately used to stress-test outputs from our SRGAN model against hallucinations.}
\subsubsection{\tip{Annotating hallucinations to tune sFRC parameters}}
\label{sec:expert_annotation}
This annotating process has been thoroughly covered in the \tobereviewd{S1, SI}. To summarize, we used 5 CT images (\tobereviewd{as shown in subplot (d) in figs.\,S1 through S5}) from the test set corresponding to the patient L$067$ to annotate hallucinated ROIs to tune the sFRC parameters. After a consultation with our internal medical officer, we were told/shown that SRGAN outputs in the tuning set exhibited many subtle irregularities that \infoclear{may} have clinical significance depending upon clinical applications/settings. These five images were contrast-enhanced abdominal CT scans from the portal-venous phase obtained using patients with positive cases in terms of the presence of hepatic metastasis \cite{ldct_data_2016}. Hence, we zeroed in on contrast-based plaque-like artifacts (as shown using the green-boxed ROI in fig.\,\ref{img:intro_plot_on_artifacts}(c)) as a critical hallucination to be annotated for our sFRC analysis. Likewise, based on the literature survey \cite{fast_mri_2020_result,antun_instabilities,mri_fn_annotation_paper}, we also noted indentations and perturbation/ unnatural demorphing as other important hallucinations to be annotated in the tuning set. We meticulously annotated these three types of hallucinations in the tuning set of sFRC. This process of identifying the three types of hallucinated features was performed on SRGAN outputs as well as their corresponding mid-frequency-based image components \tobereviewd{(as shown in subplot (f) in figs.\,S1 through S5)}. 

\subsubsection{\tip{Tuning sFRC parameters using expert-based annotations}}
\label{sec:tune_sfrc_4r_ct}
The sFRC parameters comprise \textbf{patch size, FRC threshold, and hallucination threshold} ($x_{h_t}$). When tuning patch size, we found that using patch sizes ($\geq96\times96$) decreases the chances of correctly detecting hallucinations, for the imperceptible hallucinations are usually limited to small ROIs. Using a large patch size overinflates the correlation outcomes from the faithfully restored regions. Likewise, using small patches (usually, $\leq 32 \times 32$) increases the likelihood of wrongly labeling ROIs as hallucinations (i.e., an increase in false positives). We found that the sFRC analysis using $64 \times 64$ sized patches yields optimal outcomes in terms of maximizing true positives and minimizing false positives for the super-resolution problem. We set $0.5$ as the FRC threshold, as it would target the FRC curve drop-off of the hallucinated ROIs along the rings, $0.25\cdot r \text{ to } 0.5\cdot r$ \tobereviewd{(see fig.\,S7, SI)}. 

\tip{As previously explained in section \ref{sec:rationale}, setting  $x_{h_{t}}$ as $0.25 \cdot r (\approx 0.25\times\frac{1}{2\times0.48})$  mm$^{-1}$, we expect almost no patch to be identified as hallucinated by sFRC because we downsampled CT images by a factor of $4$ (\tobereviewd{see fig.\,S8, SI})}. $0.48$ mm is the pixel size of the CT image. Accordingly, \tip{per algorithm\,\ref{alg:sfrc}}, $x_{h_{t}}$ resulted in $0.35$. We relaxed $x_{h_{t}}$ from $0.35$ to $0.33$ before applying sFRC to the testing data. This is primarily because, at a slightly lower hallucination threshold, sFRC was still able to detect the three types of hallucinations related to plaques, perturbations, and indentations (\tobereviewd{see fig.\,S11, SI}). \tip{Note that, we are interested in understanding sFRC's robustness} to reapply its tuned knowledge on the spatial frequency-based decay in the FRC curve due to a particular type of hallucination to the rest of the testing datasets where the profile of hallucinations could be quite drastically different in terms of intensity, shape, and size (\textbf{i.e., can the $x_{h_{t}}$ tuned using plaques, perturbations, and indentations detect a new type of hallucination?}). 

\subsubsection{\tip{Super-Resolution Wasserstein GAN (SR-WGAN) model}}\label{sec:srwgan_model}
\tip{To demonstrate the generalizability of our sFRC analysis when applied to results from a more advanced AI model, we trained a Wasserstein GAN with a gradient penalty for the same CT super-resolution problem, using the same six-patient sharp CT dataset. We refer to this model as SR-WGAN. Details of the SR-WGAN training procedure are provided in \tobereviewd{S3, SI}}.
\tip{\subsection{MRI subsampled restoration case study}} 
\label{sec:MRI_case_description}
 \blockcomment{We used this case study to demonstrate that the parameters essential to our sFRC analysis can be tuned by using any other objective assessment-based or imaging theory-based definition of hallucinations.}
\subsubsection{\tip{Linear operator theory–based hallucinations and U-Net post-processing}}\label{sec:hall_map_n_unet}
 We used Bhadra et al.'s \cite{varun_hallu} \tip{approach} to map out hallucinations due to inaccuracies that may result from using a wrong data-driven prior\footnote{Bhadra et al.\, defined wrong data prior in the sense that a U-Net model – trained on adult scans - was used on pediatric scans during the inference/testing phase.} and from using deficient forward- or inverse-imaging processes (such as measurement, regularization parameter, numerical methods,  noise models, etc.) during a reconstruction procedure. The author termed the former inaccuracy a null space hallucination map and the latter an error map. Fig.\,\ref{img:set_xht_definition} depicts hallucinated ROIs due to both types of inaccuracies on a U-Net output. Bhadra et al. trained their image domain-based U-Net to learn a mapping that transforms MR images reconstructed at an acceleration factor of $3$ to their corresponding images, reconstructed at a fully-sampled rate for a \revfresponse{stylized single-coil MR forward model}. Their training data employed $2$D axial adult MR brain scans provided in the fastMRI data repository \cite{fastMRI_data_repository}. \blockcomment{For a thorough detail on how Bhadra et al. incorporated forward/inverse MR models and trained the U-Net in their study, we refer to the authors' paper in \cite{varun_hallu}.} 

We used the source codes, pre-trained network weights, and demo test dataset (labeled as out-of-distribution (OOD)) provided by Bhadra et al. in their GitHub repository at \url {https://github.com/comp-imaging-sci/hallucinations-tomo-recon}. \revfresponse{This OOD data corresponds to five skull-stripped T1-weighted magnitude-only MR images sequestered in a pediatric epilepsy study conducted by Mallo et al.\,\cite{MR_ped_epilepsy}. These images exhibit a $1\times1$ mm$^{2}$ resolution, $[-0.05, 0.53]$ range, and were considered as ground truth in our study. Their corresponding fully sampled k-space data was also obtained from Bhadra et al.\,'s repository. Gaussian noise had already been added to the k-space data by the authors (see fig.\,\ref{img:row_based_gt_ifft_unet}(b) rows 1 and 2). Using the background of the MR images reconstructed by applying the iFFT operator on the fully sampled k-space data, we estimated the standard deviation of the added noise to be $\sim\,0.02$.} The subsampled MR images were simulated by applying Cartesian equidistant undersampling masks and iFFT operations - for acceleration factors $1, 2,\text{ and } 3$ - incorporated in fastMRI code package \cite{fastMRI_data_repository}. 

\subsubsection{\tip{Tuning sFRC Parameters Using Imaging Theory–Based Hallucination Maps}}\label{sec:sfc_tuning_4r_mr}
\begin{figure}[!hbt]
\centering
\includegraphics[width=1.0\linewidth]{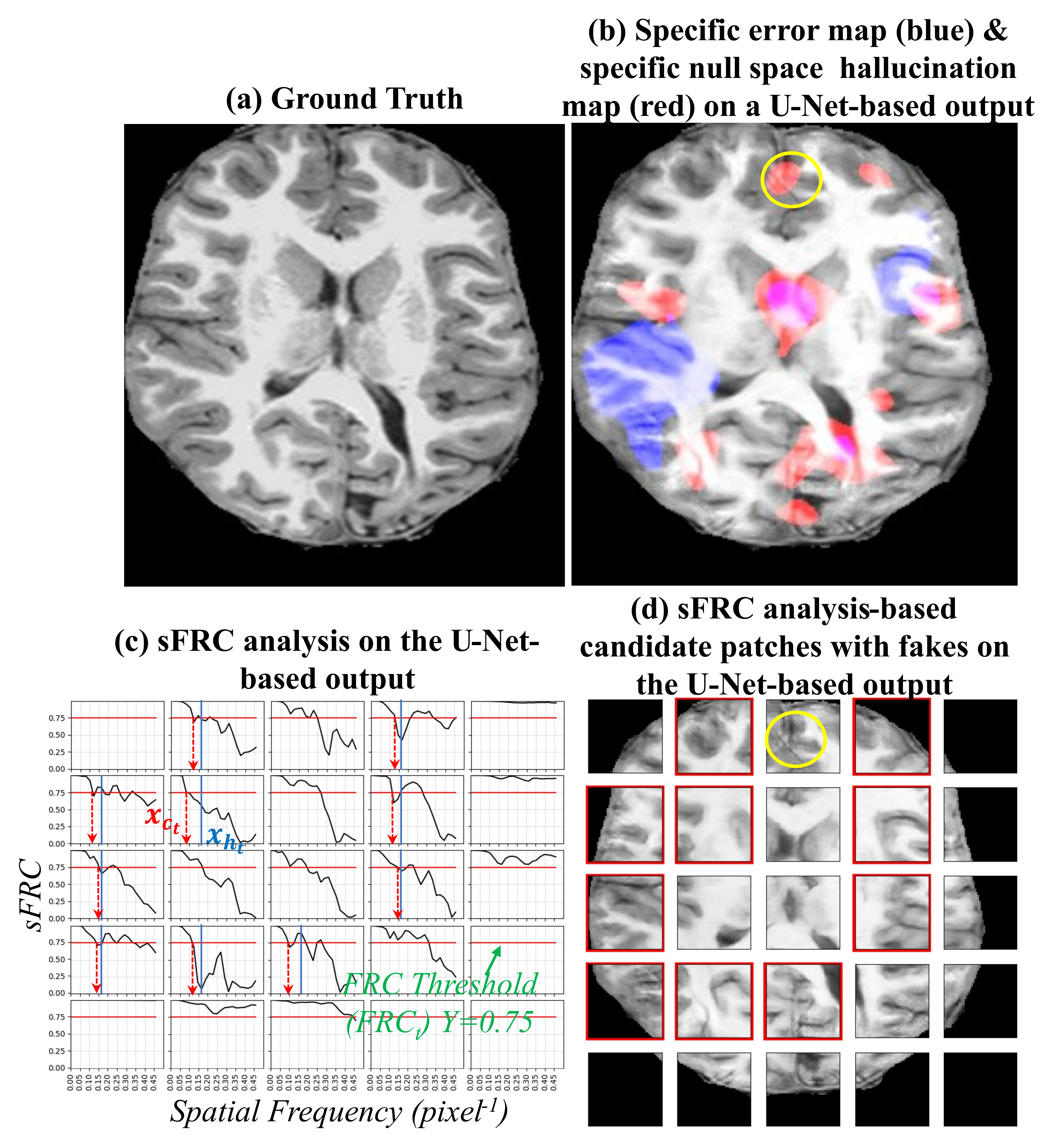}
\caption{An illustration of the imaging theory-based approach to set $x_{h_{t}}$. In our sFRC analysis of the MRI output post-processed, using a U-Net in (c), $0.75$ is set as its FRC threshold and $0.16$ as its $x_{h_{t}}$ such that the patches labeled as hallucinations from our analysis (indicated by the red bounding boxes in (d)) overlap with the false structured obtained using a theoretical \tip{approach} proposed by Bhadra et al. in \cite{varun_hallu}.}
\label{img:set_xht_definition}
\end{figure}

We applied the trained U-Net to all five subsampled MRI images restored at an acceleration factor of $3$ and used Bhadra et al.'s \cite{varun_hallu} \tip{approach} to map out hallucinations in the five images. Out of the five outputs, one was used to tune the sFRC parameters (also depicted in Fig.\,\ref{img:set_xht_definition}(b)), and the remaining four were used in the final analysis in section \ref{sec:MRI_result}. 
Using the tuning image, we systematically searched across different patches sizes (i.e., $32\times32$-sized, $48\times48$-sized, and $64\times64$-sized patches), different FRC thresholds (i.e., $Y=[0.5,0.8]$ at a step-size of $0.05$) and different hallucination thresholds, $x_{h_{t}}$ (with the initial guess of $x_{h_{t}}$ set as $\frac{1}{3}\times0.5\times\text{pixel}^{-1}$). Accordingly, we found that the setting that comprised $48\times48$-sized patches, $0.75$ as the FRC threshold, and $0.16$ as the $x_{h_{t}}$ (i.e., the upper bound of all $x_{c_{t}}$s corresponding to the hallucinated ROIs as per eq.\,\ref{eq:xht_bound_eq}), yields the optimal overlap in hallucinations obtained using our approach and Bhadra et al.'s approach. \tip{Note that, compared to the tuning subroutine in Algorithm \ref{alg:sfrc}, this procedure indirectly allows one to tune sFRC parameters.} Figs.\,\ref{img:set_xht_definition}(c-d) depict the outputs from our sFRC analysis after the tuning process. A visual comparison of hallucinated ROIs in figs.\,\ref{img:set_xht_definition}(b) and \ref{img:set_xht_definition}(d) reveals that the ROIs from the two methods mostly overlap. Nonetheless, there is not a \revfresponse{hundred} percent agreement between the ROIs from the two methods. For instance, look at the ROIs within the yellow rings in figs.\,\ref{img:set_xht_definition}(b, d). Even though we tuned the sFRC parameters using the hallucinated ROIs (blue and red) obtained from  Bhadra et al.'s \tip{approach}, the inner workings of the two approaches are inherently different. At the basic level, our approach performs Fourier ring-based comparison over $2$D patches. In contrast, Bhadra et al.'s approach works by determining the null component of the system matrix (used in a reconstruction). As such, some differences in regions labeled as hallucination between the two approaches are an expected outcome. 

\subsubsection{\tip{PLS-TV reconstruction model and sFRC tuning for PLS-TV}}\label{sec:plstv}
We also employed our sFRC to analyze the reconstructed OOD test set obtained using a conventional regularization method called the Penalized Least Squared with the total variation prior (PLS-TV). We considered the PLS-TV method in this paper to demonstrate the effectiveness of the sFRC in detecting hallucinations for any method (beyond deep learning) that seeks to resolve ill-conditioned imaging problems faithfully. The PLS-TV reconstruction was, in turn, completed using the Berkeley Advanced Reconstruction Toolbox (BART) \cite{bart}. The regularization parameter for the PLS-TV was set as $0.02$. Like the U-Net case, we used the PLS-TV counterpart of fig.\,\ref{img:set_xht_definition}(b) to tune the sFRC parameters for detecting hallucinations in PLS-TV outputs. Accordingly, an optimal outcome was obtained for a setting that comprised $48\times48$-sized patches, $0.75$ as the FRC threshold, and $0.17$ as the $x_{h_{t}}$.

\tip{\subsection{CT sparse view case study} \label{sec:ct_sparse_case_study}
Although in the previous CT and MRI case studies, the sFRC parameters were shown to be tuned using radiologist-annotated and image-theory-based hallucination maps, we would like to acknowledge that the sFRC parameters have explicit physical relevance to the specific image modality and restoration task. We demonstrate this for a CT sparse view problem resolved using a state-of-the-art (SOTA) reconstruction algorithm called Progressive Artifact Image Learning (PAIL) \cite{pail_paper}.  Refer to \tobereviewd{S5, SI} for a detailed description of the AI prior, compressed-sensing prior, and CT-physics constraints incorporated in PAIL. 

Concretely, $x_{h_{t}}$ can be initialized as sampling rate $\times$ Nyquist frequency and sequentially increased to analyze what percentage of fully sampled data is recovered without sFRC labeling any patch as hallucinated. See \ref{sec:sfrc_on_sparse_view} for results from this type of hallucination analysis using sFRC. 
}

All the codes and data used in this manuscript can be found at \url {https://github.com/DIDSR/sfrc}.
\begin{figure}[!bt]
\centering
\includegraphics[width=0.90\linewidth]{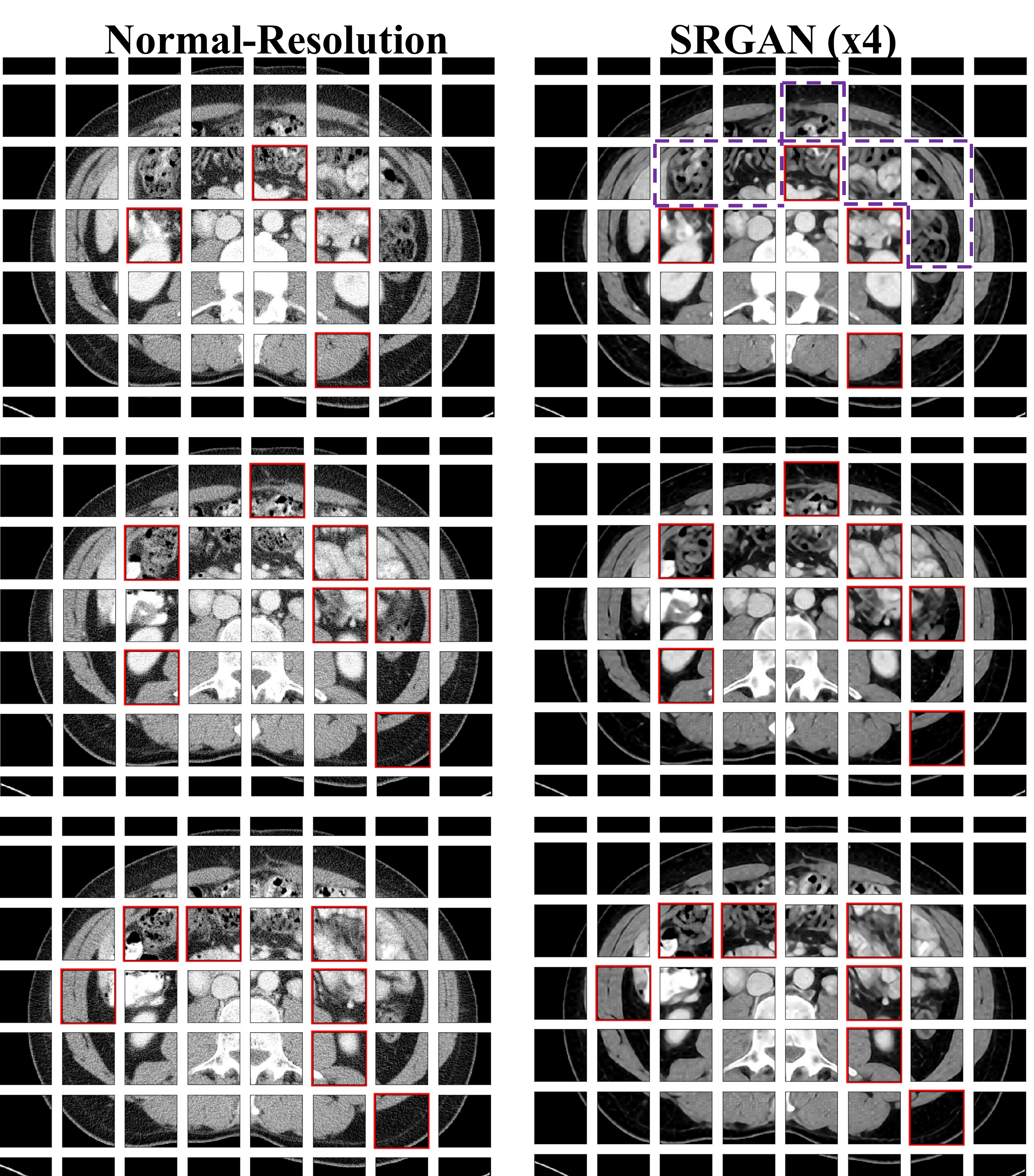}
\caption{The purple ROIs indicate hallucinated patches in the SRGAN outputs that were initially missed by our sFRC analysis in the \tip{top right} scan (000077.IMA). The regions corresponding to the missed hallucinated patches get flagged in the subsequent CT scans in the \tip{center right} (000080.IMA) and the \tip{bottom right} (000082.IMA)  within the abdominal CT volume of patient L$067$ \cite{ldct_data_2016}. Display window is (W:$400$ L:$50$).}
\label{img:ct_test_3d_robustness}
\end{figure}
\section{Results}\label{sec:results}
\subsection{sFRC analysis on CT super-resolution}
\subsubsection{\tip{Stack-based effectiveness of sFRC}}\label{sec:sFRC_result_in_stacks}
After a comprehensive study of the outputs from our sFRC analysis, we found that sFRC works more effectively when it processes a stack of images from a particular body part (or clinical application). This is shown in fig.\,\ref{img:ct_test_3d_robustness}. The figure illustrates three neighboring normal-resolution and the SRGAN $2$D CT pairs (in the sharp test set) that correspond to the abdominal region of patient L$067$. The patches with the red bounding box indicate hallucinated ROIs  from sFRC analysis. On each of the CT images in the right column, we can easily observe multiple hallucinated patches that were missed by our approach. In particular, the ROIs bounded using the purple dotted line in the top right image in fig.\,\ref{img:ct_test_3d_robustness}. However, the missed ROIs eventually get flagged in the subsequent slices in the second and third rows in the figure. We do not have any specific formula to perfectly capture (or predict the pattern of) all the hallucinations that may be present in a DL solution. We rely upon our a priori hallucinated ROIs from a tuning set (that in turn may be associated with certain organ- or structure-based  features) to set $x_{h_{t}}$ and patch size. As such, a hallucinated ROI may be distributed over multiple patches. Subsequently, the distributed hallucinations may be small enough (and even be surrounded by uniform areas) such that their corresponding $x_{c_{t}}$s may exceed $x_{h_{t}}$ and be missed by sFRC analysis. Nonetheless, our $2$D sFRC \tip{approach} is able to efficiently flag the hallucinated patches over a stack of $2$D images for a given volume. 

\begin{table}[!hbt]
    \tip{
    \begin{center}
    \caption{sFRC-detected hallucinations on sharp and smooth test sets for SRGAN and SR-WGAN models. Both  test sets consist of $188$ CT images (i.e., $188\times64$ patches)}
    \label{tab:CT_ood_ind}
    \begin{tabular}{l|c|c}
	\hline
	  Model (test set kernel) & No. of patches detected as & sFRC hallucination\\ 
    & hallucinations & rate (in $[0,1]$) \\
	\hline
	SRGAN (sharp) & $787$ &  $0.0653$\\
	SRGAN (smooth) & $451$ & $0.0374$\\
    \hline
    SR-WGAN (sharp) & $901$ &  $0.0748$\\
	SR-WGAN (smooth) & $502$ & $0.0417$\\
	\end{tabular}
	\end{center}
    }
\end{table}   
\begin{figure}[!htb]
\centering
\includegraphics[width=1.0\linewidth]{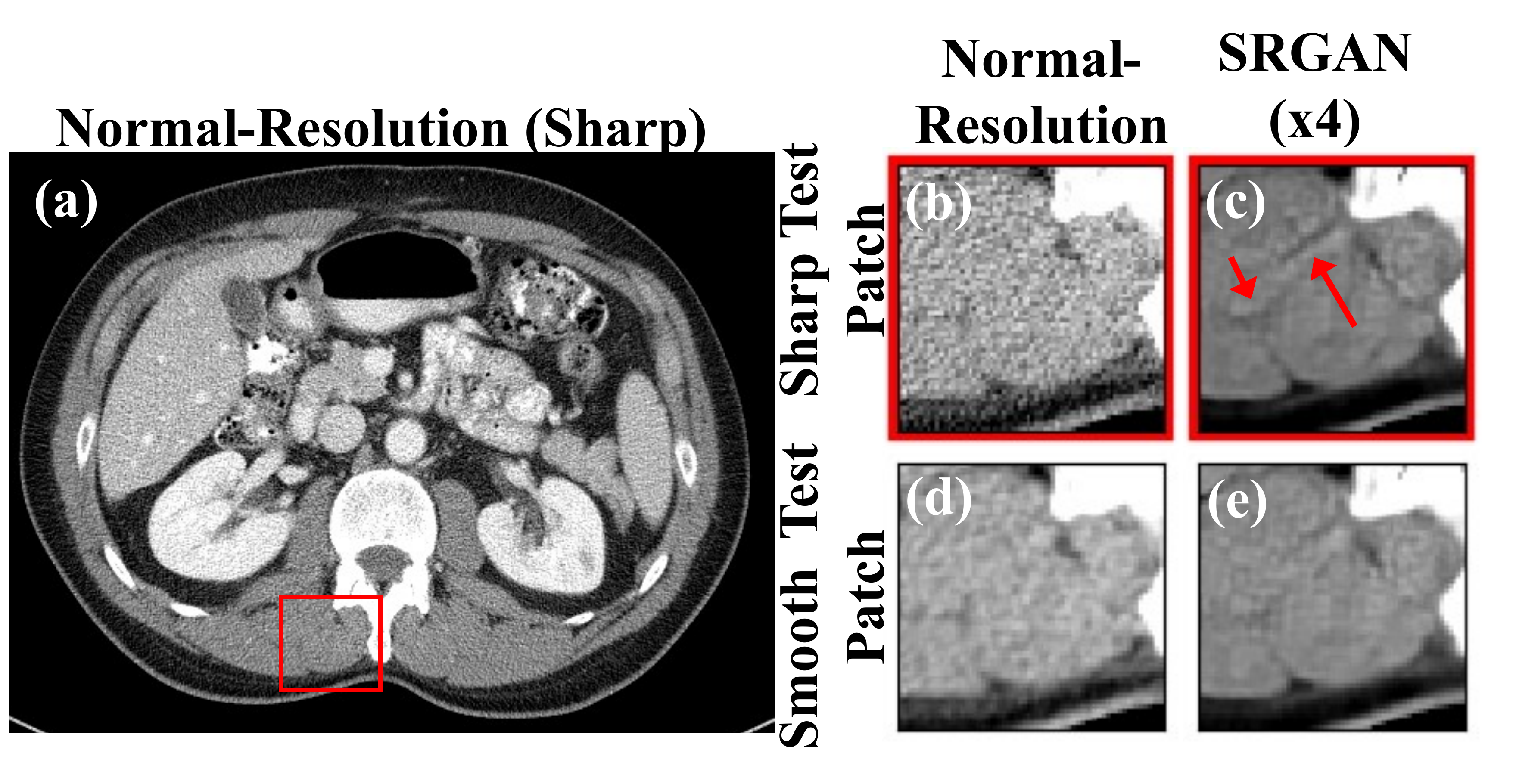}
\caption{An explicit illustration of the decay in performance of the SRGAN super-resolution model - trained using smooth data - when analyzed against (b) sharp and (d) smooth kernel-based normal-resolution patches using our sFRC analysis. The smoothly trained SRGAN generated hallucinated indentations in (c), which were detected by our sFRC analysis. In contrast, the same SRGAN upsampled smooth test data more faithfully in (e) and is not flagged by our sFRC analysis. The patches in (b-e) correspond to an ROI of an abdominal CT scan in (a). Display window is (W:$400$ L:$50$).}
\label{img:sh_sm_pic}
\end{figure}
\subsubsection{\tip{Empirical validation of sFRC under distribution shift}}\label{sec:sFRC_result_in_vs_out}
\begin{figure}[!th]
\centering
\includegraphics[width=0.7\linewidth]{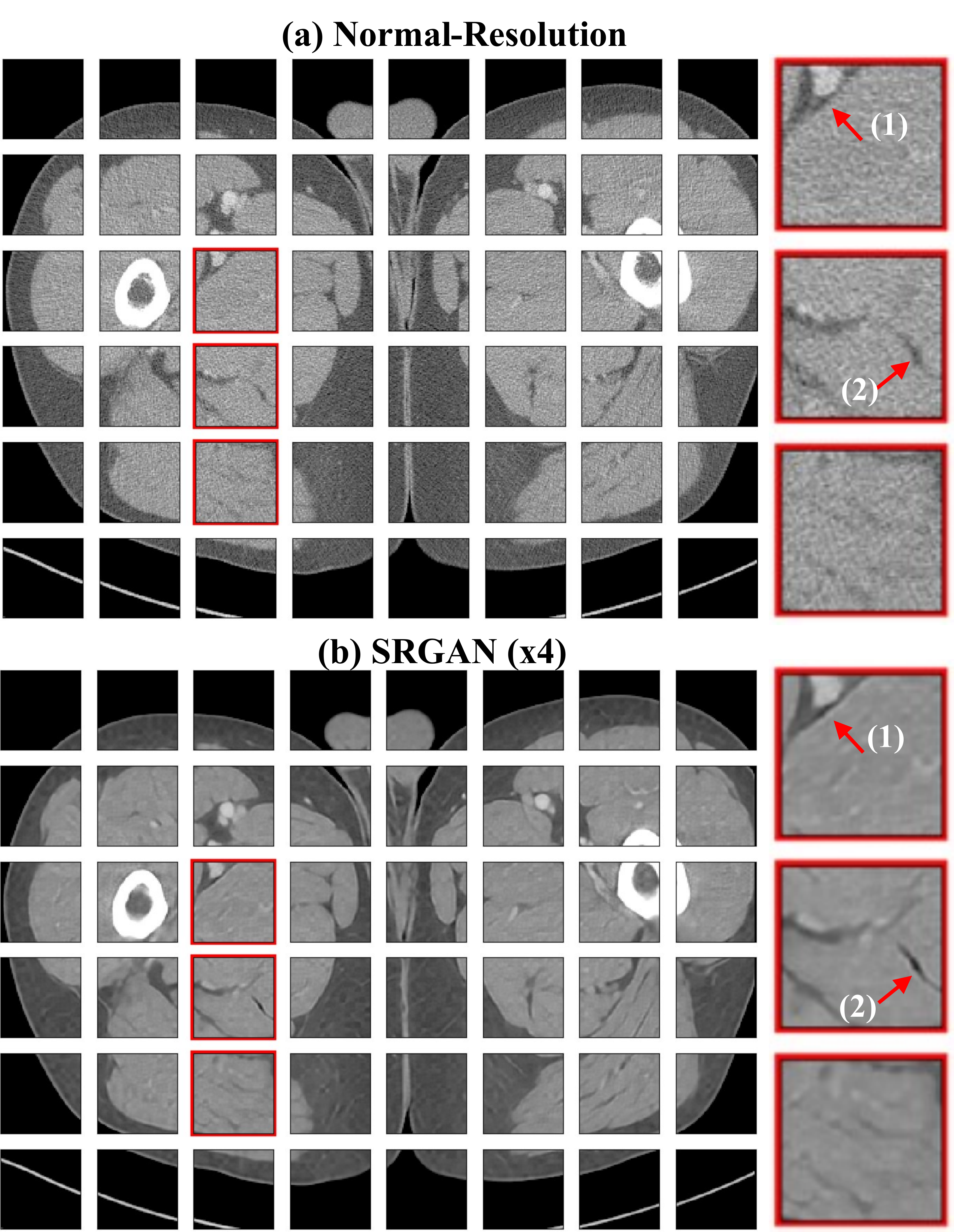}
\caption{\revfresponse{Plaque-like hallucination that is wrongly added along a bordering region (displayed using the arrow (1) and underfitting of fatty attenuation to air (displayed using the arrow (2)) in the SRGAN-based output as compared to its normal-resolution FBP counterpart. Radiologists are trained to look for even small amounts of air. Such air-based hallucinations \infoclear{may} mislead a radiologist’s clinical decision. Display window is (W:$700$ and L:$50$).}}
\label{img:srgan_air}
\end{figure}

Here we demonstrate our sFRC’s ability to reflect the empirically observed behavior of a DL model, i.e., decay in its performance when it processes in-distribution versus out-of-distribution dataset. This is shown  using the smooth counterpart of the sharp test data collected using patient L$067$ (described in section \ref{sec:train_srgan_model}). The number of patches labeled as hallucination on the sharp test data from our analysis was $787$.  This number decreased considerably to $451$ (also listed in Table \ref{tab:CT_ood_ind}) since we trained the SRGAN model using smooth data. Also, the ability of the SRGAN to faithfully \revfresponse{upsample} smooth data (as opposed to sharp data) is depicted in fig.\,\ref{img:sh_sm_pic} (e,c). The figure illustrates hallucinated indentations when the SRGAN processed an out-of-distribution sharp image (indicated by red arrows in fig.\,\ref{img:sh_sm_pic}(c)). This hallucination is substantially diminished when the SRGAN model processed the in-distribution smooth counterpart (in fig.\,\ref{img:sh_sm_pic}(e)). The sharp and smooth test sets with the patches that were identified as hallucinations by our sFRC are provided as movie files at \url {https://fdahhs.box.com/s/vvfcbqxd66a2x09yld1tyk2weqs72i7s}. \clinicalcomments{The hallucinations in the movie files encompassed a wide range of imaging errors, such as \textbf{over-smoothing, in-homogeneity, tiny structural changes, removal of subtle features, distortion of small organelles, addition of minute indentation-/blood vessel-/plaque-like structures, coalescing of small organelles, unwarranted foldings, etc.}} Fig.\,\ref{img:srgan_air} highlights the clinical repercussion of AI-based hallucinations. The SRGAN model underfits fatty attenuation to air (as shown using arrow (2) in fig.\,\ref{img:srgan_air}). This hallucinated ROI can have a severe clinical impact, as radiologists are trained to look for even small amounts of air in unexpected areas (like fatty anatomical regions).
\tip{
\subsubsection{sFRC results on SR-WGAN}\label{sec:sFRC_result_in_srwgan}
Higher PSNR and SSIM values were obtained for both sharp and smooth test datasets from SR-WGAN than SRGAN (Table \ref{tab:CT_MRI_metric_summary}). Yet, reusing the sFRC parameters from SRGAN when analyzing the SR-WGAN results, sFRC-detected hallucinations increased in the SR-WGAN results relative to SRGAN (Table \ref{tab:CT_ood_ind}). Despite this increase, the overall hallucination trend remained similar between the two models. Hallucinations in the SR-WGAN results were distributed across multiple patches, making sFRC more effective at detecting hallucinations across multiple 2D slices (\tobereviewd{fig.\,S16, SI}). SR-WGAN also exhibited more hallucinations in the out-of-distribution sharp test set than in the in-distribution smooth test set (Table~\ref{tab:CT_ood_ind}; \tobereviewd{fig.\,S17, SI}). Finally, SR-WGAN also demonstrated clinically impactful hallucinations, similar to those observed in SRGAN (e.g., an HU-based hallucination in \tobereviewd{fig.\,S18, SI} and fig.\,\ref{img:srgan_air}). Note that, the SR-WGAN result in \tobereviewd{fig.\,S18, SI} exhibits a false-positive patch and a more exaggerated cut-like hallucination. This suggests that the increased hallucination rate for SR-WGAN could be due to false positives or due to SR-WGAN hallucinating more features than SRGAN.}

\begin{figure}[!hbt]
\centering
\includegraphics[width=0.9\linewidth]{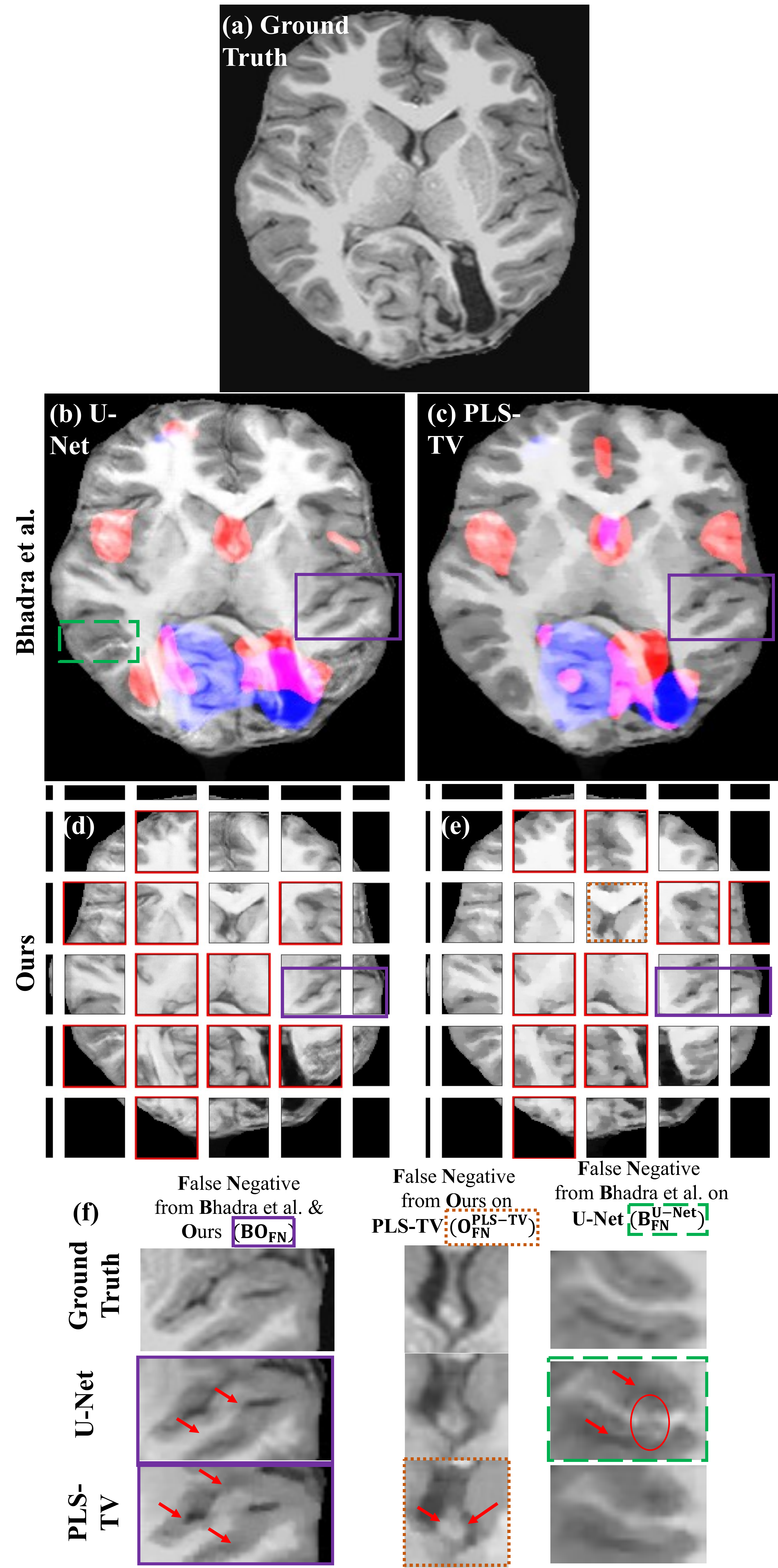}
\caption{A ground truth (in (a)) was used to perform subsampled MRI acquisition by an acceleration factor of 3. The subsampled output is post-processed by a U-Net (in (b), (d)) and reconstructed using PLS-TV (in (c), (e)). Bhadra et. al’s linear operator theory-based \tip{approach} (in (b), (c)) and our sFRC analysis (in (d), (e)) are applied on the U-Net and PLS-TV outputs to identify hallucinations. In fig.\,(f), the first column depicts hallucinated ROIs missed (false negative) by our and Bhadra et al.’s \cite{varun_hallu} approaches; the second column depicts an ROI missed by our approach on the PLS-TV output; and the third depicts an ROI missed by Bhadra et al's approach on the U-Net.}
\label{img:mri_test_result}
\end{figure}
\subsection{sFRC analysis on MRI subsampled restoration}
\label{sec:MRI_result}
\subsubsection{\tip{Qualitative comparison between sFRC and Bhadra et al.’s hallucination maps}}\label{sec:sfrc_vs_varun}
We compared the hallucinations that Bhadra et al. mapped against those given by our sFRC approach on the U-Net and PLS-TV outputs. We found the hallucinations from the two approaches to overlap overwhelmingly. An explicit example is provided in fig.\,\ref{img:mri_test_result}. 

Some ROIs were missed (False Negative (FN)) by both approaches in the test data for the U-Net and PLS-TV outputs. \clinicalcommentr{A particular instance, whereby both approaches missed is depicted in figs.\,\ref{img:mri_test_result}(b,c,d,e) using the purple box. The corresponding zoomed view is presented in the leftmost column of fig.\,\ref{img:mri_test_result}(f) and is labeled as $\text{BO}_{\text{FN}}$. The red arrows in the $\text{BO}_{\text{FN}}$ column indicate \textbf{anomalies related to sulci, constriction of the white matter, and thickening of the grey matter.} Similarly, we found some ROIs missed by one approach while correctly identified by the other. A particular ROI that patently looked to be \textbf{disfigured (via removing sulcus and impacting the overall gyri structures)} on the U-Net output is depicted in fig.\,\ref{img:mri_test_result}\,(b) using the green dotted box. This ROI is not picked by Bhadra et al.'s approach. Its zoomed view is depicted in the rightmost column of fig.\,\ref{img:mri_test_result}(f) and is labeled as $\text{B}^{\text{U-Net}}_{\text{FN}}$. The zoomed view further discloses \textbf{a grey matter migration anomaly} within the red circle. Likewise, our sFRC analysis does not pick up the inaccuracy on the PLS-TV output to preserve the fornices going into the third ventricle. This is labeled as $\text{O}^{\text{PLS-TV}}_{\text{FN}}$ in fig.\,\ref{img:mri_test_result}(f).}
\begin{figure}[!htb]
\centering
\includegraphics[width=0.8\linewidth]{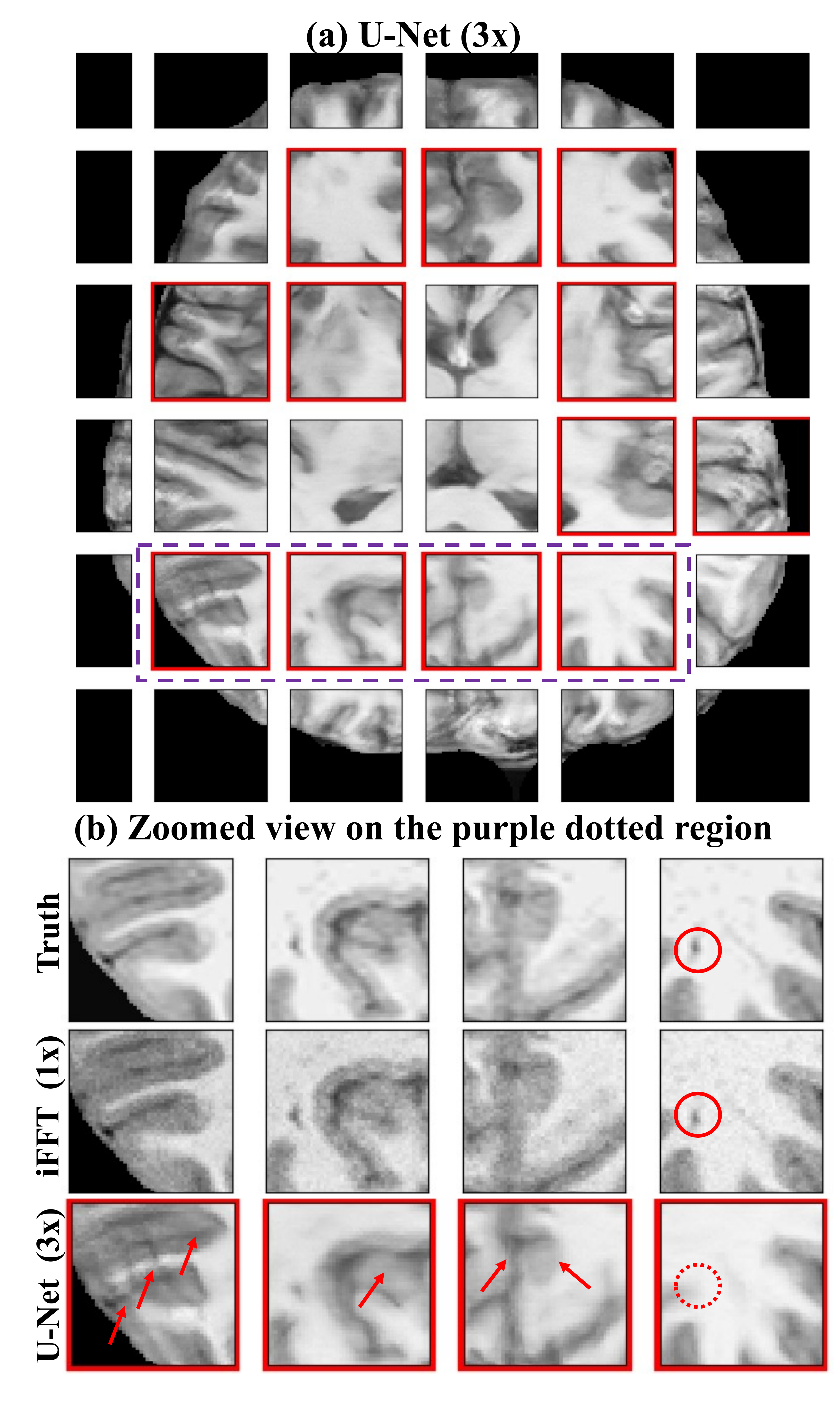}
\caption{(a) Red bounding boxes as hallucinations detected by our sFRC analysis on a U-Net output. It was restored from a subsampled MRI data acquired using an acceleration factor of three ($3$x). \tip{(b) Zoomed view of the purple dotted region from (a). The first two rows in (b) correspond to the ground truth and iFFT ($1$x) counterparts of the purple dotted region from (a).}}
\label{img:row_based_gt_ifft_unet}
\end{figure}

The primary purpose of this MRI case study is not to get a perfect alignment between our approach and a different one on the regions labeled as hallucinations. Instead, it is meant to demonstrate the flexibility of our approach to adhere to a predefined imaging theory-based definition or objectively defined criteria in its discourse to label ROIs as hallucinations. Further, such hallucination detection approaches are generally geared toward finding false structures relative to a specific task and not identifying every possible imaging error with a hundred percent accuracy. 

\subsubsection{\tip{Clinical relevance of sFRC-detected hallucinations}}\label{sec:sfrc_mr_clinical}
\tip{Together with a medical officer, we meticulously reviewed the red bounding–boxed hallucinated patches detected by sFRC across four MRI test images}. Overwhelmingly, we found that the patches labeled as hallucinations by our sFRC analysis exhibited anomalies of one form or the other. An explicit case is provided in fig.\,\ref{img:row_based_gt_ifft_unet}. \clinicalcommentr{The first three columns of the third row in fig.\,\ref{img:row_based_gt_ifft_unet}(b) illustrates \textbf{aliasing, thickening of grey matter with the loss of a subtle sulcus feature, and banding artifacts with black stripes crossing into grey matter} across the three U-Net patches as compared to their ground truth and iFFT ($1$x) counterparts. In the last column in fig.\,\ref{img:row_based_gt_ifft_unet}(b), we can see the \textbf{omission of a dark signal} in the U-Net output}. Still, there might be clinical applications where a loss of minute features might be tolerated in favor of faster acquisition with DL-based restoration. Further study \tip{would be helpful} to determine the optimal binning of pixels during the Fourier ring-based correlation calculations so that sFRC analysis becomes less sensitive to clinically less relevant artifacts for a given task. 
\begin{figure}[!htb]
\centering
\includegraphics[width=0.9\linewidth]{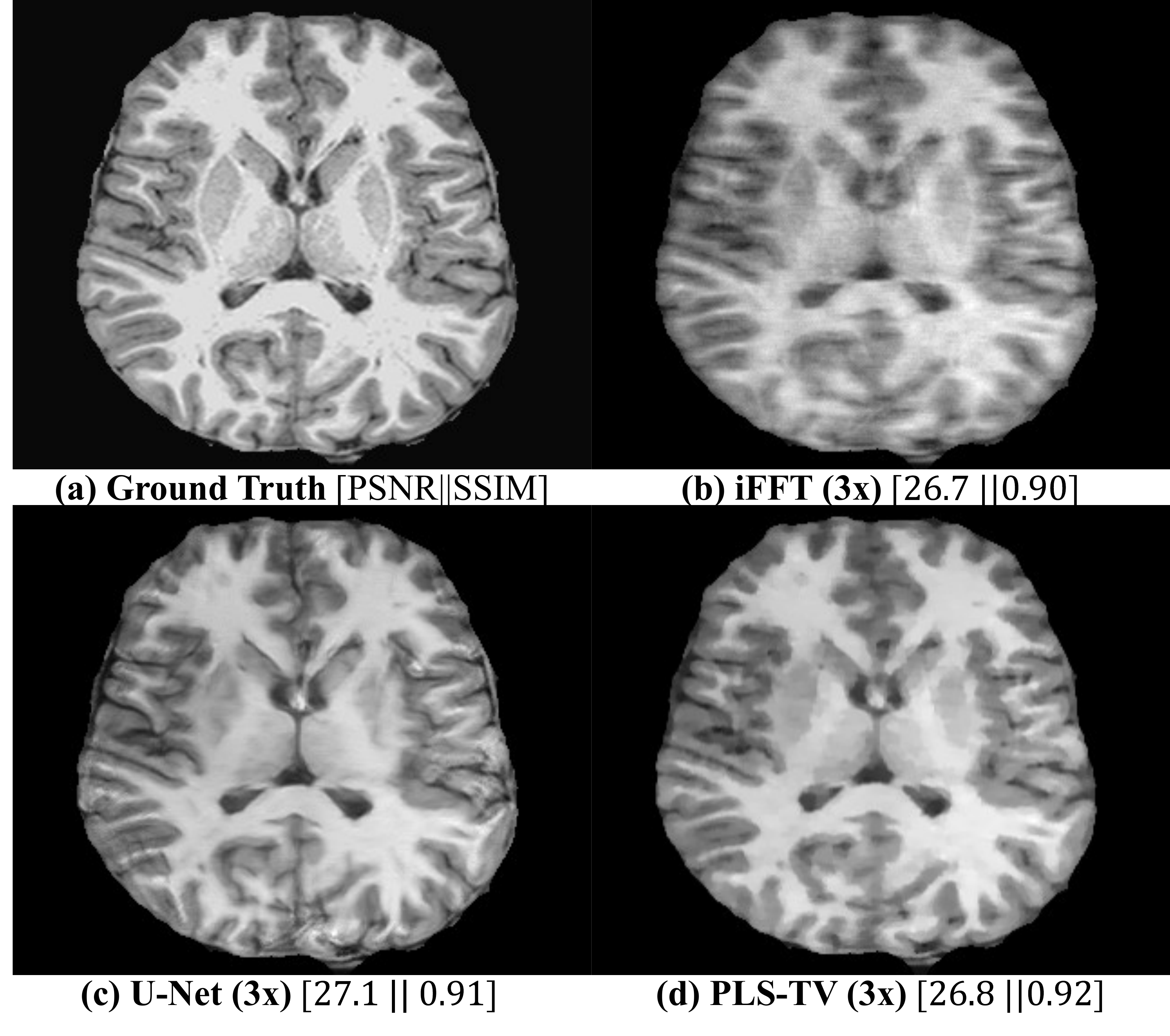}
\caption{(a) Ground truth image and its corresponding (b) iFFT, (c) the U-Net, and (d) PLS-TV outputs restored using a subsampled acquisition (i.e., accelerated by a factor three ($3$x)). PSNR \infoclear{and} SSIM were calculated over signal-only regions. Plot (c) exhibits a high PSNR and SSIM values, and visually it looks as good as plot (a). Still, plot (c) is contaminated with hallucinations, as shown in fig.\,\ref{img:row_based_gt_ifft_unet}.}
\label{img:mri_full_img_per_method}
\end{figure}
\begin{table}[!hbt]
    \tip{
    \begin{center}
    \caption{sFRC analysis-based hallucinations relative to different subsampling rates and restoration methods from the four MRI test scans (totaling $196$ patches)}
    \label{tab:MRI_fakes_per_method}
    \begin{tabular}{l|c|c}
	\hline
	  Method & No. of patches & sFRC hallucination \\
           & detected as hallucinations &  rate (in $[0,1]$) \\
	\hline
	      iFFT ($1$x)   & $2$   & $0.010$\\ 
	      iFFT ($2$x)   & $22$  & $0.112$\\
        iFFT ($3$x)   & $60$  & $0.306$\\
        U-Net ($3$x)  & $43$  & $0.219$\\
        PLS-TV ($3$x) & $42$  & $0.214$\\
    \hline
	\end{tabular}
	\end{center}
    }
\end{table}   

\subsubsection{\tip{sFRC robustness over side-by-side likability inspection}}\label{sec:sfrc_over_likability}
\tip{A side-by-side visual inspection (or Likert study) of the results in fig.\,\ref{img:mri_full_img_per_method} may suggest that the U-Net and PLS-TV reconstructions are more visually pleasing (i.e., receive higher Likert scores) than the standard iFFT method for subsampled acquisitions with an acceleration factor of $3$.} However, as previously discussed and illustrated in fig.\,\ref{img:row_based_gt_ifft_unet} (which employed the same U-Net-based output as in fig.\,\ref{img:mri_full_img_per_method} (c)), our sFRC analysis efficiently flags patches where features may have been wrongly added or removed when modern deep-learning methods like the U-Net (or even a regularization-based spare restoration method like the PLS-TV) are used to restore images from a subsampled acquisition. 

\subsubsection{\tip{sFRC is consistent with data processing inequality}}\label{sec:sfrc_same_as_dpi}
Table \ref{tab:MRI_fakes_per_method} lists the total number of hallucinations detected per restoration method by our sFRC analysis on the four MRI test scans (or equivalently after scanning over $196$ patches as a result of using $48\times48$-sized patches over the four $320\times320$-sized MRI test scans). The table illustrates that as we progressively employed the subsampled acquisition---i.e., by using \revfresponse{$\sim\, 99.69\%$} of the acquired data with the $1$x acceleration to $\sim\,50\%$ with the $2$x acceleration to $\sim\,33.33\%$ with the $3$x acceleration---the number of hallucinations observed in the conventional iFFT approach sequentially increased. This is an expected outcome as the conventional iFFT or the BackProjection techniques were designed to perform a stable reconstruction for the data acquired at a full mode (i.e., for well-conditioned problems) and not for data that is heavily subsampled (such as the $2$x, $3$x cases). On the contrary, deep learning methods (such as the U-Net) \infoclear{and} regularization methods with handcrafted priors (such as the PLS-TV) are proposed to mitigate issues related to the iFFT or the BackProjection method for subsampled acquisition to faithfully \infoclear{and} stably restore the underlying true images. Still, table \ref{tab:MRI_fakes_per_method} lists that the number of hallucinations observed in the U-Net and the PLS-TV outputs for the acceleration factor $3$x continued to remain high and do not reach the level observed with the iFFT at $1$x or $2$x. 

Gleaning through the information in figs.\,\ref{img:row_based_gt_ifft_unet}, \ref{img:mri_full_img_per_method} and table \ref{tab:MRI_fakes_per_method}, it appears that a DL method may yield an output that perceptually looks appropriate from a subsampled acquisition. Still, there is no guarantee that the output from the DL method is faithfully restored, and it may contain imperceptibly localized hallucinations. In this sense, our sFRC analysis subscribes to the \textit{data processing inequality} \cite{dpi_theory}, as the use of data-based or handcrafted prior, may not be sufficient to compensate for the loss of information due to the subsampled acquisition.

\begin{figure*}[!htb]
\centering
\includegraphics[width=1.0\linewidth]{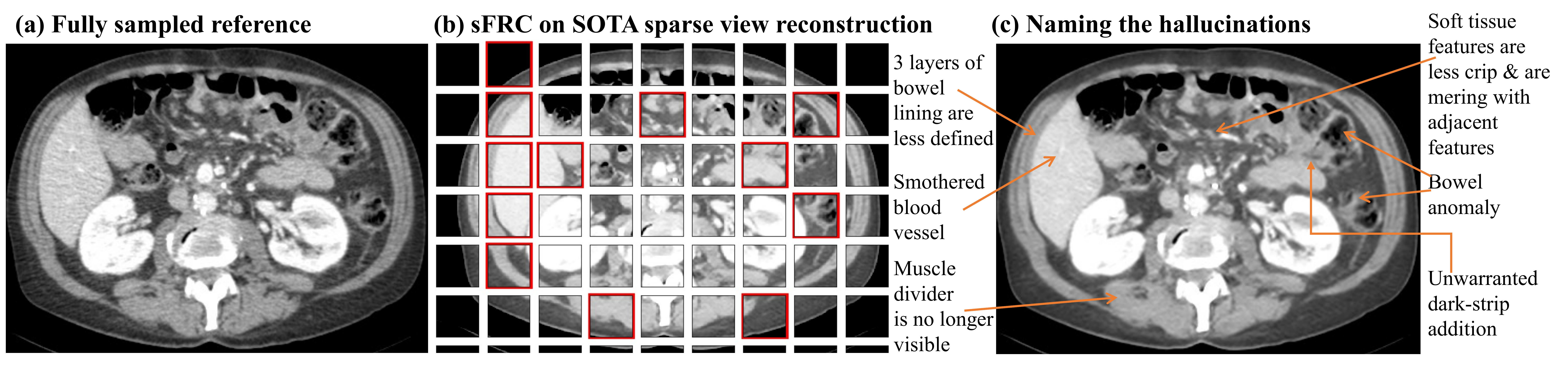}
\caption{\tip{The reference CT image in (a) and its 36 view–based reconstruction from a state-of-the-art (SOTA) model in (c) are processed using sFRC to detect potentially hallucinated local regions, shown as red bounding boxes in (b). Display window is (W:$400$ L:$50$)}}
\label{img:pail_3plot}
\end{figure*}

\infoclear{\subsection{sFRC analysis on CT sparse view problem}\label{sec:sfrc_on_sparse_view}
PAIL reconstructions were obtained using $36$-view CT projection data. These views correspond to a $180^{\circ}$ projection orbit, implying at least $10\% (=(180/360)\times(36/180))$ of the image content to be faithfully restored. Applying sFRC with $x_{h_{t}}=0.33$ (i.e., $31.68\%$ of a fully sampled reconstruction as $0.3168\times\frac{1}{2 \times 0.48}$, where $0.48$ pixel size in mm) and a patch size of $64\times64$ resulted in no patches being labeled as hallucinated.

Increasing $x_{h_{t}}$ to $0.50$ (i.e., $48\%$ of fully sampled reconstruction) and reducing the patch size to $48\times48$ led sFRC to flag multiple hallucinated regions (fig.\,\ref{img:pail_3plot}(b)). In consultation with a medical officer, these findings were validated by direct comparison of sFRC-identified regions with the fully sampled reference image (fig.\,\ref{img:pail_3plot}(a)). \textbf{Confirmed hallucinations included bowel anomalies, smoothing of the inner bowel wall, spurious dark bands, loss of muscle-divider visibility, and reduced soft-tissue sharpness causing apparent feature merging(fig.~\ref{img:pail_3plot}(c)).} These inaccuracies were subtle and difficult to discern without careful comparison to the fully sampled FBP reference.}

\begin{table*}[!t]
\tip{
\caption{Variation in restoration performance across different evaluation metrics. Higher values indicate better performance for PSNR ($\in [0,\infty]$) and SSIM ($\in [0,1]$), whereas lower values indicate better performance for the Hellinger distance ($\in [0,1]$) and normalized sFRC ($\in [0,1]$).}
\label{tab:CT_MRI_metric_summary}
\centering
\renewcommand{\arraystretch}{1.15}
\begin{tabular}{lccccccc}
\hline
    Restoration    & Training & Test & No. of Testing       & PSNR        & SSIM              & Hellinger        & sFRC \\
    Method         & set      & set  & [Images $||$ Patches] &(std)       & (std)             & Distance (std)   & hallucination rate \\
\hline\hline
\multicolumn{8}{|c|}{\textit{CT Super-resolution problem}} \\
\hline
    Bicubic (x$4$) & -      & smooth  & $[188\ ||\ 12032]$ & $33.85\ (0.87)$ & $0.904\ (0.012)$ & $\mathbf{0.216\ (0.012)}$ & $0.368$\\   
    SRGAN (x$4$)   & smooth & smooth  & $[188\ ||\ 12032]$ & $36.74\ (2.73)$ & $0.945\ (0.010)$ & $0.328\ (0.096)$ & $\mathbf{0.037}$\\
    SR-WGAN (x$4$) & smooth & smooth  & $[188\ ||\ 12032]$ & $\mathbf{39.93\ (1.45)}$ & $\mathbf{0.970\ (0.005)}$ & $0.259\ (0.054)$ & $0.042$\\
\hdashline
    Bicubic (x$4$) & -      & sharp   & $[188\ ||\ 12032]$ & $32.86\ (0.76)$          & $0.818\ (0.022)$ & $0.277\ (0.030)$ & $0.390$\\   
    SRGAN (x$4$)   & smooth & sharp   & $[188\ ||\ 12032]$ & $33.72\ (1.95)$          & $0.860\ (0.020)$ & $0.302\ (0.066)$ & $\mathbf{0.065}$\\
    SR-WGAN (x$4$) & smooth & sharp   & $[188\ ||\ 12032]$ & $\mathbf{36.05\ (0.97)}$ & $\mathbf{0.878\ (0.021)}$ & $\mathbf{0.242\ (0.037)}$ & $0.075$\\
\hline\hline
\multicolumn{8}{|c|}{\textit{MR subsampling (accelerating) problem}} \\
\hline
	iFFT ($1$x)   & -     & pediatric & $[4\ ||\ 196]$ & $\mathbf{30.54\ (1.12)}$ & $\mathbf{0.96\ (0.011)}$ & $0.264\ (0.026)$ & $\mathbf{0.010}$ \\
	iFFT ($2$x)   & -     & pediatric & $[4\ ||\ 196]$ & $29.23\ (0.95)$ & $0.95\ (0.012)$ & $0.240\ (0.023)$ & $0.112$ \\
    iFFT ($3$x)   & -     & pediatric & $[4\ ||\ 196]$ & $27.67\ (0.63)$ & $0.92\ (0.014)$ & $0.233\ (0.023)$ & $0.306$ \\
    U-Net ($3$x)  & adult & pediatric & $[4\ ||\ 196]$ & $28.18\ (0.77)$ & $0.93\ (0.014)$ & $0.230\ (0.018)$ & $0.219$ \\
    PLS-TV ($3$x) & -     & pediatric & $[4\ ||\ 196]$ & $28.13\ (0.86)$ & $0.93\ (0.011)$ & $\mathbf{0.223\ (0.043)}$ & $0.214$ \\
\hline
 \textbf{Is the metric consistent}   & -     & -         & -              & No        & No              & No        & \textbf{Yes} \\   
 \textbf{with hallucinatory findings?}    &      &          &               &        &               &         & \\

\hline
\end{tabular}
}
\end{table*} 

\tip{\subsection{Comparing sFRC with another hallucination approach and other metrics}\label{sec:compare}
Tivnan et al. \cite{tivan_hi} introduced a metric called Hallucination Index based on the Hellinger distance to quantify distributional differences between reconstructed and reference images. Our approach is fundamentally distinct. Specifically, sFRC is designed to detect locally confined hallucinatory regions, whereas the Hellinger distance provides a global distributional summary. The sFRC outputs candidate ROIs exhibiting hallucinatory structure, while Tivnan et al.’s method outputs Hellinger distance-based single valued score characterizing the overall discrepancy between two image distributions. As with global fidelity metrics (e.g., RMSE, PSNR) and perceptual metrics (e.g., SSIM), global distributional measures tend to be dominated by the overall image content (or the rest of the non-hallucinatory features restored faithfully in the entire image). Thereby giving a false impression of accurate reconstruction, as illustrated in fig.\,\ref{img:mri_full_img_per_method}, and not being able to capture local inaccuracies corresponding to diagnostically important small features/signals. Full-image–based FRC also suffers from a similar issue, as demonstrated in \tobereviewd{fig.\,S15, SI}.

A further challenge inherent to global or distribution-based metrics is the identification of a clinically meaningful and reusable threshold for distinguishing hallucinated from faithful reconstructions. This limitation is shown in Table \ref{tab:CT_MRI_metric_summary}. The point estimates and standard deviations of the Hellinger distance for four MRI test images (described in Section \ref{sec:MRI_result}) suggest superior performance for U-Net ($3$x) compared to iFFT ($1$x), despite the explicit presence of locally limited hallucinated ROIs in U-Net ($3$x) results as shown in figs.\,\ref{img:mri_test_result} and \ref{img:row_based_gt_ifft_unet}. In contrast, sFRC parameters can be tuned using confirmed hallucinations to automatically differentiate hallucinated from faithfully reconstructed ROIs. Moreover, \textbf{sFRC parameters are directly linked to an imaging system’s sampling rate and pixel size. As shown in section \ref{sec:sfrc_on_sparse_view}, by using the subsampling rate, one can systematically assess what percentage of the fully sampled data is restored by a DL method.}

Likewise, iFFT (1x) achieves the highest PSNR and SSIM values among all methods for the MR subsampling problem, consistent with expectations and with its faithfully reconstructed images. However, U-Net (3x) is only $0.025$ lower in SSIM and $2.41$ dB lower in PSNR than iFFT (1x), indicating that PSNR and SSIM can give a misleading impression of restoration efficacy, in which U-Net (3x) appears comparable to the non-hallucinatory reconstructions from iFFT (1x).

Even for the CT super-resolution problem, SR-WGAN yields better PSNR, SSIM, and  Hellinger distance values than SRGAN for both smooth and sharp kernel CT test sets. However, SR-WGAN exhibits hallucinatory outputs similar to those observed with SRGAN (see \ref{sec:sFRC_result_in_srwgan}). In contrast to these image-fidelity and distribution-based analyses, our sFRC values align with the observation that reconstruction efficacy degrades with increased subsampling and when moving from in-distribution to out-of-distribution data. Importantly, \textbf{sFRC \infoclear{may} be used to estimate hallucination rate of an image restoration method (that has explicit physical relevance in terms of locally limited hallucination numbers). Furthermore, red-bounding-box–based visualizations on restored images \infoclear{may} be used by AI developers to determine the nature of hallucinations, assess the impacted anatomical structures, and accordingly optimize their algorithm.}}
\section{Discussion}
\label{sec:discussion}
\tip{\subsection{Using other metrics for the local ROI analysis}}\label{sec:othermetrics}
In the initial stage of developing a metric to detect hallucinations, besides FRC, we considered other metrics such as the Fr\'echet Inception Distance (FID), SSIM, and normalized RMSE (NRMSE). We used them to analyze hallucinated and non-hallucinated ROIs (similar to those depicted in figs.\,\ref{img:intro_plot_on_artifacts}(c,d) and \ref{img:fourier_decomposition}(g,m), and those reported in literature \cite{fast_mri_2020_result}). Further, using the FID, SSIM, and NRMSE, we analyzed three different \revfresponse{frequency-based image components of hallucinated patches (corresponding to very low-, low- to mid-, and high to very high-frequency bands as depicted in fig.\,\ref{img:fourier_decomposition}}). However, we could not find any pattern that allowed us to categorize hallucination and non-hallucination ROIs systematically. 

In contrast, sFRC was robust in categorizing hallucinated and non-hallucinated ROIs in a sweeping manner. This is primarily because the FRC curve dropped rapidly whenever there was unwarranted addition, removal, distortion, over-smoothing, compression, or coalescing of tiny structures along the mid-frequency components (column 4 in fig.\,\ref{img:fourier_decomposition}). Subsequently, it became easy for sFRC to bound the hallucinated ROIs using its \tip{hallucination threshold} ($x_{h_{t}}$).  We have found sFRC to be generic, which allows one to slide $x_{h_{t}}$---that groups patches into hallucinated and non-hallucinated---from very low- to high-frequency depending upon a given imaging problem. 

For the CT super-resolution problem, the three types of hallucinations - plaques, indentations, and perturbations - were used to tune sFRC. \textbf{In the testing phase, sFRC additionally picked other hallucinations} such as over-smoothing, in-homogeneity, tiny structural changes, removal of subtle features, addition of minute blood-like structures, coalescing of small organelle, and transferring of attenuation. Likewise, even for the MRI subsampling problem---where sFRC was tuned using hallucination maps corresponding to sulci---sFRC was robust enough to pick contrast migration hallucinations and the removal of a dark signal. One may also consider banded frequency-wise difference plots or a metric like the high-frequency error norm (HFEN) \cite{hfen_paper}, in a patch-wise manner, to group between hallucinations and non-hallucinations. However, careful attention \infoclear{is warranted} to ensure that edge-based and high-intensity-based features do not impact such metrics when comparing a given patch pair restored from reference and DL methods.

\subsection{\tip{Generalization of sFRC parameters}}\label{sec:sfrcgeneralization}
sFRC parameters appear to be generalizable within the same imaging problem (i.e., image processing task, acquisition setting, and body part). For example, in the 3x subsampled MRI case, parameters tuned using images restored by PLS-TV and U-Net share the same patch size and FRC threshold, and have similar hallucination thresholds ($0.17$ for U-Net and $0.16$ for PLS-TV). Thus, once calibrated for a given imaging problem, sFRC is expected to efficiently detect hallucinations even when the AI method changes, provided the undersampling level and clinical application remain the same.

This robustness stems from sFRC’s local patch-based design and Fourier-based correlation analysis, which transfer hallucination tolerance learned during tuning into a spatial-frequency threshold that specifies up to which frequencies Fourier content must be reliably recovered for faithful reconstruction. \tip{Moreover, for the same imaging problem, sFRC parameters tuned for an existing method provide a good initialization for evaluating new methods (e.g., those incorporating both physics-based models and AI-based priors) and determining whether previously observed hallucinations have been resolved, as demonstrated with the SR-WGAN model. Accordingly, one can sequentially increase the fraction of fully sampled data, using sFRC’s hallucination threshold, to infer the extent of faithful recovery achieved by the new method.} 

\tip{
\subsection{Potential use of sFRC on other imaging problems}\label{sec:sfrc_potential}
We demonstrated sFRC’s effectiveness in detecting hallucinations in the CT super-resolution problem (restored using DL post-processing), the CT sparse view problem (restored using hybrid physics- and AI-based models), and the MRI subsampling problem (restored using DL post-processing and regularization-based reconstruction methods). Similarly, sFRC can efficiently detect hallucinations when DL denoisers are used to restore images from aggressive low-dose acquisitions. However, sFRC may be sensitive to spurious noise in denoising applications; therefore, additional tuning of the ring bin size may be necessary when performing sFRC analysis for such problems.

Conventional imaging artifacts (e.g., distortion, streaking, and shading) typically span all frequency zones, from very low to very high (\tobereviewd{fig.\,S12, SI}). Consequently, directly applying sFRC to images containing such deterministic artifacts leads to most patches being labeled as hallucinations (\tobereviewd{fig.\,S13, SI}). For visibly perceptible errors arising from linear reconstruction methods (e.g., FBP or iFFT), whole-image FRC—or simply PSNR and SSIM—is suitable for discriminating between accurate and inaccurate reconstructions (see \tobereviewd{S2, SI}). 

For detecting hallucinations in images restored using non-linear methods and acquired with high–spatial-resolution imaging modalities, such as electron microscopy or nano-tomography, \infoclear{it may be advisable to consider} high-frequency regions (i.e., $0.5\cdot r$ to $0.75\cdot r$ in fig.~\ref{img:fourier_decomposition}) when setting the hallucination threshold. Analogously, for imaging modalities with lower resolving power, such as low-field MR, the hallucination threshold \infoclear{may be set along} very-low- and low-frequency regions (i.e., $0\cdot r$ to $0.1\cdot r$ and $0.1\cdot r$ to $0.25\cdot r$ in fig.\,\ref{img:fourier_decomposition}).

Furthermore, in other medical imaging applications such as digital pathology, microscopes are primarily used to scan glass slides, with little to no image-processing involved in restoring whole-slide images (WSI). However, modern image-processing techniques—such as upsampling and virtual staining—are emerging for WSI. Our sFRC approach would be applicable to WSI when DL-based conditional upsampling is performed (e.g., restoring 40× resolution from 20× magnification). In this context, hallucinations introduced by DL methods \infoclear{may} be detected using sFRC, similar to the CT super-resolution case study (\ref{sec:ct_case_study}). In contrast, virtual staining---which is often implemented using generative AI and lacking reference staining images---is not applicable to sFRC.
} 

\tip{\subsection{Current sFRC limitations} \label{sec:sfrc_limiations}
\blockcomment{In its current form, sFRC is not equipped to provide uncertainty estimates or quantify true positives, false positives, true negatives, and false negatives. Please refer to the subsection \ref{sec:hoc_future_work} for details on how we plan to incorporate these elements in future sFRC version.
}
In its current form, sFRC is not applicable for detecting hallucinations on the fly without a reference counterpart. Therefore, \textbf{sFRC is also not applicable to unconditional image generation}, since there is no direct one-to-one corresponding reference image for an AI-generated output. sFRC is only applicable to conditional medical imaging problems—such as deep learning–based post-processing or conventional regularization-based reconstruction—where reference “true” images can be derived analytically under standard acquisition conditions (like the full-dose or full sampling rate) from the object/patient to be imaged.}
\blockcomment{It is important to note that, similar to how SSIM, PSNR, and  NRMSE are commonly used to evaluate AI-restored images, distance-based or distribution-based metrics—such as Hellinger distance, Fréchet Inception Distance (FID), Inception Score, or Jensen–Shannon divergence—are widely used to assess the image quality of AI-synthesized images relative to real images. However, it is well documented that metrics popularized in natural-image applications may not be relevant for downstream medical imaging tasks \cite{varun_gan}. Consequently, it may be necessary to consider clinically meaningful statistical analyses, such as quantitative measures (e.g., gray-to-white matter ratio for brain MR synthesis; fat-to-glandular ratio analysis for breast CT synthesis), class- or disease-prevalence analyses (e.g., prevalence across fatty, scattered, heterogeneous, and dense categories for breast CT synthesis), or the use of statistics beyond means and variances (e.g., morphological features, image moments, skeletal statistics, fractal statistics, etc.)\cite{aapm_23_gan,rucha_high_order}.
}
\subsection{\tip{Study limitations}} \label{sec:study_limitations}
The numerical studies presented here have limitations. In the CT super-resolution case study, we considered a conventional GAN and \tip{a Wasserstein GAN} models instead of more advanced models like the diffusion model (claimed to perform better than GANs \cite{srdiff_paper}). In the MRI subsampled case study, we considered a stylized single-coil-based forward model that incorporated a uniform Cartesian subsampling technique as opposed to a multi-coil-based forward model with a variable-density-based subsampling technique \cite{lustig_cs_paper}. Yet, note that we counterbalanced these two limitations by keeping both problems relatively simple so that the DL methods in the two case studies could accurately restore the images. For instance, in the CT super-resolution study, we incorporated neither the noise model nor the blurring kernel of the CT system used to acquire the test data. Similarly, in the MRI study, we considered an acceleration factor only up to $3$. Still, the outputs from the DL methods were plagued with many subtle hallucinations that our sFRC analysis efficiently identified. \tip{Furthermore, although SR-WGAN was shown to yield higher PSNR and SSIM than SRGAN, it was still plagued by hallucinations that sFRC efficiently detected. Even for the CT sparse-view problem reconstructed using a state-of-the-art model—shown to achieve higher PSNR and SSIM and more stable reconstructions than prior iterative, regularization-based, and deep learning methods—sFRC efficiently detected very subtle hallucinations (as discussed in \ref{sec:sfrc_on_sparse_view}).

Also, note that the performance testing provided in this study is not intended to evaluate a model’s generalizability, covariate effects, or site-based effects. Instead, using sFRC, we demonstrate that DL reconstruction methods may introduce imperceptible hallucinations, even when they achieve very high perceptual, data-fidelity, or distribution-based metrics. In essence,  sFRC is employed to provide  counterexamples to the generally held assumption that DL reconstruction is accurate. We further show that sFRC \infoclear{may} be readily and effectively used to detect such hallucinations in a objective manner, similar to traditional MTF- or NPS-based bench testing for resolution and noise analyses. \textbf{Only after ensuring a reasonable level of faithful restoration (i.e., restoration is indeed devoid of imperceptible hallucinations) \infoclear{may} one proceed to demonstrate a DL method’s generalizability or accuracy across patient-based or scanner-based covariates.}}

\begin{figure}[!thb]
\centering
\includegraphics[width=0.8\linewidth]{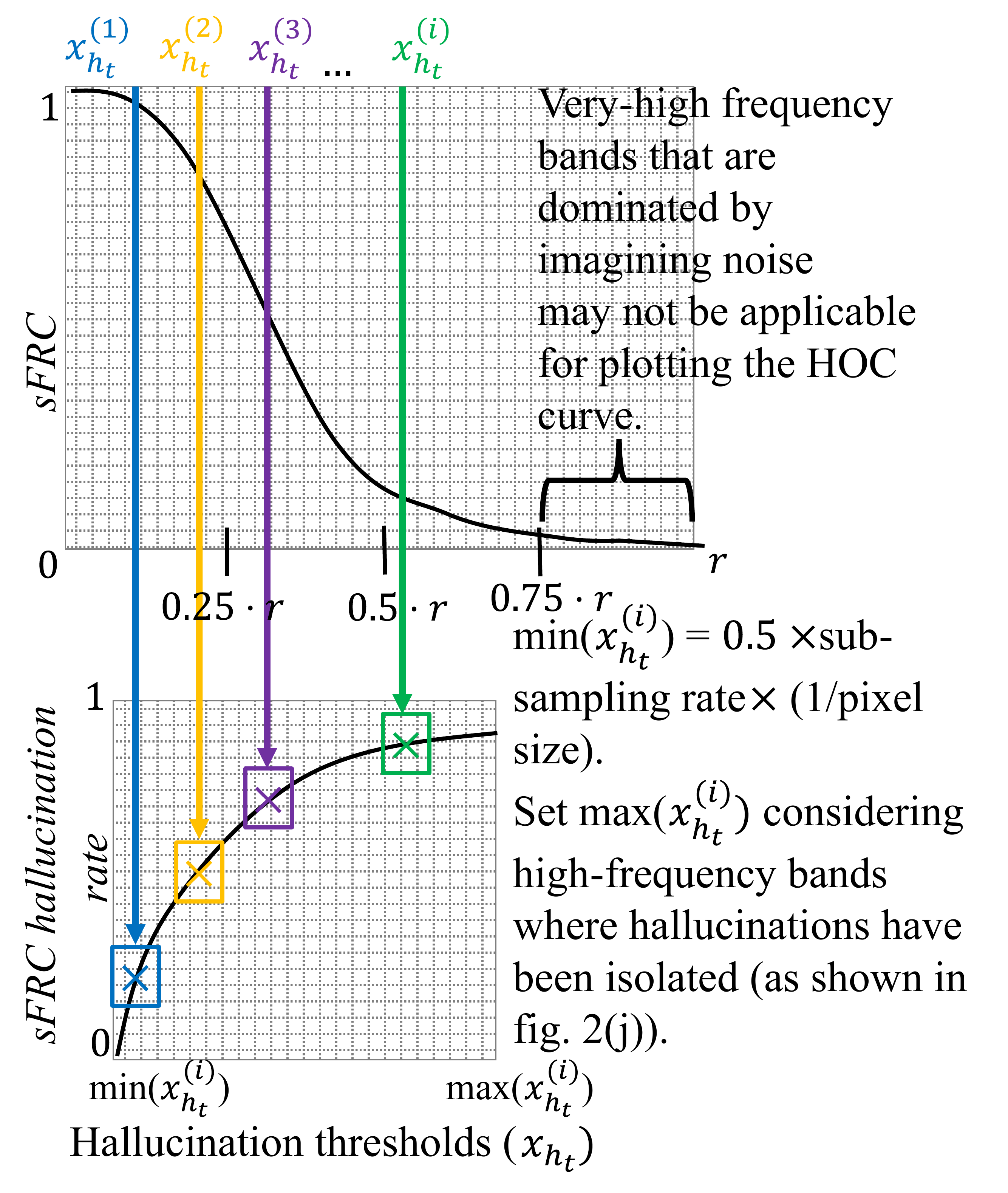}
\caption{\tip{A visual depiction of how our patch-wise (scanned) FRC (sFRC) plot enables estimation of the hallucination rate as a function of the sFRC's hallucination threshold ($x_{h_t}$). Concretely, the minimum $x_{h_t}$ value is set based on the sampling rate for a given restoration problem, while the maximum value is determined by considering the resolving power of the imaging modality and the high-frequency bands where confirmed hallucinations have been isolated.}
}
\label{img:hoc_illustration}
\end{figure}

\begin{figure}[!thb]
\centering
\includegraphics[width=0.65\linewidth]{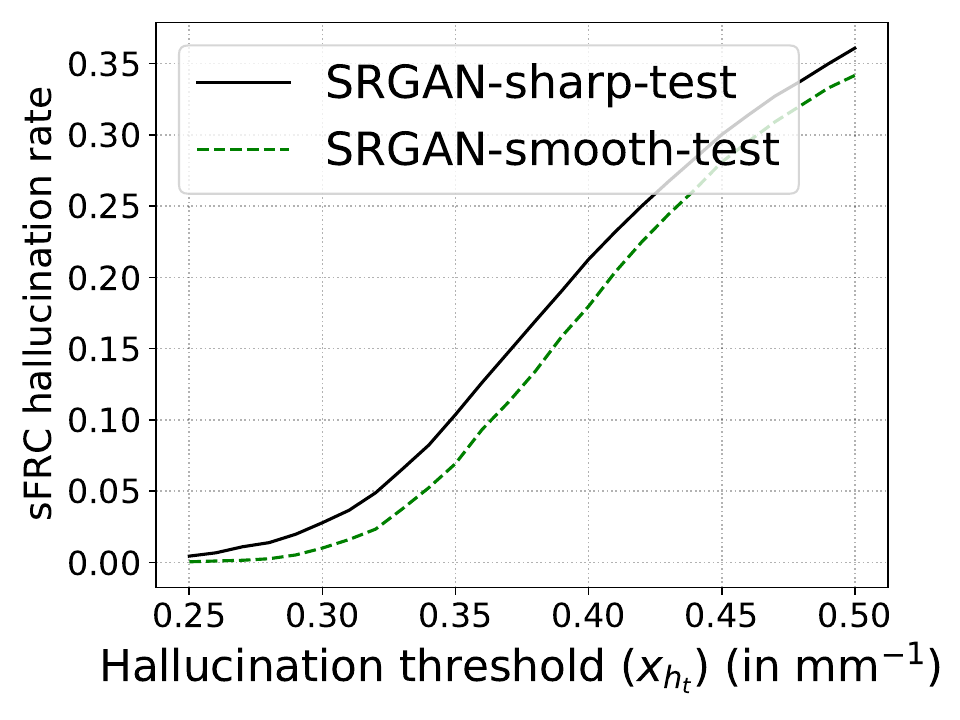}
\caption{\tip{Hallucination Operating Characteristic (HOC) curve for the SRGAN model - trained on smooth data - evaluated using smooth and sharp test sets.}}
\label{img:ct_hoc_curve}
\end{figure}

\tip{\subsection{Future work on Hallucination Operating characteristic (HOC) curve}\label{sec:hoc_future_work}
Recall that hallucination threshold ($x_{h_{t}}$) is parallel to the $y$-axis. As $x_{h_{t}}$ shifts from $0$ to Nyquist frequency ($r$), sFRC becomes more strict (i.e., more patches are labeled as hallucinated ROIs, as sFRC performs correlation analysis across very-low to very-high frequency bands). Then, \textbf{plotting sFRC hallucination rate as a function of $x_{h_{t}}$ will yield a curve similar to an ROC (termed as Hallucination Operating characteristic (HOC) curve)}. For a given  restoration problem, one can define a meaningful range for $x_{h_t}$ for a HOC curve as follows: 
\begin{itemize}
    \item Set $\min(x^{(i)}_{h_t}) = \text{subsampling rate} \times \frac{1}{2\times\text{pixel size}}$; 
    \item Set $\max(x^{(i)}_{h_t})$ based on the resolving power of the imaging modality (\tobereviewd{fig.\,S7, SI}) and the high-frequency bands where hallucinations have been confirmed (fig.~\ref{img:fourier_decomposition}(j)).
\end{itemize}

An illustration of an HOC curve is provided in fig.\,\ref{img:hoc_illustration}. For the CT super-resolution problem with downsampling by a factor of 4, a reasonable range for $x_{h_t}$ in this plot is $[0.25,0.5]$, since approximately 25\% of the Fourier information is expected to remain intact, and the hallucinations are isolated within mid-frequency bands (extending up to $0.5\cdot r$, as shown in fig.\,\ref{img:fourier_decomposition}(j)). Consistent with this expectation, fig.\,\ref{img:ct_hoc_curve} shows that the hallucination rate approaches zero as $x_{h_t} \to 0.25$.}
\tip{
As $x_{h_t}$ increases, the hallucination rate also increases, indicating that the SRGAN did not accurately restore the mid-frequency bands (similar to the hallucinations shown in fig.\,\ref{img:fourier_decomposition}(j)). Finally, the HOC curve for the smooth test set is lower than that for the sharp test set, consistent with the observations in fig.\,\ref{img:sh_sm_pic} and the empirically known fact that deep-learning models exhibit degraded performance on out-of-distribution test data compared with in-distribution data.

Analogous to the Area Under the ROC (AU-ROC) used for diagnostic comparison in clinical science, the Area Under the HOC (AU-HOC) \infoclear{may} be computed to summarize hallucinatory performance between AI-restored and reference methods across different $x_{h_t}$ thresholds. Uncertainty in the HOC \infoclear{may} be estimated by varying important sFRC parameters---such as patch size or bin size---when calculating the patch-wise Fourier Ring Correlation. However, it will involve significant future effort to truly validate the number of hallucinated ROIs at different $x_{h_t}$ values. 

A brute-force approach to clinically validate the HOC curve \infoclear{could} encompass ground truth ROIs and hallucinated ROIs at different spatial frequencies. However, acquiring such a reference dataset may be infeasible, particularly because it is not as straightforward as setting a true signal (such as a lung nodule present) and its corresponding false signal (lung nodule not present) for a binary signal detection task that is commonly employed for ROC analysis in different diagnostic tests. This is further aggravated by the fact that the hallucinatory output (in terms of the impacted signals/pathologies/organs, occurrence, and downstream clinical repercussions) may change with different AI restoration techniques. Although sFRC \infoclear{may} be employed to detect candidate hallucinated ROIs at different spatial frequencies, thereby building a library of hallucinated ROIs for a given AI method. It also involves a cumbersome effort of further labeling true positives, false positives, true negatives, and false negatives of sFRC outputs using an multireader multicase (MRMC) \cite{obuchowski_mrmc_paper} study.

A more practical yet robust method to validate HOC may be to use a dataset with ROI labels for diagnostically important lesions or pathologies across multiple organs (e.g., DeepLesion \cite{deeplesion} or fastMRI+ \cite{fastmri+}). One could then perform downstream computer-aided detection (CAD) using different AI restoration methods---including transformers and diffusion models---and compare the resulting HOC output against ROC performance. If a given AI restoration method exhibits hallucinatory behavior, we \tip{could} expect its performance to degrade in downstream CAD evaluation. Consequently, HOC could be shown to correlate with ROC-based CAD performance across multiple organs and a range of diagnostic signals spanning low- to high-frequency bands. However, this approach would involve substantial effort to carefully construct an appropriate study design for such an HOC versus downstream CAD comparison. Refining sFRC into a threshold-free HOC analysis—along with incorporating uncertainty estimates, validating the HOC curve using downstream task performance, and analyzing hallucinatory behavior across different restoration methods—will be an important direction for future work.
}
\section{Conclusion}
\label{sec:conclusion}
AI-based image restoration algorithms have shown success in computer vision and are increasingly being proposed for biomedical imaging. In recent years, there has been rapid growth in FDA submissions for AI-based image reconstruction and post-processing software as medical devices, with premarket authorizations increasing by approximately 25\% annually since 2018 [\href{https://www.fda.gov/medical-devices/software-medical-device-samd/artificial-intelligence-and-machine-learning-aiml-enabled-medical-devices?utm_medium=email&utm_source=govdelivery}{FDA AI/ML-enabled devices}]. These applications span dose reduction, deblurring, spatial resolution enhancement, domain transfer, and \tip{accelerated data acquisition} across multiple imaging modalities. However, due to AI restoration methods' possibility of hallucinatory behavior, there is a risk of misdiagnosis if such methods are used without thorough analysis to ensure patient safety. 

\infoclear{The sFRC approach described in this paper enables detection of hallucinations in medical image restoration problems in an effective, objective and automatic manner}. sFRC is effective in that it accommodates multiple imaging modalities, subsampling scenarios and non-linear methods, including post-processing or reconstruction techniques based on AI/ML and regularization. It is objective because it can incorporate predefined criteria—derived from imaging theory or labeled by experts—to define hallucinations and to automatically apply them on test datasets to detect new hallucinations. AI developers may use this metric to characterize hallucinations in AI-restored images in a bench testing manner. 

\bibliographystyle{IEEEtran}
\bibliography{mreport}
\end{document}


\title{sFRC for assessing hallucinations in medical image restoration \protect\\ (Supplemental Information)}
\author{Prabhat Kc, Rongping Zeng, Nirmal Soni, and Aldo Badano
\thanks{\tip{\textbf{DISCLAIMER:} This article reflects the views of the authors and does not represent the views or policy of the U.S. Food and Drug Administration, the Department of Health and Human Services, or the U.S. Government.  The mention of commercial products, their sources, or their use in connection with material reported herein is not to be construed as either an actual or implied endorsement of such products by the Department of Health and Human Services.}}
}
\maketitle
\section{Tuning parameters for an sFRC analysis}
\begin{figure*}
\centering
\includegraphics[width=\textwidth]{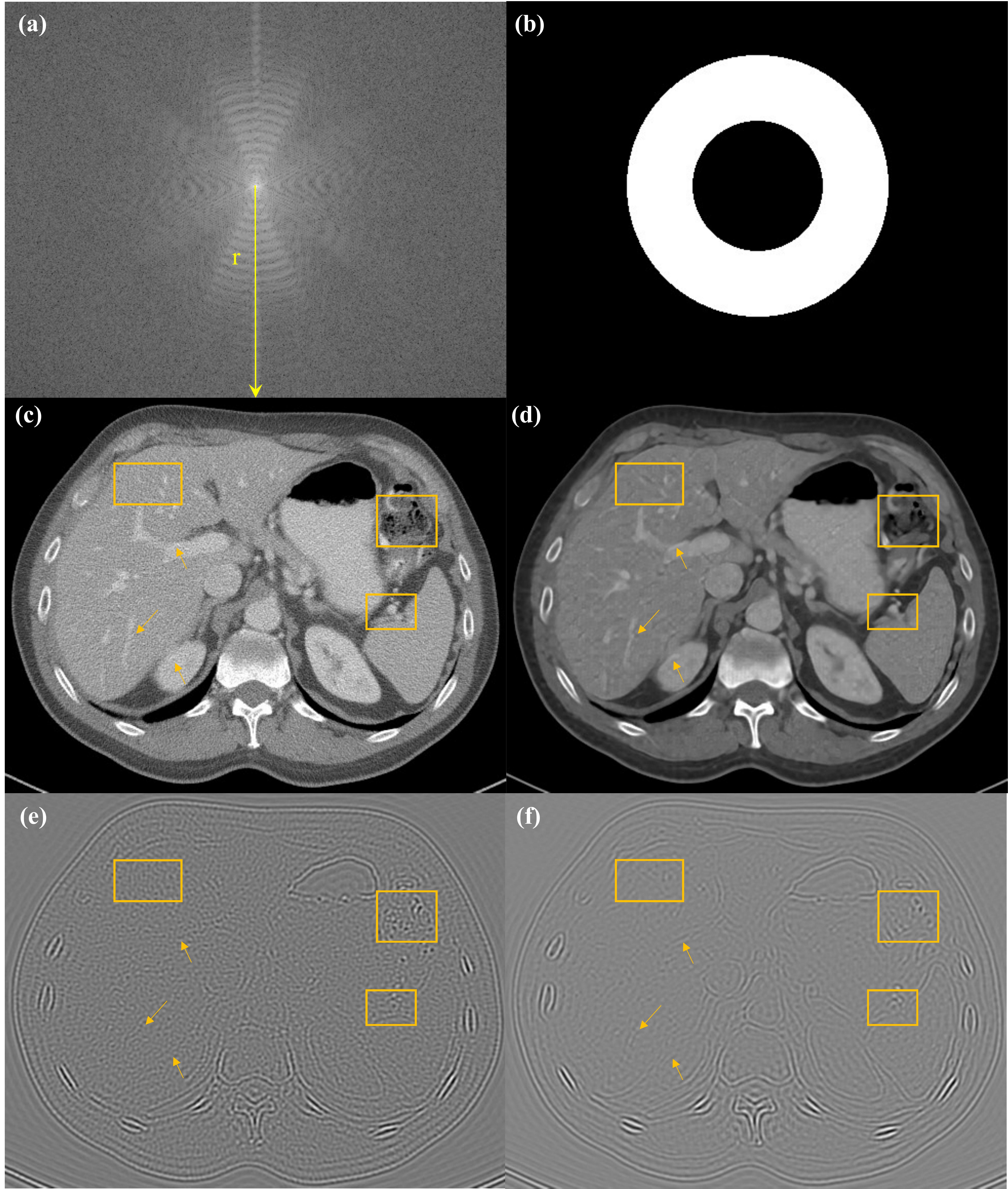}
\caption{\revfresponse{An illustration of user-defined hallucinated ROIs (as bounding boxes) and markers (as arrows) annotated in the SRGAN upsampled output in (d) by comparing the image against its normal-resolution reference in (c). The comparison was assisted using the complementary components (e, f) of the two CT images (c, d) retrieved by convolving a bandpass filter ($0.25\cdot r$ to $0.5\cdot r$) in (b) to (c, d), respectively. Prior information that the normal-resolution CT reference and low-resolution test data correspond to contrast-enhanced abdominal CT examinations in the portal-venous phase of enhancement was also utilized when comparing. The normal-resolution CT scan in (c) corresponds to the axial scan 000039.IMA of patient L$067$ (acquired at a $3$mm slice thickness using a sharp kernel). This normal-resolution CT scan was provided in the LDGC repository \cite{ldct_data_2016}. The display window for CT images in (c, d) is (W:$700$ L:$50$).}}
\label{img:039_annotation}
\end{figure*}

\begin{figure*}
\centering
\includegraphics[width=\textwidth]{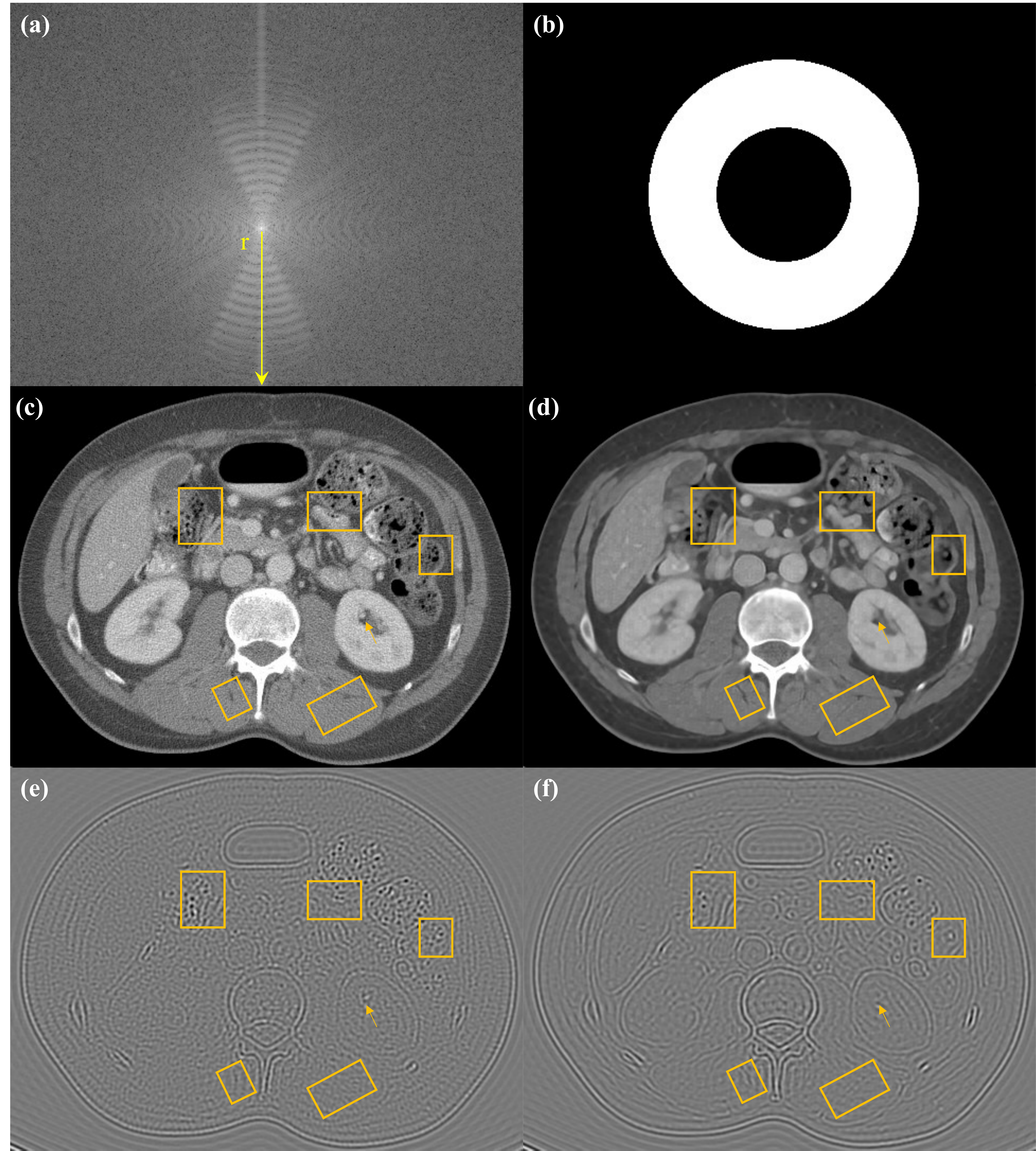}
\caption{\revfresponse{An illustration of user-defined hallucinated ROIs (as bounding boxes) and a marker (as an arrow) annotated in the SRGAN upsampled output in (d) by comparing the image against its normal-resolution reference in (c). The comparison was assisted using the complementary components (e, f) of the two CT images (c, d) retrieved by convolving a bandpass filter ($0.25\cdot r$ to $0.5\cdot r$) in (b) to (c, d), respectively. Prior information that the normal-resolution CT reference and low-resolution test data correspond to contrast-enhanced abdominal CT examinations in the portal-venous phase of enhancement was also utilized when comparing. The normal-resolution CT scan in (c) corresponds to the axial scan 000069.IMA of patient L$067$  (acquired at a $3$mm slice thickness using a sharp kernel). This normal-resolution CT scan was provided in the LDGC repository \cite{ldct_data_2016}. The display window for CT images in (c, d) is (W:$700$ L:$50$).}}
\label{img:069_annotation}
\end{figure*}

\begin{figure*}
\centering
\includegraphics[width=\textwidth]{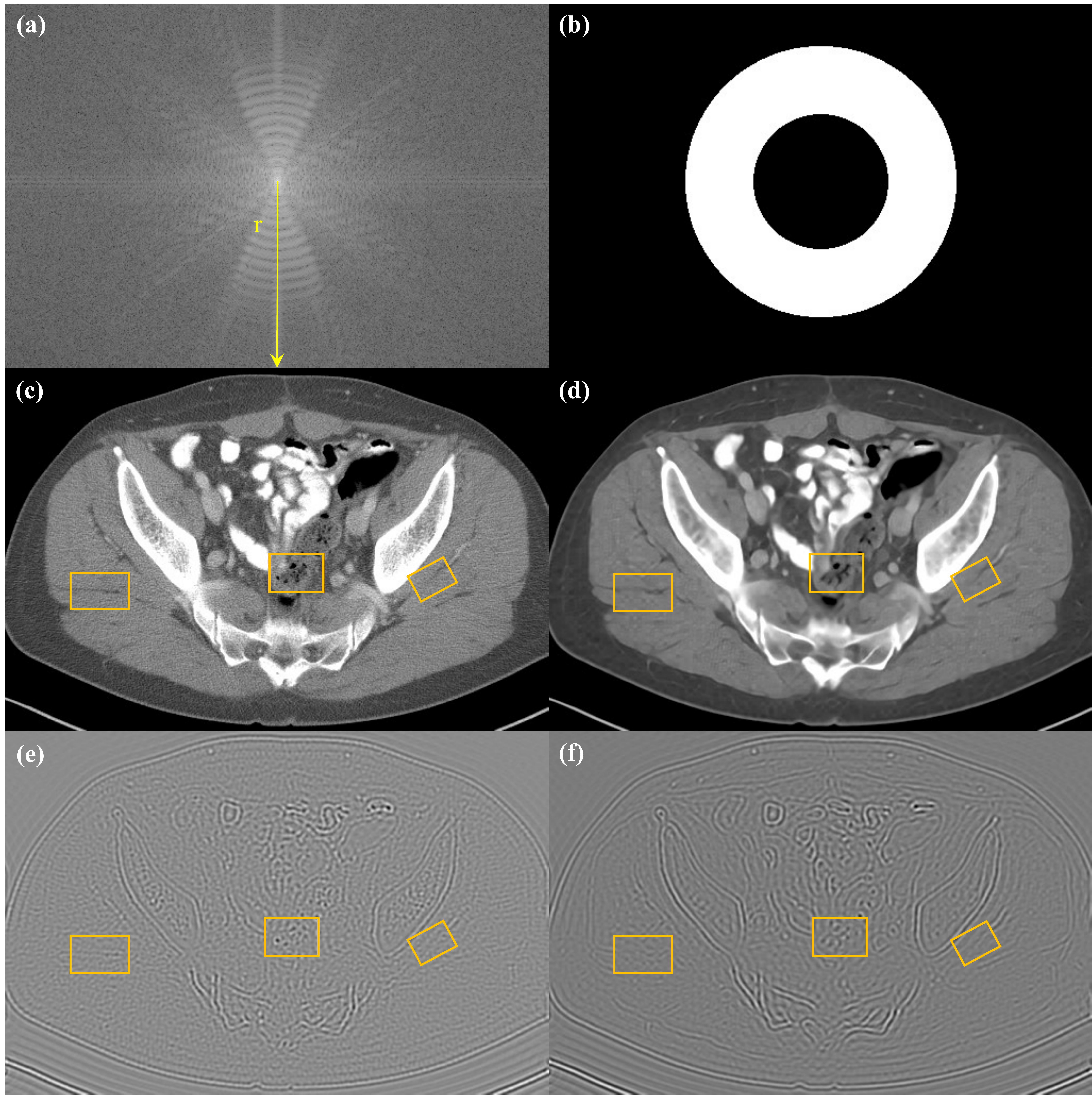}
\caption{\revfresponse{An illustration of user-defined hallucinated ROIs (as bounding boxes) annotated in the SRGAN upsampled output in (d) by comparing the image against its normal-resolution reference in (c). The comparison was assisted using the complementary components (e, f) of the two CT images (c, d) retrieved by convolving a bandpass filter ($0.25\cdot r$ to $0.5\cdot r$) in (b) to (c, d), respectively. Prior information that the normal-resolution CT reference and low-resolution test data correspond to contrast-enhanced abdominal CT examinations in the portal-venous phase of enhancement was also utilized when comparing. The normal-resolution CT scan in (c) corresponds to the axial scan 000149.IMA of patient L$067$ (acquired at a $3$mm slice thickness using a sharp kernel). This normal-resolution CT scan was provided in the LDGC repository \cite{ldct_data_2016}. The display window for CT images in (c, d) is (W:$700$ L:$50$).}}
\label{img:149_annotation}
\end{figure*}

\begin{figure*}
\centering
\includegraphics[width=\textwidth]{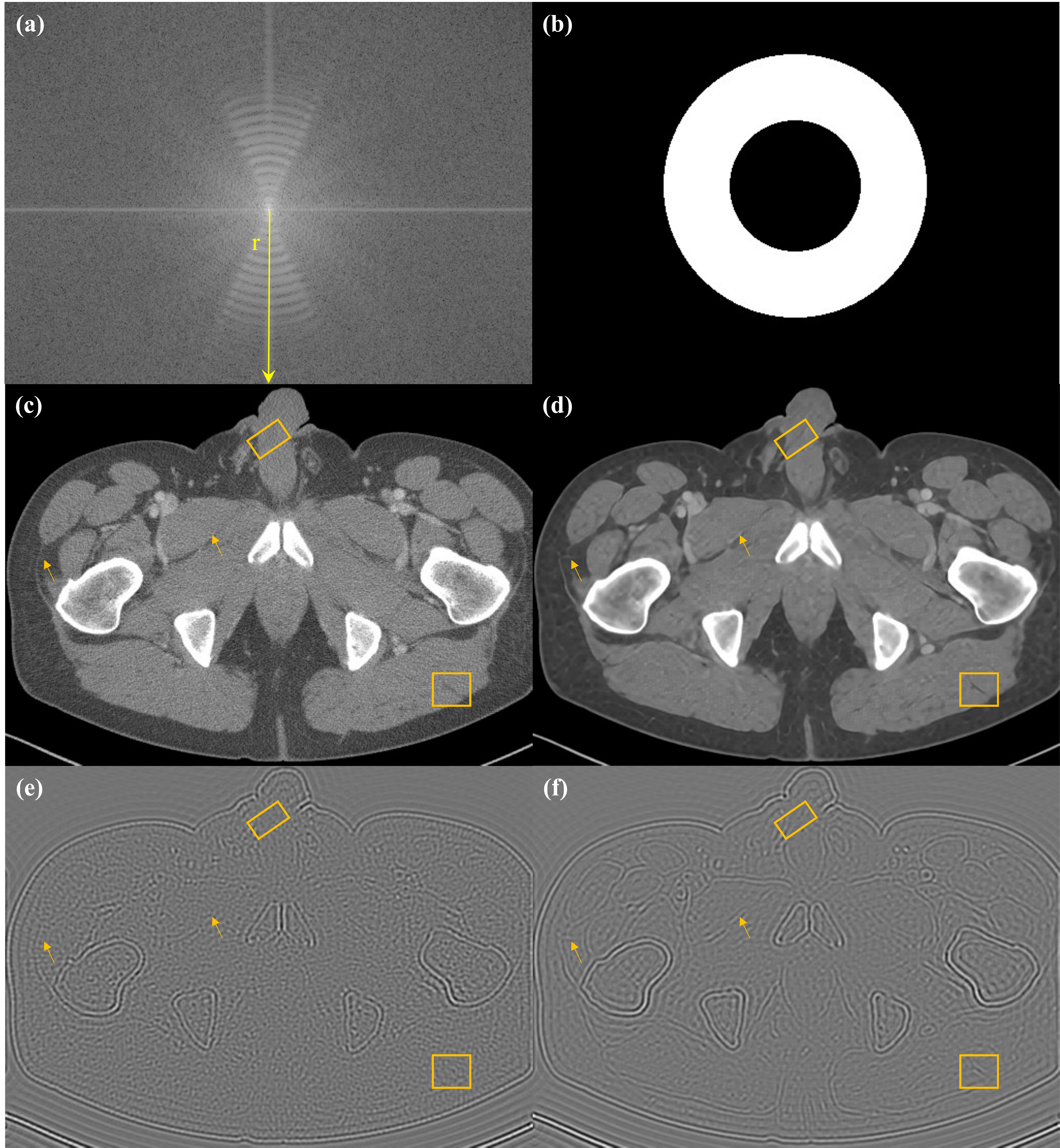}
\caption{\revfresponse{An illustration of user-defined hallucinated ROIs (as bounding boxes) and markers (as arrows) annotated in the SRGAN upsampled output in (d) by comparing the image against its normal-resolution reference in (c). The comparison was assisted using the complementary components (e, f) of the two CT images (c, d) retrieved by convolving a bandpass filter ($0.25\cdot r$ to $0.5\cdot r$) in (b) to (c, d), respectively. Prior information that the normal-resolution CT reference and low-resolution test data correspond to contrast-enhanced abdominal CT examinations in the portal-venous phase of enhancement was also utilized when comparing. The normal-resolution CT scan in (c) corresponds to the axial scan 000189.IMA of patient L$067$ (acquired at a $3$mm slice thickness using a sharp kernel). This normal-resolution CT scan was provided in the LDGC repository \cite{ldct_data_2016}. The display window for CT images in (c, d) is (W:$700$ L:$50$).}}
\label{img:189_annotation}
\end{figure*}

\begin{figure*}
\centering
\includegraphics[width=\textwidth]{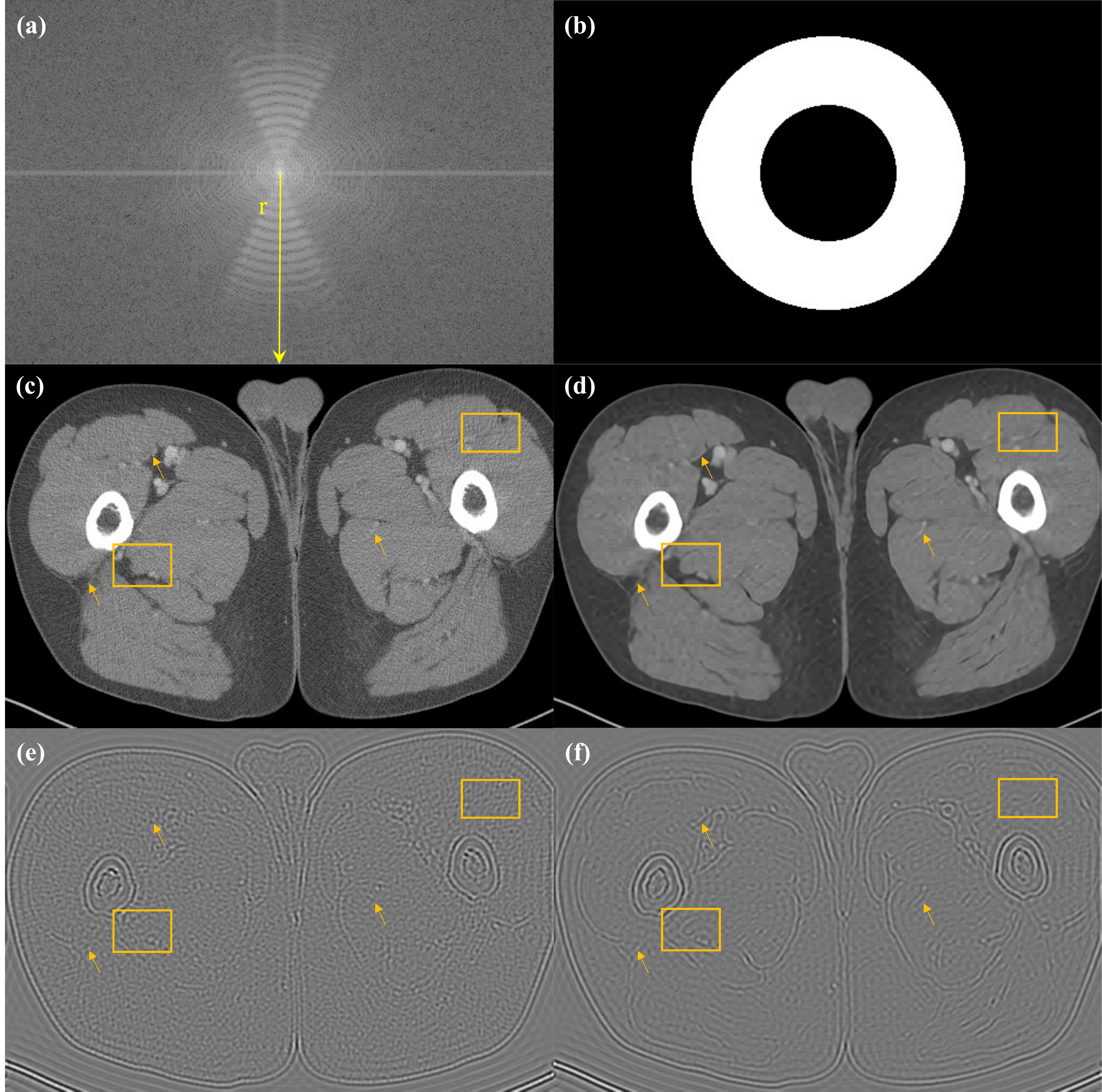}
\caption{\revfresponse{An illustration of user-defined hallucinated ROIs (as bounding boxes) and markers (as arrows) annotated in the SRGAN upsampled output in (d) by comparing the image against its normal-resolution reference in (c). The comparison was assisted using the complementary components (e, f) of the two CT images (c, d) retrieved by convolving a bandpass filter ($0.25\cdot r$ to $0.5\cdot r$) in (b) to (c, d), respectively. Prior information that the normal-resolution CT reference and low-resolution test data correspond to contrast-enhanced abdominal CT examinations in the portal-venous phase of enhancement was also utilized when comparing. The normal-resolution CT scan in (c) corresponds to the axial scan 000217.IMA of patient L$067$ (acquired at a $3$mm slice thickness using a sharp kernel). This normal-resolution CT scan was provided in the LDGC repository \cite{ldct_data_2016}. The display window for CT images in (c, d) is (W:$700$ L:$50$).}}
\label{img:217_annotation}
\end{figure*}

For the CT super-resolution problem – considered in section \tobereviewd{IV-B} in the main paper – the normal-resolution CT images consisted of images from contrast-enhanced abdominal CT examinations acquired during the portal-venous phase of the enhancement. These CT images were collected using patients with positive cases relative to the presence of hepatic metastasis identified by histology (surgery or biopsy), or progression on serial cross-sectional exams, or regression on serial exams (with treatment) \cite{ldct_data_2016}. A consultation with our internal medical officer on the SRGAN outputs (in figs.\,\ref{img:039_annotation} through \ref{img:217_annotation}) used to tune our sFRC parameters revealed that the Super-Resolution Generative Adversarial Network (SRGAN) outputs exhibited hallucinations relative to plaque-like structures as compared to their normal-resolution counterpart. Also, a thorough review of the literature revealed hallucinations such as the addition of vessels \cite{fast_mri_2020_result}, changing of the shape of pathological signals \cite{mri_fn_annotation_paper}, removal of the clinically important signals \cite{gan_domain_transfer_hallu}, and perturbation of local structures \cite{antun_instabilities}. 

Given these two pieces of information – from the medical officer and literature survey – we meticulously annotated
hallucinations in the tuning set of sFRC (i.e., considering hallucinations related to plaque-like structures in the blood vessels, indentations and perturbation of small organelles). This annotation process was assisted by using image components of the CT pairs (reference and SRGAN) retrieved by convolving a bandpass filter ($0.25\cdot r \text{ to } 0.5\cdot r$) to each of them (as depicted in subplots (e, f) in figs.\,\ref{img:039_annotation} through \ref{img:217_annotation})) and was completed using ImageJ software. Note that beyond this set of hallucinations, \textbf{we intentionally did not annotate other hallucinations/artifacts in the tuning images.} This is primarily because we wanted to test the robustness of sFRC in detecting a range of hallucinations once sFRC is calibrated/tuned using clinical information that was known to us and had been thoroughly validated using literature surveys and a medical officer. To align the clinical condition of the patient cases and the acquisition protocol used in this study, we did not use CT scans from the lung, heart, and chest region of the testing patient L$067$. We used CT scans spanning from the upper abdomen to the lower leg of patient L$067$ for the sFRC analysis.

\begin{figure*}
\centering
\includegraphics[width=1.0\textwidth]{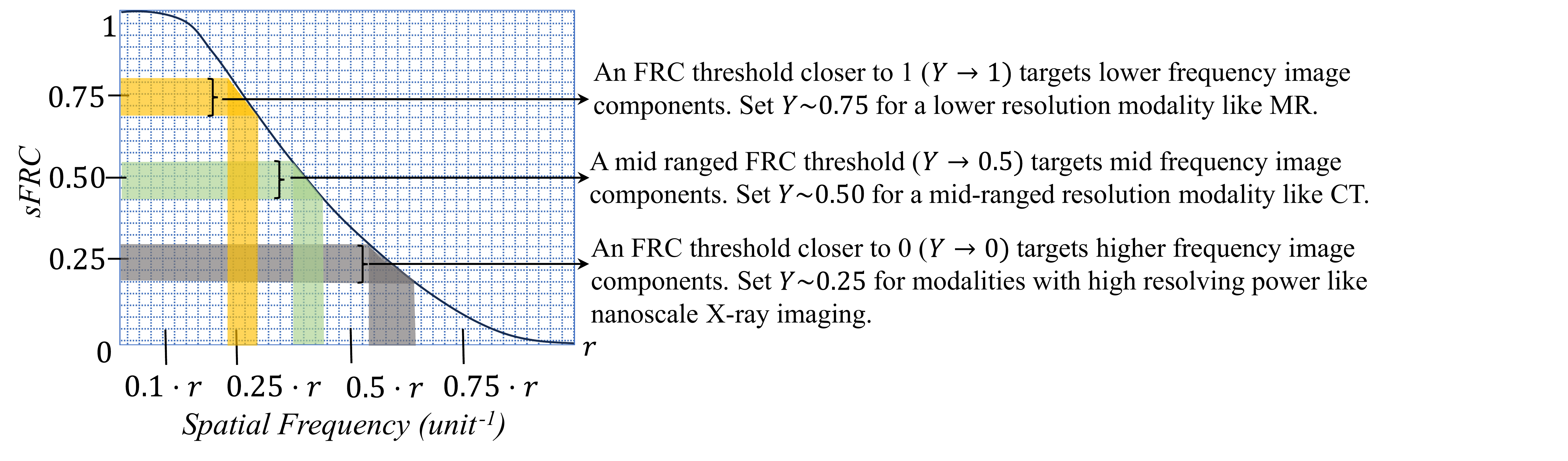}
\caption{\tip{A visual depiction of how the FRC threshold ($Y$) targets different spatial frequencies in sFRC analysis between deep learning and reference images, with Y set in the range $[0, 1]$.}}
\label{img:frc_threshold_targeting}
\end{figure*}

We then proceeded to tune the sFRC parameters (i.e., \textbf{patch size, FRC threshold, and hallucination threshold}). We do not expect a high correlation along rings greater than $0.5$ times the Nyquist frequency ($r$), even for faithfully restored patches with enough features. This is primarily due to:
\tip{
\begin{itemize}
    \item the patch-wise nature of our analysis, 
    \item those very high-frequency rings predominantly contain edges (or very fine details), and
    \item  pairs of patches for the analysis are restored using two different methods (rather than realizations).
\end{itemize}
}
Next, due to the use of the data prior to resolving the CT downsampling problem by factor $4$ (i.e., loss of information by $75\%$), we expect rings up to $0.25$ of $r$ (or very low- to low-frequency-based image components) to be properly restored (as shown in fig.\,$2$ in the main paper). This expectation was also validated by our ability to discern the hallucinated ROIs and markers in the image components corresponding to the rings - $0.25\cdot r \text{ to } 0.5\cdot r$ - as shown in subplots (e, f) in figs.\,\ref{img:039_annotation} through \ref{img:217_annotation}. We set $0.5$ as the FRC threshold (as shown in fig.\,$3$ in the main paper), as it would target the FRC curve drop-off of the hallucinated features along the rings, $0.25\cdot r \text{ to } 0.5\cdot r$. Note that setting the FRC threshold closer to $0$ (i.e., $Y\rightarrow0$) would result in a comparison along the tapering region of the FRC curve that is oscillatory in nature. This would impede the hallucinated threshold, $x_{h_{t}}$, from distinguishing the mismatch, $x_{c_{t}}$,  relative to a faithfully restored versus hallucinated patch. Similarly, if the FRC threshold is set closer to $1$, (i.e., $Y\rightarrow1$), the comparison would be limited to the very low-frequency bands (i.e., rings$\rightarrow 0 \cdot r$). \tip{This property of the FRC threshold---namely, how it targets different frequency bands depending on the chosen value (which, in turn, depends on the resolving power of the imaging system)---is visually depicted in fig.\,\ref{img:frc_threshold_targeting}.}

As previously explained in \tobereviewd{section IV-B4 in our main paper}, we found that using large patch sizes ($\geq96\times96$) decreases the chances of correctly detecting hallucination, for the imperceptible hallucinations are usually limited to small ROIs. Using a large patch size overinflates the correlation outcomes from the faithfully restored regions. Likewise, using small patches (usually, $\leq 32 \times 32$) increases the likelihood of wrongly labeling ROIs as hallucinations (i.e., an increase in false positives). We found that the sFRC analysis using $64 \times 64$ sized patches yields optimal outcomes in terms of maximizing true positives and minimizing false positives for the super-resolution problem. The nature of the imaging problems (such as missing wedge, ultra-high noise, geometric distortion, etc.) can also guide when setting patch size. Subhead \ref{sec:sfrc_on_nonfakes} provides more information on this aspect.

\begin{figure*}
\centering
\includegraphics[width=.9\textwidth]{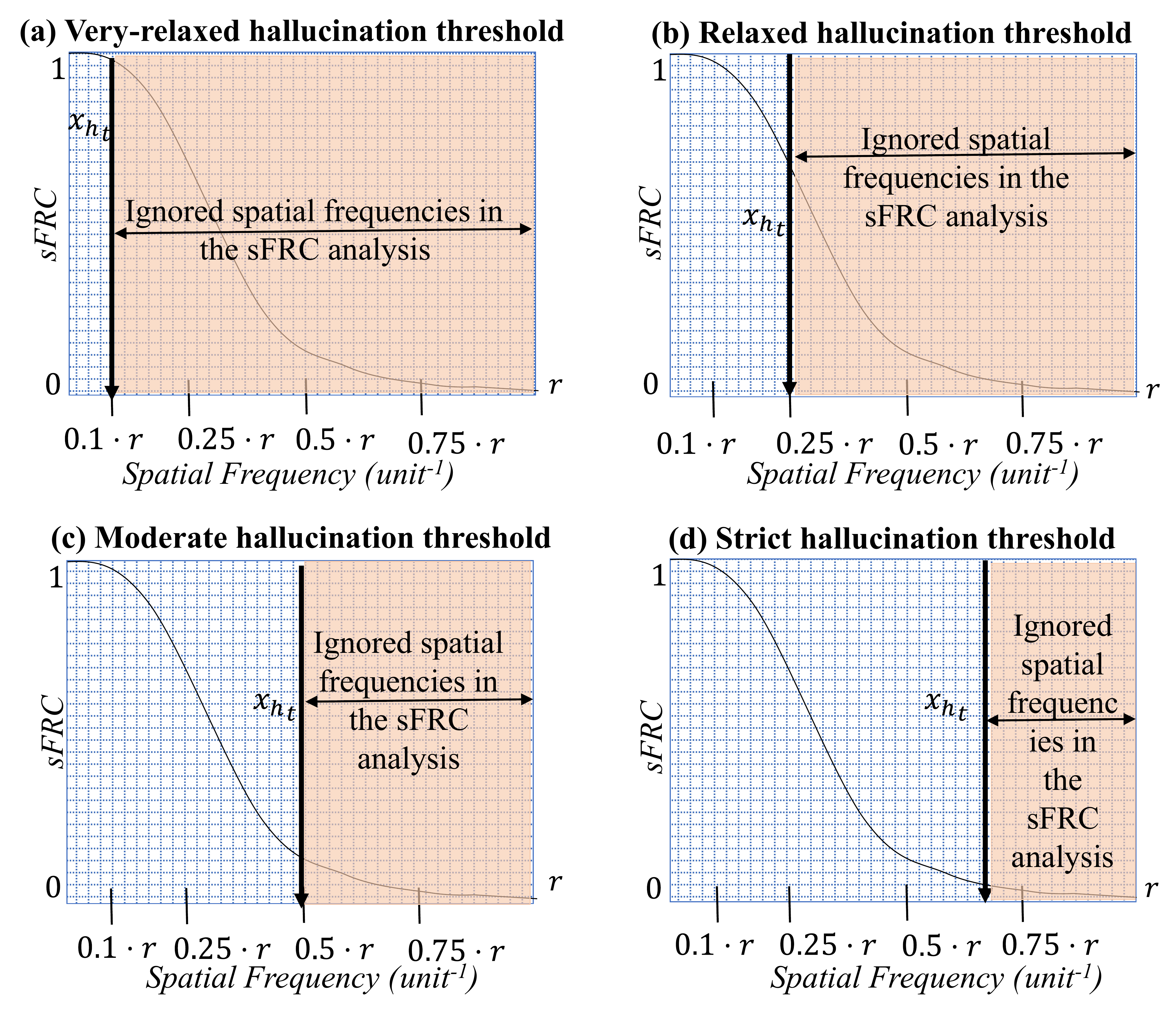}
\caption{\tip{Illustration of hallucination threshold ($x_{h_{t}}$) getting stricter (i.e., label more patches to be hallucinated) as $x_{h_{t}}$ approaches Nyquist frequency ($r$). For instance, $x_{h_{t}}$ in plot (a) is very relaxed, as sFRC will only consider very low-frequency bands (i.e., blurred image components) when performing correlation analysis between Deep learning and reference image pairs. In contrast, $x_{h_{t}}$ in plot (d) is strict, as sFRC considers image components across low- to high-frequency bands when performing the correlation analysis.}}
\label{img:xht_strictness_plot}
\end{figure*}

\begin{figure*}
\centering
\includegraphics[width=0.98\linewidth]{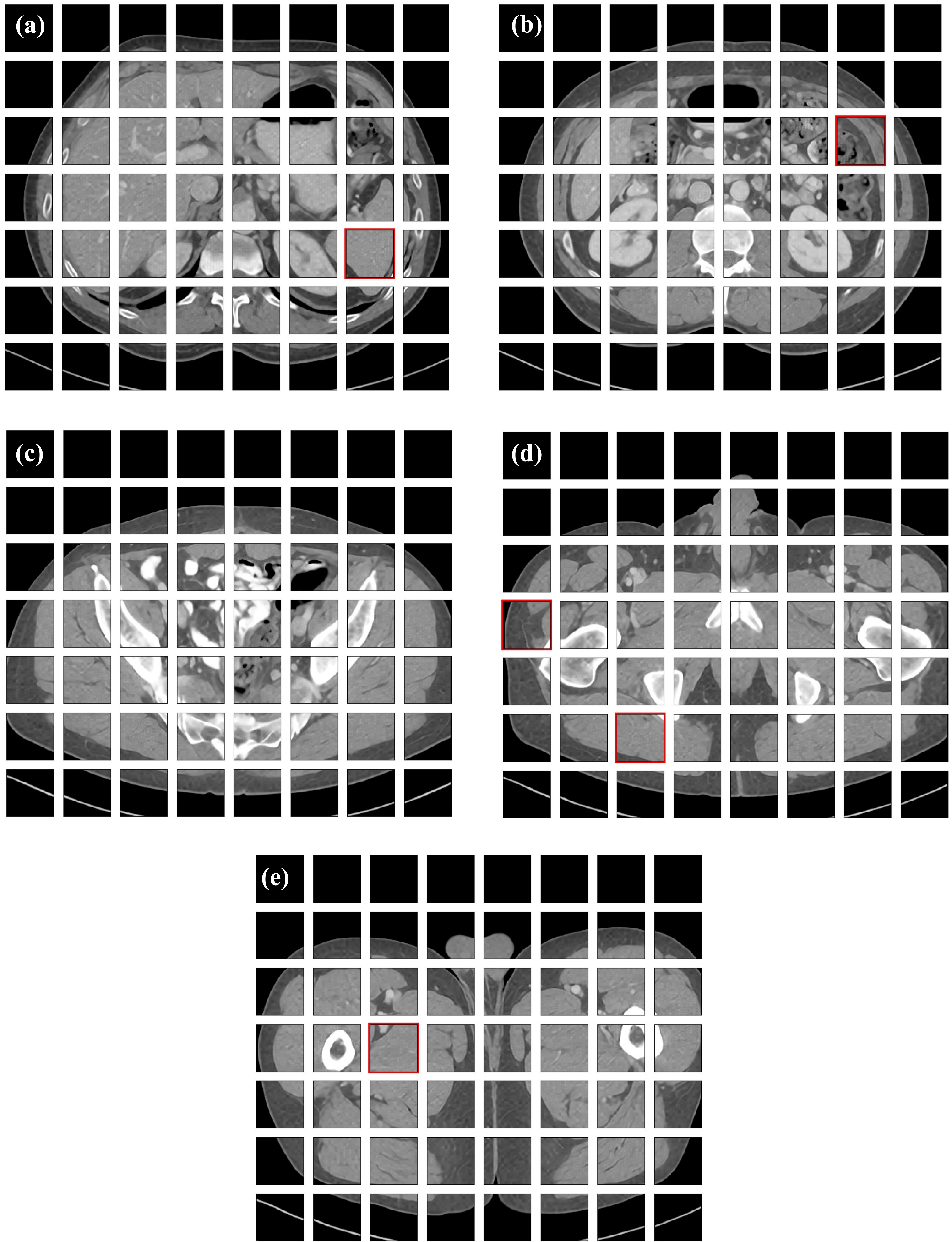}
\caption{\revfresponse{Hallucinations identified in the red bounding boxes from our sFRC analysis in the SRGAN upsampled outputs. The five SRGAN upsampled CT images correspond to images used to tune the sFRC parameters. The sFRC parameters comprised of $64\times64$ as the patch size, $0.5$ as the FRC threshold, and $\mathbf{0.25}$ as the hallucination threshold ($x_{h_{t}}$). At $0.25$ $x_{h_{t}}$, almost all the hallucinated ROIs - labeled in figs.\,\ref{img:039_annotation} through \ref{img:217_annotation} - were missed by the sFRC. The display window for all images is (W:$700$ L:$50$).}}
\label{img:sfrc_on_ct_sh_tuning_ht_0.25}
\end{figure*}

\begin{figure*}
\centering
\includegraphics[width=0.98\linewidth]{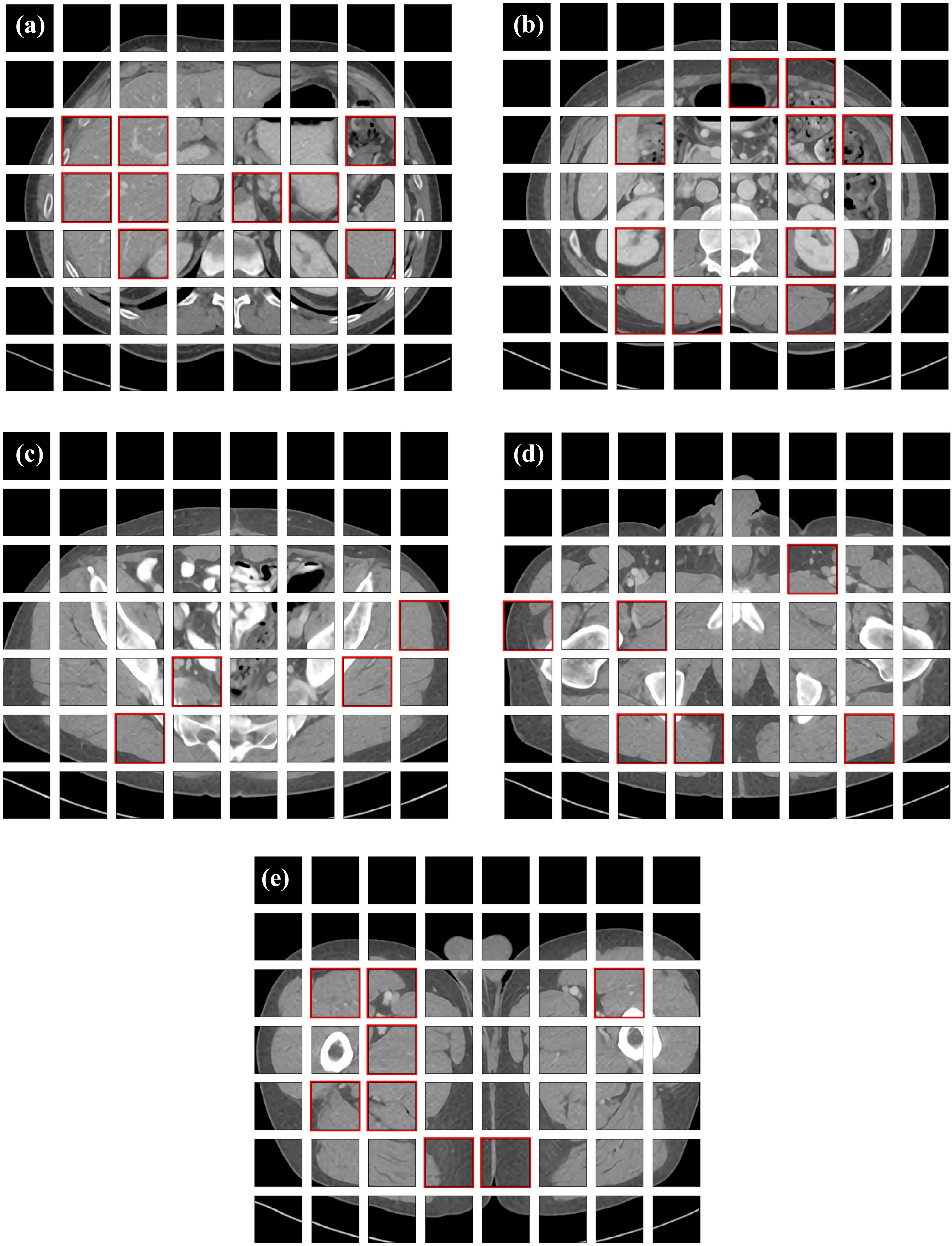}
\caption{\revfresponse{Hallucinations identified in the red bounding boxes from our sFRC analysis in the SRGAN upsampled outputs. The five SRGAN upsampled CT images correspond to images used to tune the sFRC parameters. The sFRC parameters comprised of $64\times64$ as the patch size, $0.5$ as the FRC threshold, and $\mathbf{0.35}$ as the hallucination threshold ($x_{h_{t}}$).  At $0.35$ $x_{h_{t}}$, almost all the hallucinated ROIs - labeled in figs.\,\ref{img:039_annotation} through \ref{img:217_annotation} - were captured by the sFRC. The display window for all images is (W:$700$ L:$50$).}}
\label{img:sfrc_on_ct_sh_tuning_ht_0.35}
\end{figure*}

\tip{One may directly extract all patches annotated as hallucinated and use \tobereviewd{eq.\,(2)} in the main paper to set $x_{h_t}$. Alternatively, $x_{h_t}$ may be initialized to correspond to the sampling rate and then sequentially increased until all hallucinated feature types are detected by sFRC.} To show this, we began by setting $x_{h_{t}}$ as $0.25 \cdot r (\approx 0.25\times\frac{1}{2}\times\frac{1}{0.48})$  mm$^{-1}$ because we downsampled the CT data by a factor of $4$. At $0.25\ x_{h_{t}}$, most of the labeled hallucinations in fig.\,\ref{img:039_annotation} through \ref{img:217_annotation} were missed by the sFRC (as shown in fig.\,\ref{img:sfrc_on_ct_sh_tuning_ht_0.25}). Then, $x_{h_{t}}$ was sequentially increased at a stepsize of $0.01$. At $0.35\ x_{h_{t}}$, almost all the hallucinations labeled in figs \,\ref{img:039_annotation} through \ref{img:217_annotation} were captured by our sFRC-based bounding boxes (as shown in fig.\,\ref{img:sfrc_on_ct_sh_tuning_ht_0.35}). The $k_{x}= x_{h_{t}}$, is a vertical line through the frequency value. In the sFRC analysis, this value plays a critical role in setting the cutoff between a faithful versus a hallucinated restoration for a given patch. More concretely, any mismatch (i.e., $x_{c_{t}}$ as an $x$-coordinate corresponding to the intersection of the FRC curve and FRC threshold) to the right of this vertical line, $k_{x} = x_{h_{t}}$, is not labeled as a hallucination. But a mismatch to the left of this line will be labeled as a hallucination. In essence, any mismatch from the Nyquist frequency (i.e., $r$) to $k_{x}=x_{h_{t}}$ is tolerated. The closer the value $k_{x}$ is to $r$, the lesser the sFRC's tolerance for passing uncorrelated patches as non-hallucinations (i.e., sFRC will label a higher number of patches as hallucinations by considering a correlation between patches at high-frequency zones, rings$\rightarrow r$). \tip{This property of $x_{h_{t}}$ is visually illustrated in fig.\,\ref{img:xht_strictness_plot}.}

We relaxed the hallucination threshold from $0.35$ to $0.33$ before applying sFRC to the testing data. This is primarily because even at $0.33\ x_{h_{t}}$, sFRC was able to detect the hallucinations related to plaques, perturbations, and indentations, as shown in fig.\,\ref{img:sfrc_on_ct_sh_tuning_ht_0.33} below. We wanted to calibrate sFRC using a handful of thoroughly validated hallucinations. Then, in the testing phase, we wanted to check sFRC's robustness to reapply its tuned knowledge on the spatial frequency-based decay in the FRC curve due to a particular type of hallucination to the rest of the testing datasets where the profile of hallucinations could be quite drastically different in terms of intensity, shape, and size. Subsequently, we found that sFRC was robust in accurately detecting hallucinations relative to unwarranted folding (overfitting) of intestinal as contiguous loops in \tobereviewd{fig.\,6 in the main paper}. We also found that the contrast hallucinations (annotated as the plaque artifact in our tuning set) were sufficient to detect overattenuation along the boundaries of structures (as shown using the \tobereviewd{arrow (1) in fig.\,8 in the main paper}). sFRC was robust in detecting a new type of hallucination as shown using \tobereviewd{arrow (2) in fig.\,8} in the main paper (whereby SRGAN underfitted low-intensity based fatty attenuation as air). As noted in the results \tobereviewd{section in V-A2 in the main paper}, the annotation from the 5 CT tuning images was robust enough to detect a wide range of hallucinations such as smoothing, in-homogeneity, tiny structural changes, removal of subtle features, distortion of small organelles, addition of minute indentation-/blood vessel-/plaque-like structures, coalescing of small organelles, unwarranted foldings in the testing stack of patient L$067$.  Even in the MRI subsampling case study (\tobereviewd{in section IV-C in the main paper}), we found that our sFRC, mostly tuned using hallucination maps corresponding to sulci, was robust to pick  hallucinations related to contrast migration, as shown in \tobereviewd{fig.\,9(f)} in the main paper. sFRC also picked the removal of a dark signal in \tobereviewd{fig.\,10(b)} in the main paper.

Overall, using different AI-based restoration techniques, sFRC could be used to build an atlas of hallucinations for a given imaging problem. This could be achieved by first sequestering a ground truth (or clinically annotated) data lakes with labels corresponding to pathology \cite{fastmri_plus_dataset} and lesion \cite{deep_lesion_dataset} across different body parts at the fully sampled mode. On the one arm, \tip{multireader multicase (MRMC) \cite{obuchowski_mrmc_paper}} with human observers-based grading task could be used to set objective grading on faithfully restored and hallucinated ROIs relative to an AI algorithm-based undersampled restoration. On the second arm, sFRC could be applied to AI-based outputs. Then, ROIs detected as hallucinations from sFRC at different $x_{h_t}$ could be compared to those from the readers in the MRMC study to understand the nature of sFRC's sensitivity- and specificity-based performance. \tip{This would also allow us to estimate the \textbf{hallucination operating characteristic (HOC)} curve (\tobereviewd{see Section VI-F in the main paper}). In the main paper, we discuss validating the HOC curve by comparing it against ROC performance from downstream CAD evaluations of AI-restored scans across multiple organs and clinically relevant signals. Nevertheless, substantial additional research is warranted to appropriately establish study designs for such MRMC-based grading or downstream CAD-based evaluations that incorporate diverse clinically important signals and pathologies (beyond simple binary signal detection)}. 
\section{sFRC on conventional artifacts (non-hallucinations)}
\label{sec:sfrc_on_nonfakes}
In order to understand the outcomes of sFRC on conventional imaging artifacts (that are considered non-hallucinations in the main paper), we simulated missing wedge, geometric distortion, blur, and noise artifacts in the following manner: 

\subsection{Method}
\begin{figure*}
\centering
\includegraphics[width=0.9\textwidth]{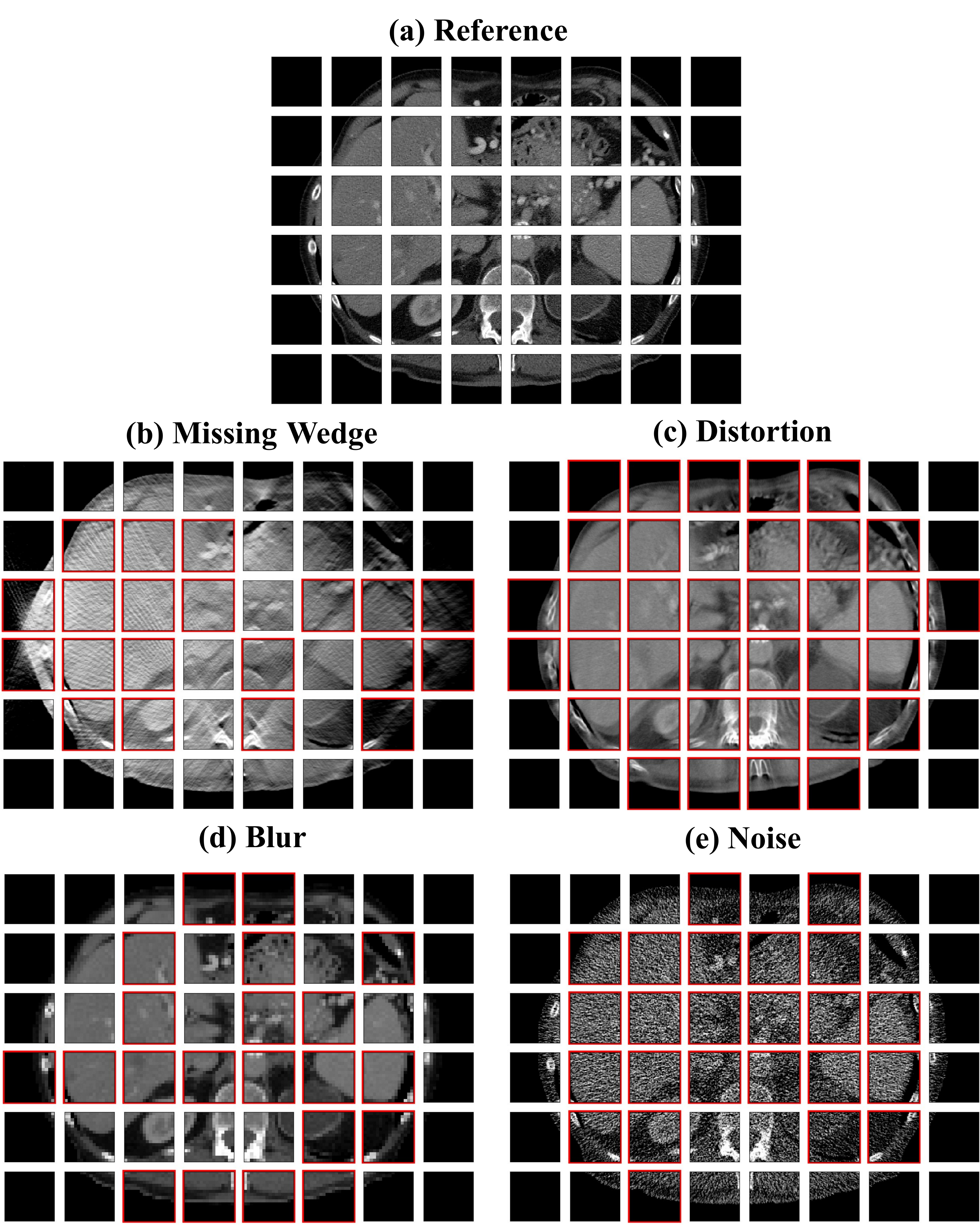}
\caption{\revfresponse{An illustration of a reference CT image in (a) that was used to simulate different types of conventional CT artifacts. (b-e) plots outcomes from sFRC, as a red-bounded box, on the CT images with artifacts. (b) Missing wedge is due to a limited number of projection angles ($[30^{\circ},150^{\circ})$ at a $2^{\circ}$ stepsize). (c) Distortion is due to a mismatch in the forward projection and backprojection phases by $10^{\circ}$. (d) Blurring artifact is due to interpolation-based pixel upsampling and downsampling by a factor of $4$. (e) Noise is due to CT physics-based noise insertion for an acquisition corresponding to a $5\%$ dose of the reference CT image in (a). The display window for all the CT images is (W:$780$ L:$260$).}}
\label{img:sfrc_on_nfk_artifacts}
\end{figure*}

We used CT images from the Low-Dose Grand Challenge (LDGC) dataset \cite{ldct_data_2016} and the Michigan Image Reconstruction Toolbox\footnote{\url {https://web.eecs.umich.edu/~fessler/code/}} to simulate the conventional artifacts. A $2$D axial CT slice from patient L506 acquired at the full dose using a sharp kernel and at 3mm slice thickness was randomly selected to set the reference image to simulate different artifacts. The CT image was further pre-processed using a total variation-based denoising algorithm with its regularization parameter set at $0.006$ to gently remove noise in the reference image. The sole purpose of pre-processing the full-dose CT image was to minimize the effect of noise when analyzing the application of sFRC on other types of artifacts. The subsequent reference image is depicted in fig.\,\ref{img:sfrc_on_nfk_artifacts}(a). Our CT forward model comprises an arc-based fan beam geometry. Its acquisition parameters comprised of $100$ kVp peak voltage, $1085.6$ mm source-to-detector, $595$ mm source to the center of rotation, $986$ equispaced sensors, $250$ mm field of view, and $512\times512$ reconstruction matrix. Simulation details on each of the abovementioned conventional artifacts are provided below, and their corresponding codes can be found in our paper's GitHub repository\footnote{\url{https://github.com/DIDSR/sfrc}}.

\subsubsection{Missing Wedge}
Missing wedge-based measurements were simulated by changing the range of forward projection angles and performing reconstruction using the Filtered Backprojection (FBP) with a Ram-Lak filter \cite{ramlak_paper} as the filtering kennel. In particular, the missing wedge was progressively decreased from $[30^{\circ},150^{\circ})$ at a $2^{\circ}$ stepsize to $[0^{\circ},180^{\circ})$ at a $1^{\circ}$ stepsize to $[0^{\circ},360^{\circ})$ at a $0.5^{\circ}$ stepsize. Figs.\,\ref{img:three_mw_setup_n_frc}(b-d) illustrate reconstructions corresponding to these three missing wedge setups.

\subsubsection{Distortion}
A geometric distortion-based CT reconstruction was acquired by imposing a $10^{\circ}$ mismatch between the forward projection and backprojection parts. Specifically, sinogram measurements were acquired for angles $[0^{\circ},360^{\circ})$ at a $0.5^{\circ}$ stepsize using a reference image. Then FBP was performed employing angles $[0^{\circ},350^{\circ})$ at a $0.4861^{\circ}$ stepsize. See fig.\,\ref{img:sfrc_on_nfk_artifacts}(c) for an illustration of an output corresponding to the distortion artifact.

\subsubsection{Blur}
A blurring-based CT image for the sFRC analysis was acquired by first downsampling the reference image by a factor $4$ using OpenCV's\footnote{\url {https://docs.opencv.org/3.4/index.html}} resizing function with \texttt{INTER\_AREA} set as the pixel resampling mode. Next, we used the same function to upsample by factor $4$ with \texttt{INTER\_NEAREST} set as the resampling mode. See fig.\,\ref{img:sfrc_on_nfk_artifacts}(d) for an output from this blur artifact-based simulation. 

\subsubsection{Noise}
The reference CT image was forward projected employing angles $[0^{\circ},360^{\circ})$ at a $0.5^{\circ}$ stepsize with maximum incidence flux set as $1.35\times10^{5}$. Then, Poisson and electronic noise were inserted into the sinogram as per the CT noise model described in \cite{yu_ct_noise_model}. The dose level in the noisy realization was set to be $5\%$ of that in the reference CT image. The reconstruction was performed using the FBP algorithm. Fig.\,\ref{img:sfrc_on_nfk_artifacts}(e) depicts the noise artifact-based CT image from this simulation. The reconstructed noisy image was further evaluated using three uniform ROIs to confirm that it observed the inverse square relationship between noise and dose \cite{sarah_ct_noise_paper}.

\subsection{Results} 
Conventional imaging artifacts (like distortion, streaking, and shading) usually span across all frequency zones – from very low to very high (as depicted in fig.\,\ref{img:full_img_banded_plots} below). As such, a direct application of sFRC on images plagued with conventional artifacts (non-hallucinations) will result in most of the patches in the image being labeled as hallucinations (as depicted in fig.\,\ref{img:sfrc_on_nfk_artifacts} below).  All these artifacts are drastically different from each other. Still, a naïve application of sFRC with its hallucination threshold set as $\frac{1}{3}\cdot r \approx 0.33\text{ mm}^{-1}$  (i.e., the same value used for the CT super-resolution problem in the main paper) and without re-tuning its patch size yields most patches being labeled as hallucinations (with the red-bounding box) for the images plagued with artifacts. 

\begin{figure*}
\centering
\includegraphics[width=0.8\textwidth]{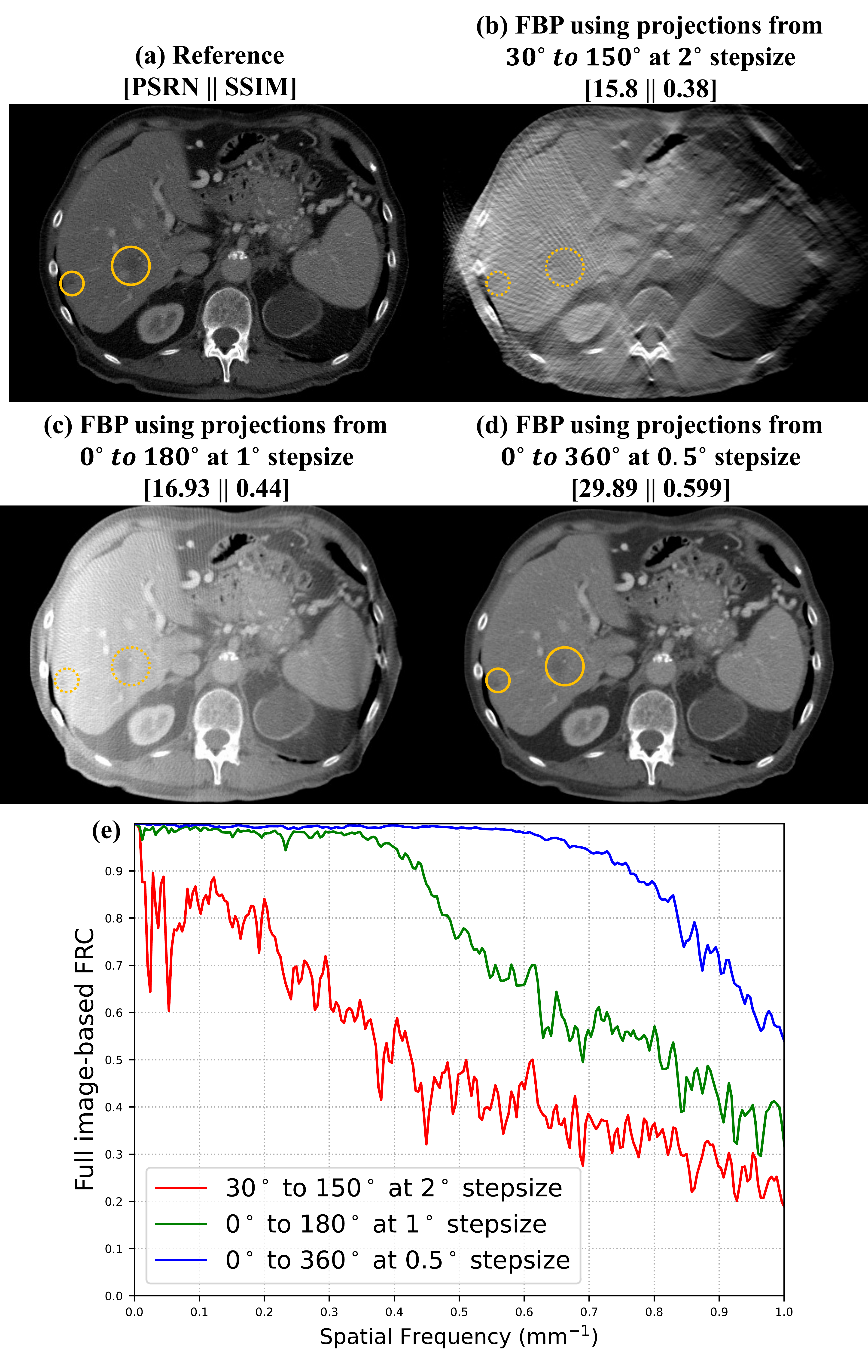}
\caption{\tip{An illustration of linear system–based improvements achieved by progressively decreasing the missing-wedge gap. These linear system–based improvements also translate into a human observer’s increased ability to discern dark lesions---shown using yellow circular regions in plots (b–d)---as well as improvements in full-image PSNR and SSIM values in plots (b–d), and Fourier Ring Correlation (FRC) curves in plot (e). The display window for all the CT images in (a-d) is (W:$780$ L:$260$).}}
\label{img:three_mw_setup_n_frc}
\end{figure*}

Hence, for visibly perceptible imaging errors, one can employ whole image-based FRC \cite{frcref_4rm_rev1}, HFEN \cite{hefn_paper}, or even a visual perception-based metric such as SSIM to discriminate data fidelity and consistency between a good reconstruction and an inaccurate reconstruction rather than using sFRC. \tip{We suggest that our previous statement be interpreted within the context of linear system-based restoration techniques, as visually illustrated in fig.\,\ref{img:three_mw_setup_n_frc} below}. The figure depicts that as we sequentially correct the missing wedge artifacts at the imaging acquisition level or the hardware level (e.g., increasing the projection angles from $[30^{\circ},150^{\circ})$ at a $2^{\circ}$ stepsize to $[0^{\circ},180^{\circ})$ at a $1^{\circ}$ stepsize to $[0^{\circ},360^{\circ})$ at a $0.5^{\circ}$ stepsize), our ability to discern clinically meaningful features (such as lesions highlighted in the figure) accordingly improves. Fig.\,\ref{img:three_mw_setup_n_frc}(e), the full image-based FRC plot, also reflects this linear system-based improvement. However, when software-based image processing techniques (either regularization- or AI-based) are used to correct any hardware-based artifacts (or loss of information due to under-sampling or use of less dose), the improvements in the full image-based FRC plots do not necessarily translate as effectiveness in the restoration of clinically important features. This phenomenon is illustrated below in fig.\,\ref{img:frc_on_blurring_n_fakes}. The use of SRGAN to upsample the low-resolution CT image not only causes a gain in the full image-based FRC but also increases hallucinations in the restored image. The second column in fig.\,\ref{img:frc_on_blurring_n_fakes} depicts hallucinations corresponding to the bowel region (highlighted using the yellow box) and the addition of plague-like features (highlighted using the green box). This outcome underscores the inaccuracy of the full-image-based comparison/analysis to efficiently capture imaging errors in local regions due to high correlation along uniform regions and high density-based features from the other parts of the full image.
\blockcomment{
In the main paper, we considered the SGRAN super-resolution algorithm as the AI method to resolve the blurring artifact for the CT downsampling by a factor of $4$. We also considered a U-Net-based approach to resolve aliasing artifacts for the MRI undersampling with an acceleration factor set as $3$. We thoroughly chronicled hallucinations identified by sFRC in solutions restored by both these AI methods. Both the ill-condition problems - in the main paper - comprised losing up to $75\%$ of information (or mismatch spanning across lower-mid to high-frequency bands, i.e., columns $4$ through $6$). If one were to consider more ill-conditioned problems – such as distortions, ultra-low-dose acquisition, and missing wedge (with severe shading, line, blurring-artifacts) as depicted in fig.\,\ref{img:full_img_banded_plots}– and employ an AI to correct such artifacts that span across low- to high-frequency bands (i.e., columns $2$ through $6$), sFRC is expected to be even more effective in detecting hallucinations in such cases. AI will create hallucinations along all the frequency bands because of the data processing inequality (as illustrated in fig\,\ref{img:three_mw_setup_n_frc} and \ref{img:frc_on_blurring_n_fakes}). Then, sFRC can easily be employed with its hallucination threshold set along very low- to mid-frequency bands (i.e., $>0\cdot r \text{ to } 0.5\cdot r$) to detect hallucinations. }

\section{\tip{Super-Resolution Wasserstein GAN (SR-WGAN) model}}\label{sec:SRWGAN}
\tip{Similar to the SRGAN model, smooth-kernel–based normal-resolution CT scans from six patients provided in the Low-Dose Grand Challenge (LDCT) \cite{ldct_data_2016} were used as target images for training the proposed SR-WGAN model. \tobereviewd{Eq.\,(3)}  in the main paper was used to generate the low-resolution input images, $\mathbf{X}_{\text{L}}$, to train SR-WGAN. The weights corresponding to the SR-WGAN's discriminator, $D_{\Theta_D}$, and generator, $G_{\Theta_G}$, networks were optimized by minimizing the following equations:

\begin{equation}
\begin{split}
    \Theta_D \leftarrow \arg\min_{\Theta_D} 
        \Biggl[ & \frac{1}{m} \sum_{i=1}^{m} D_{\Theta_{D}}(G_{\Theta_G}(X_{L}^{(i)})) \\
        &- \frac{1}{m} \sum_{i=1}^{m} D_{\Theta_D}(X_{N}^{(i)}) \\
        &+ \lambda \sum_{i=1}^{m} \left( \left\| \nabla_{\hat{X}^{(i)}} D(\hat{X}^{(i)}) \right\|_2 - 1 \right)^2 \Biggr],
\end{split}\label{eq:d_opt_wgan}
\end{equation}

where, $\hat{X}^{(i)} =\alpha_i X_{N}^{(i)} +(1-\alpha_i)G_{\Theta_G}(X_{L}^{(i)})\,\quad \alpha_i \sim \mathcal{U}(0,1);$

\begin{equation}
\begin{split}
    \Theta_G \leftarrow \arg\min_{\Theta_G} \Biggl[ & \frac{\eta}{m} \sum_{i=1}^{m} \left| X_{N}^{(i)} - G_{\Theta_G}(X_{L}^{(i)}) \right| \\
    &- \frac{1}{m} \sum_{i=1}^{m} D_{\Theta_D}(G_{\Theta_G}(X_{L}^{(i)})) \Biggr]
\end{split} \label{eq:g_opt_wgan}
\end{equation}

We adopted a Residual-in-Residual Dense Block (RRDB)-based network, as detailed in \cite{esrgan_paper}, with four RRDB blocks to construct the generator network of our SR-WGAN model. Similarly, the discriminator network was designed to be consistent with that described in \cite{srgan_paper}, except that the batch normalization operator was replaced with per-image normalization (i.e., \texttt{InstanceNorm2d}) to ensure proper functioning of the gradient penalty term (i.e., the third term inside the arg min numerical operation in the discriminator loss in eq.\,\ref{eq:d_opt_wgan} above) \cite{wgan_paper}. 

Ten percent of the six-patient CT dataset was used for systematic tuning of SR-WGAN across various data-related, optimization-related, and hyperparameter settings. This process resulted in patch-based training using $128 \times 128$ patches, which were normalized to the $[0,1]$ range and augmented using scaling-, rotation-, and flipping-based transformations. The tuning process also resulted in learning rate---for minimizing both the generator and discriminator losses--- being set to $10^{-4}$, the mini-batch size to $64$ and $\eta$ in eq.\,\ref{eq:g_opt_wgan} to $10^{3}$. $\lambda$ in eq.\,\ref{eq:d_opt_wgan} was set to be $10$ as typically set in WGAN implentations \cite{wgan_paper}.
}

\section{\tip{SRGAN data sufficiency testing}}\label{sec:srgan_data}
\begin{figure*}
\centering
\includegraphics[width=1.0\linewidth]{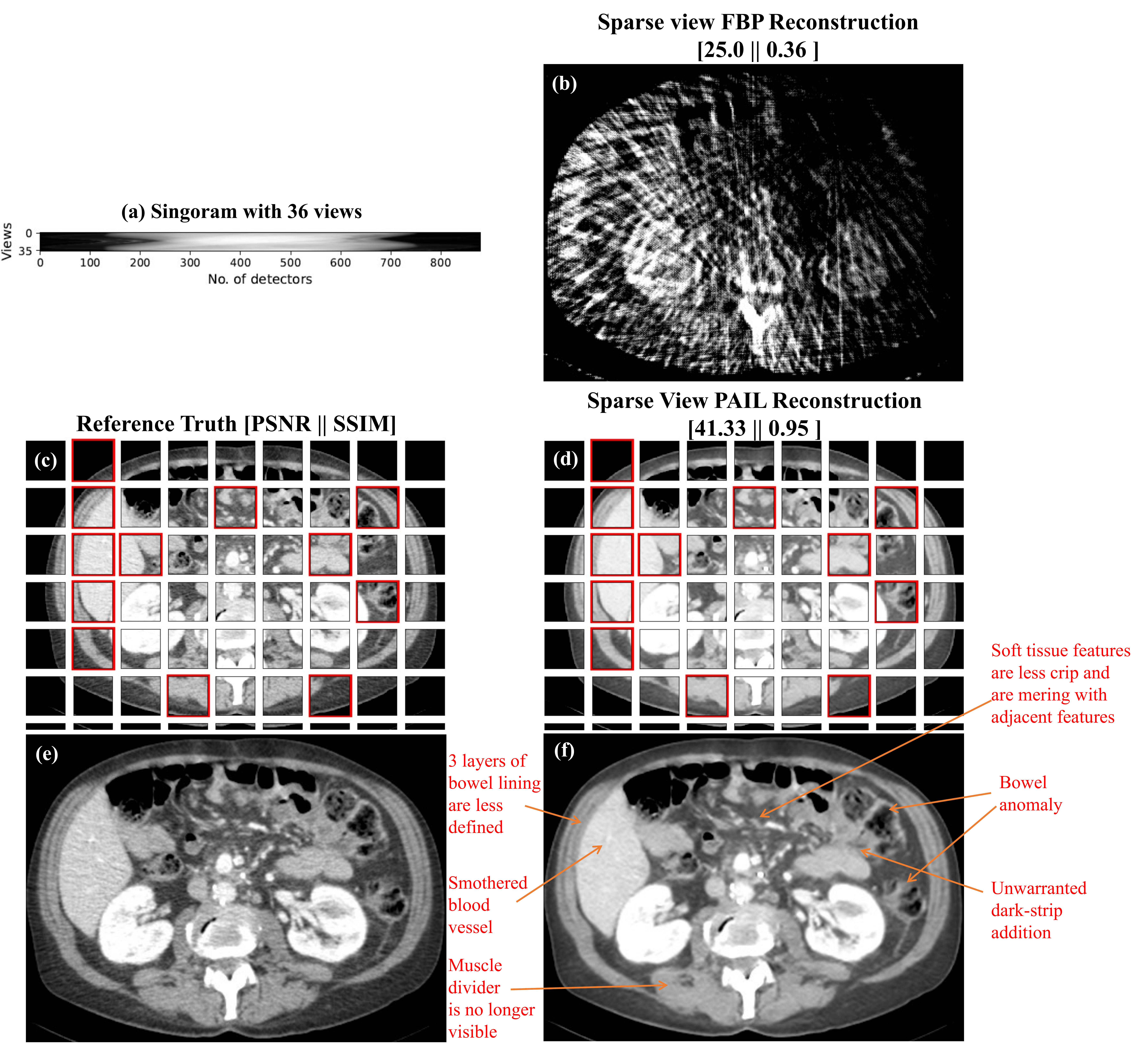}
\caption{\tip{Sparse view sinogram data is shown in (a). The corresponding FBP reconstruction is shown in (b), and the PAIL reconstruction is shown in (f). The fully sampled FBP reconstruction is provided in (e). sFRC is performed using the fully sampled FBP and PAIL reconstructions in (e) and (f), with $x_{h_{t}}=0.5$ and a patch size of $48\times 48$. The subsequent hallucinations, highlighted by red bounding boxes in (d), were confirmed and labeled in coordination with a medical officer and are shown in (f). Display window in (b-f) is (W:$400$ L:$50$).}}
\label{img:sfrc_on_pail}
\end{figure*}

\tip{
In support of the information provided in subsection \tobereviewd{IV-B2} of the main paper, we include testing results from SRGAN models trained on eight patients and six patients. The corresponding PSNR and SSIM values on the test set for the two differently trained SRGAN models are provided in  table \ref{tab:SRGAN_six_eight}. The point estimates, as well as the standard deviations, on both the smooth and sharp test sets for the eight-patient–trained SRGAN model are very similar to those of the six-patient–trained model.
}
\begin{table}[!hbt]
    \tip{
    \begin{center}
    \caption{SSIM and PSNR of SRGAN trained using six and eight patient data and tested using $188$ CT images}
    \label{tab:SRGAN_six_eight}
    \begin{tabular}{l|c|c|c}
	\hline
    No. of patients used           & Testing  & PSNR       & SSIM  \\
    to train (training kernel)     & kernel   &(std)       & (std) \\
	\hline
	6 patients (smooth)           & smooth   & $36.74\ (2.73)$ &  $0.945\ (0.010)$ \\
	8 patients (smooth)           & smooth   & $37.14\ (2.67)$ &  $0.943\ (0.001)$ \\
    \hline
    6 patients (smooth)           & sharp   & $33.72\ (1.95)$   &  $0.860\ (0.020)$ \\
	8 patients (smooth)           & sharp   & $33.94\ (1.91)$   &  $0.859\ (0.019)$ \\
	\end{tabular}
	\end{center}
    }
\end{table}   
\tip{
\section{sFRC on CT sparse view problem}
\label{sec:sfrc_on_ct_sparse}
\subsection{PAIL-based sparse view reconstruction} Recently, Zhang et al.~\cite{pail_paper} proposed an unrolling-based approach called Progressive Artifact Image Learning (PAIL) for limited-data CT reconstruction. They demonstrated that PAIL yields more accurate reconstructions than iterative-based conventional reconstruction (e.g., SIRT \cite{sirt_paper}, TV \cite{tv_4r_ct_sparse}), deep-learning post-processing approaches (e.g., FBPConvNet \cite{dense_net}, DDNet \cite{dense_net}), and other recent residual learning–based methods (e.g., IRON \cite{iron_paper}). Zhang et al. showed the efficacy of their method in terms of higher PSNR and SSIM, as well as through visual assessments using difference plots and the faithful reconstruction of finer details within regions of interest (ROIs) compared with other methods. The authors also thoroughly demonstrated the stability of their reconstruction approach.

PAIL's superior reconstruction performance is attributed to its integration of a residual domain model (RDM), an image domain model (IDM), and a wavelet domain model (WDM). Specifically, the RDM estimates residual images to reduce observable artifacts within PAIL’s iterative updates. The IDM further refines the RDM updates by suppressing unobservable artifacts and recovering fine details. Finally, the method's incorperation of a compressed sensing prior via the WDM promotes stable reconstruction.

Mathematically, PAIL solves the following joint optimization problem:
\begin{equation}
\begin{aligned}
\arg\min_{x,\,z}\;\Biggl[
&\underbrace{\tfrac{1}{2}\|\phi_{1}(z)\|_{2}^{2}}_{\text{Residual term}}
+
\underbrace{\tfrac{\alpha}{2}\|y_{0}-Ax-z\|_{2}^{2}}_{\text{Data consistency}} \\
&+
\underbrace{\beta\|H\,\phi_{2}(x)\|_{1}}_{\text{Wavelet prior \& Image domain term}} \Biggr],
\end{aligned}
\end{equation}
where $y_{0}$ denotes the measured limited-data CT projections, $A$ is the CT system matrix, $x$ is the CT image to be estimated, and $z$ is a residual variable that models the projection-domain mismatch between the measured data $y_{0}$ and the reconstructed estimate at each iteration. The operator $\phi_{1}(\cdot)$ represents the RDM following a filtered backprojection (FBP) operator, $\phi_{2}(\cdot)$ denotes the IDM, $H$ is the WDM, and $\alpha$ and $\beta$ are trainable weights that balance the three terms.

\subsection{sFRC analysis on PAIL results}
We performed our sFRC analysis using the publicly available code\footnote{\url {https://github.com/seuzjj/PAIL}} and test data provided by Zhang et al. The test data appear to correspond to the LDGC dataset, with sparse-view acquisition consisting of $36$ views out of $2200$ views. The corresponding sinogram, ground-truth image, and reconstructed images from FBP and PAIL are shown in fig.\,\ref{img:sfrc_on_pail}.

The hallucinated ROIs detected by sFRC at an $x_{h_{t}}$ of $0.5$ and a patch size of $48 \times 48$ are shown in fig.\,\ref{img:sfrc_on_pail}(d). The corresponding features inaccurately recovered by PAIL are annotated in fig.\,\ref{img:sfrc_on_pail}(e), with a detailed description provided in the main paper's section \tobereviewd{V-C}.

\blockcomment{
We first reapplied our previously used $x_{h_{t}}$ threshold of $0.33$ (i.e., assessing whether approximately $31.68$\% of the Nyquist ($r$) was faithfully reconstructed as $0.3168\times0.5\times\frac{1}{0.48}=0.33$) using a patch size of $64 \times 64$. Under this criterion, none of the patches were labeled as hallucinated by sFRC on the PAIL result.

We then re-evaluated the PAIL reconstruction using a higher $x_{h_{t}}$ threshold of $0.5$ (corresponding to $48\%$ of the Nyquist frequency) with a smaller patch size of $48 \times 48$. Under these stricter conditions, sFRC flagged several patches (fig.\,\ref{img:sfrc_on_pail}(d)). Accordingly, in coordination with our medical officer, we confirmed several hallucinated features by meticulously comparing the sFRC-labeled red bounding boxes on the PAIL-reconstructed images with the fully sampled reference image (fig.\,\ref{img:sfrc_on_pail}(c)). \textbf{The hallucinations included bowel anomalies, smoothing of the inner bowel lining, unwarranted addition of a dark strip, loss of visibility of a muscle divider, and reduced sharpness of soft-tissue features that appeared to merge with neighboring structures (shown in (fig.\,\ref{img:sfrc_on_pail}(f)).} Overall, these hallucinated features were subtle, vague, and difficult to fully characterize unless carefully reviewed alongside the fully sampled FBP reference image (fig.\,\ref{img:sfrc_on_pail}(e)).}
}

\begin{figure*}
\centering
\includegraphics[width=0.98\textwidth]{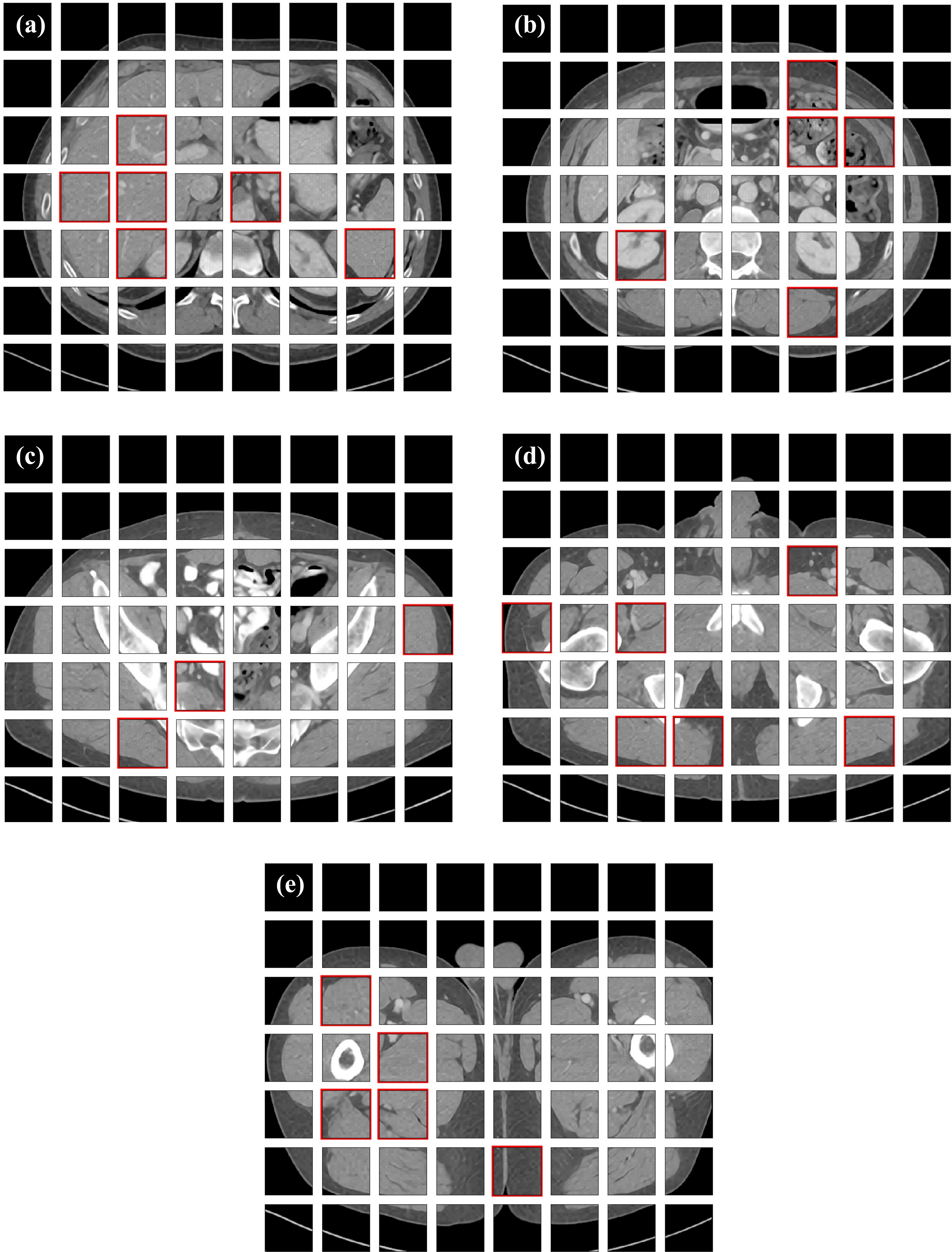}
\caption{\revfresponse{Hallucinations identified in the red bounding boxes from our sFRC analysis in the SRGAN upsampled outputs. The five SRGAN upsampled CT images correspond to images used to tune the sFRC parameters. The sFRC parameters comprised of $64\times64$ as the patch size, $0.5$ as the FRC threshold, and $\mathbf{0.34}$ as the hallucination threshold ($x_{h_{t}}$). At $0.34$ $x_{h_{t}}$, most of the hallucinated ROIs - labeled in figs.\,\ref{img:039_annotation} through \ref{img:217_annotation} - were captured by the sFRC. The display window for all images is (W:$700$ L:$50$).}}
\label{img:sfrc_on_ct_sh_tuning_ht_0.34}
\end{figure*}

\begin{figure*}
\centering
\includegraphics[width=0.98\textwidth]{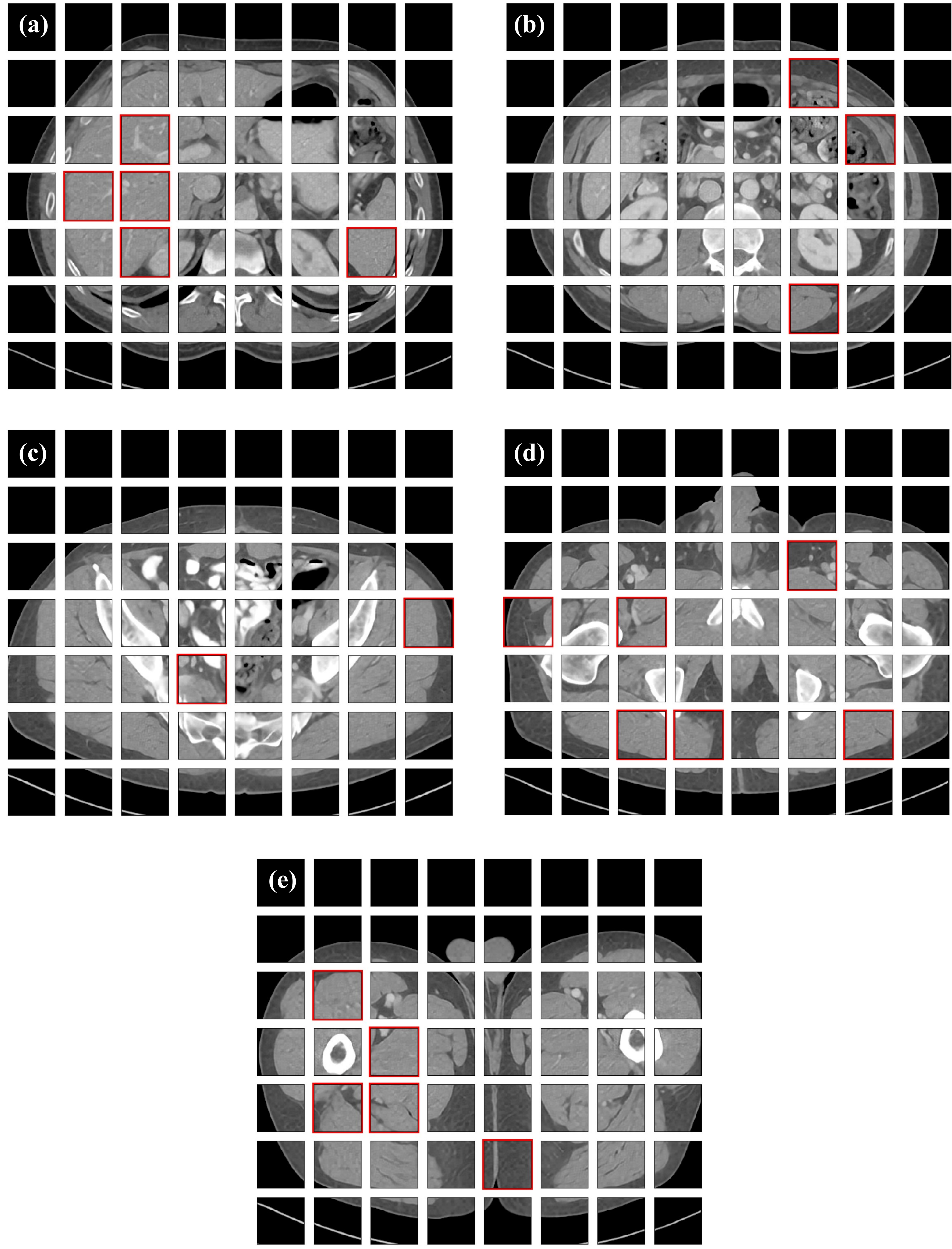}
\caption{\revfresponse{Hallucinations identified in the red bounding boxes from our sFRC analysis in the SRGAN upsampled outputs. The five SRGAN upsampled CT images correspond to images used to tune the sFRC parameters. The sFRC parameters comprised of $64\times64$ as the patch size, $0.5$ as the FRC threshold, and $\mathbf{0.33}$ as the hallucination threshold ($x_{h_{t}}$). At $0.33$ $x_{h_{t}}$, the sFRC was still able to detect all types of hallucinations (related to plaques, perturbations, and indentations) that were annotated in figs.\,\ref{img:039_annotation} through \ref{img:217_annotation}. The display window for all images is (W:$700$ L:$50$).}}
\label{img:sfrc_on_ct_sh_tuning_ht_0.33}
\end{figure*}

\begin{figure*}
\centering
\includegraphics[width=1.0\textwidth]{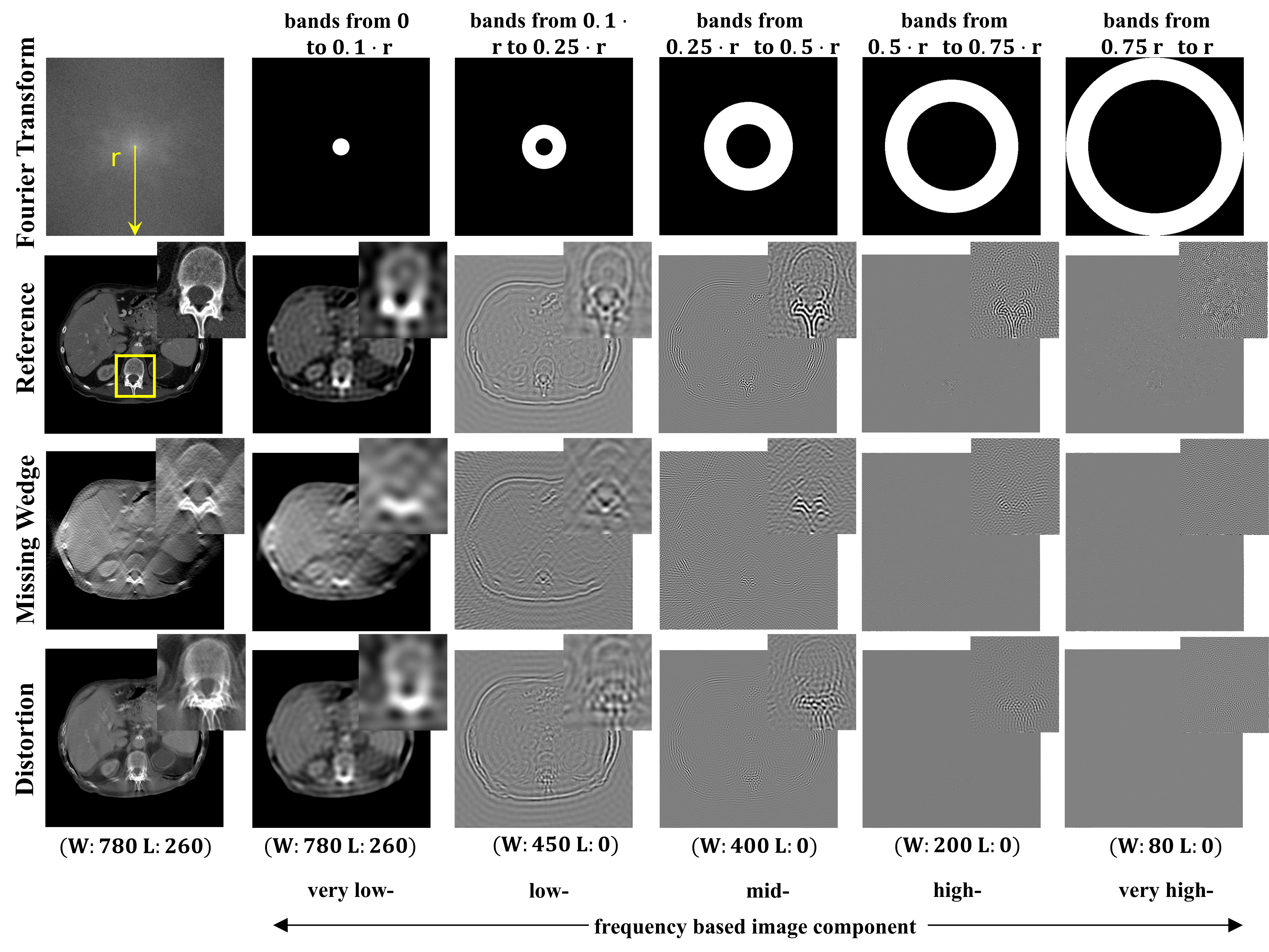}
\caption{\revfresponse{An illustration of imaging errors spanning from low- to high-frequency components of restored images plagued with Missing Wedge and Distortion artifacts. The second,
third, and fourth rows indicate plots corresponding to reference, missing wedge, and distorted images. Imaging filters corresponding to each component of an image (in the first column) are depicted in the second through sixth columns in the top row. Zoomed views of the ROI corresponding to the yellow bounding box are provided for all the image components. Display windows corresponding to each column are provided in the bottom row.}}
\label{img:full_img_banded_plots}
\end{figure*}

\begin{figure*}
\centering
\includegraphics[width=\textwidth]{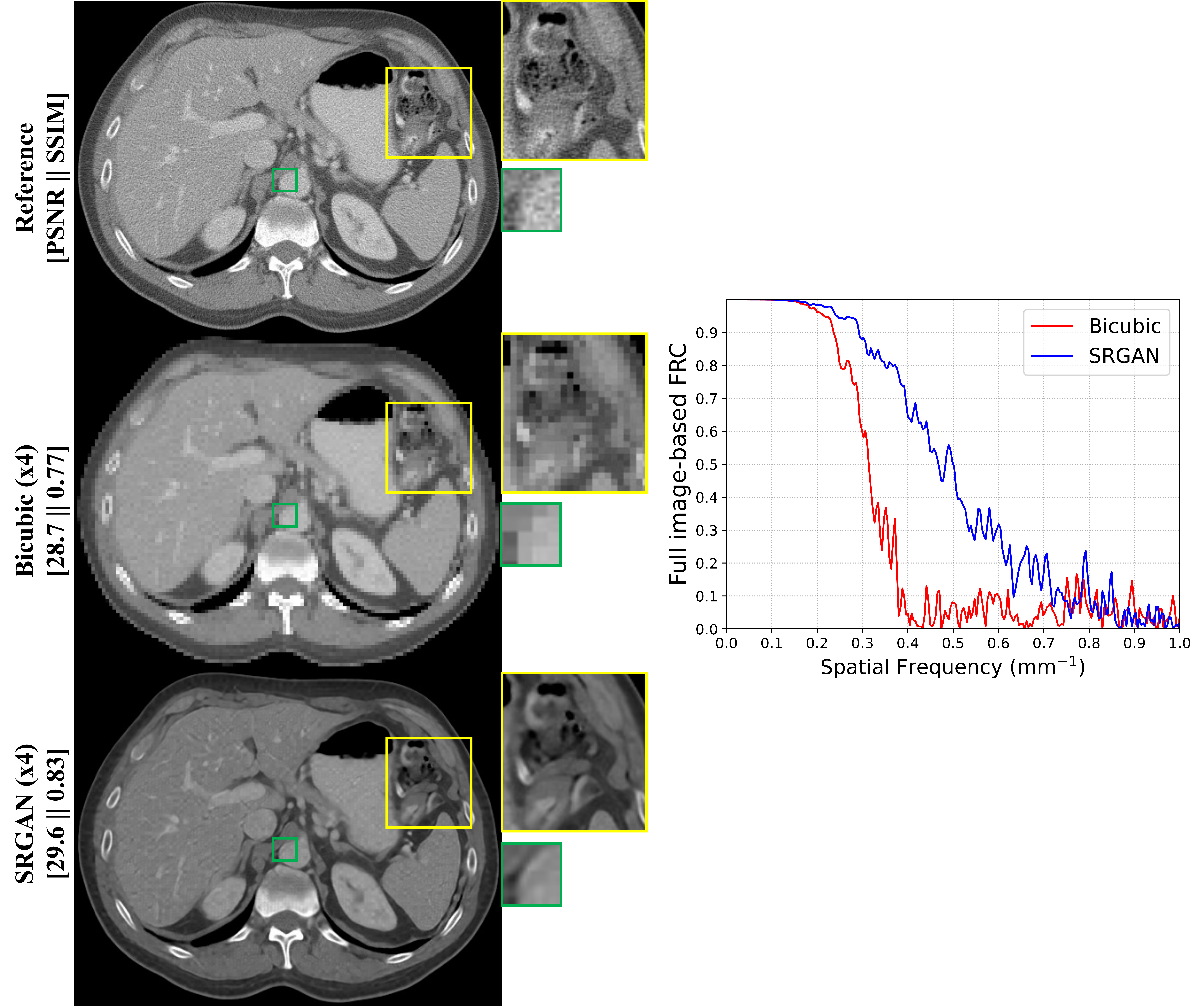}
\caption{\tip{An illustration of the lack of efficacy of the FRC, PSNR, and SSIM determined using the entire image to evaluate SRGAN, a post-processing non-linear method, accurately. The FRC curves and PSNR and SSIM values indicate apparent improvement with the use of SRGAN for a CT super-resolution problem. However, the second column reveals hallucinations in the SRGAN-based outputs, including the appearance of two bowel loops instead of a single contiguous loop within the yellow-boxed region, as well as the addition of a plaque-like structure in the green-boxed region. The display window for all the CT images in the first two columns is (W:$700$ L:$50$).}}
\label{img:frc_on_blurring_n_fakes}
\end{figure*}

\begin{figure*}
\centering
\includegraphics[width=0.8\linewidth]{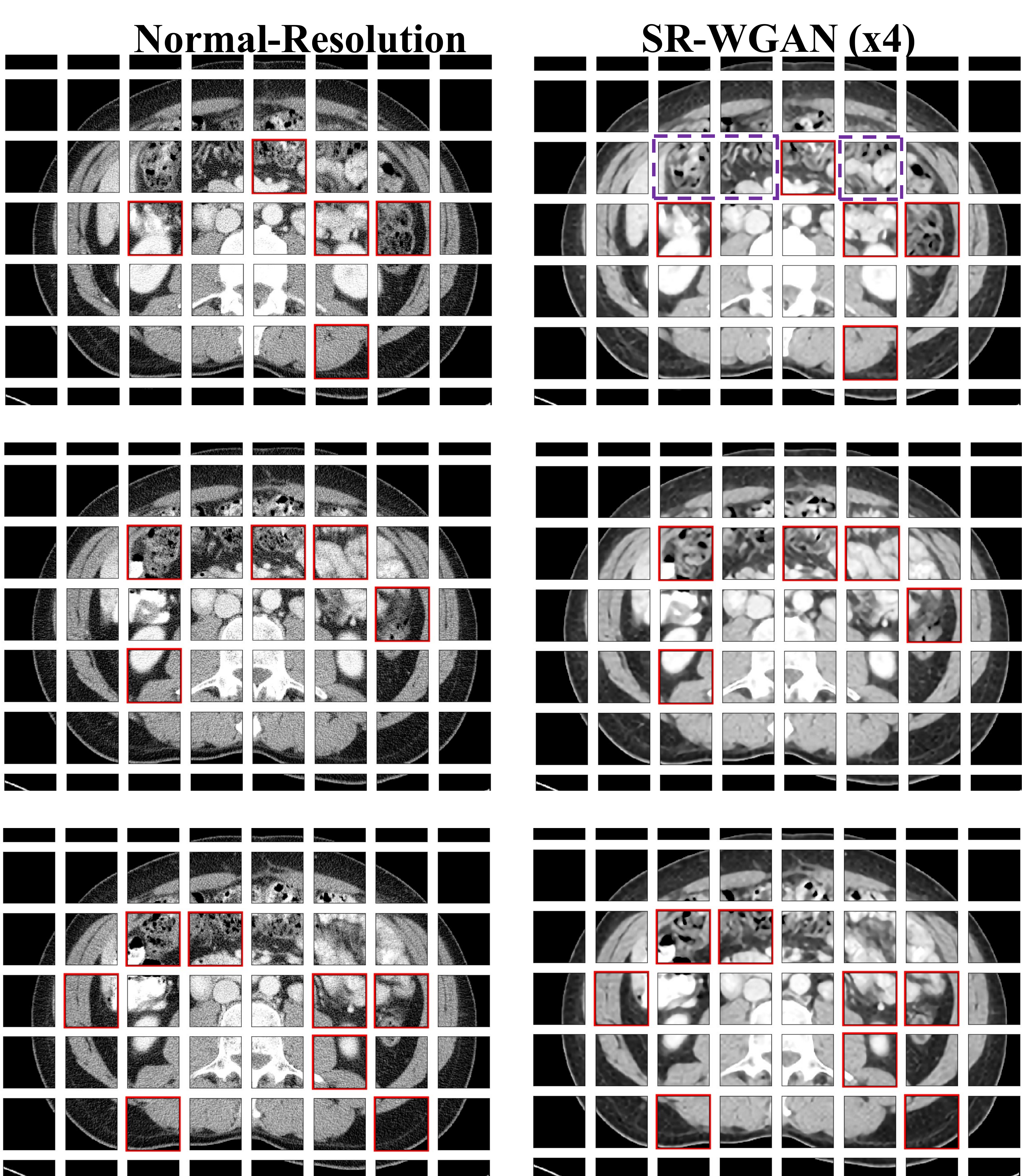}
\caption{\tip{The purple ROIs indicate hallucinated patches in the SRWGAN outputs that were initially missed by our sFRC analysis in the leftmost scan (000077.IMA). The regions corresponding to these missed hallucinations are subsequently flagged in the center (000080.IMA) and rightmost (000082.IMA) scans within the abdominal CT volume of patient L067\cite{ldct_data_2016}. This plot is similar to that obtained from the SRGAN model shown in \tobereviewd{fig.\,6} of the main paper. Display window is (W:$400$ L:$50$).}}
\label{img:wgan_ct_test_3d_robustness}
\end{figure*}

\begin{figure*}
\centering
\includegraphics[width=0.6\linewidth]{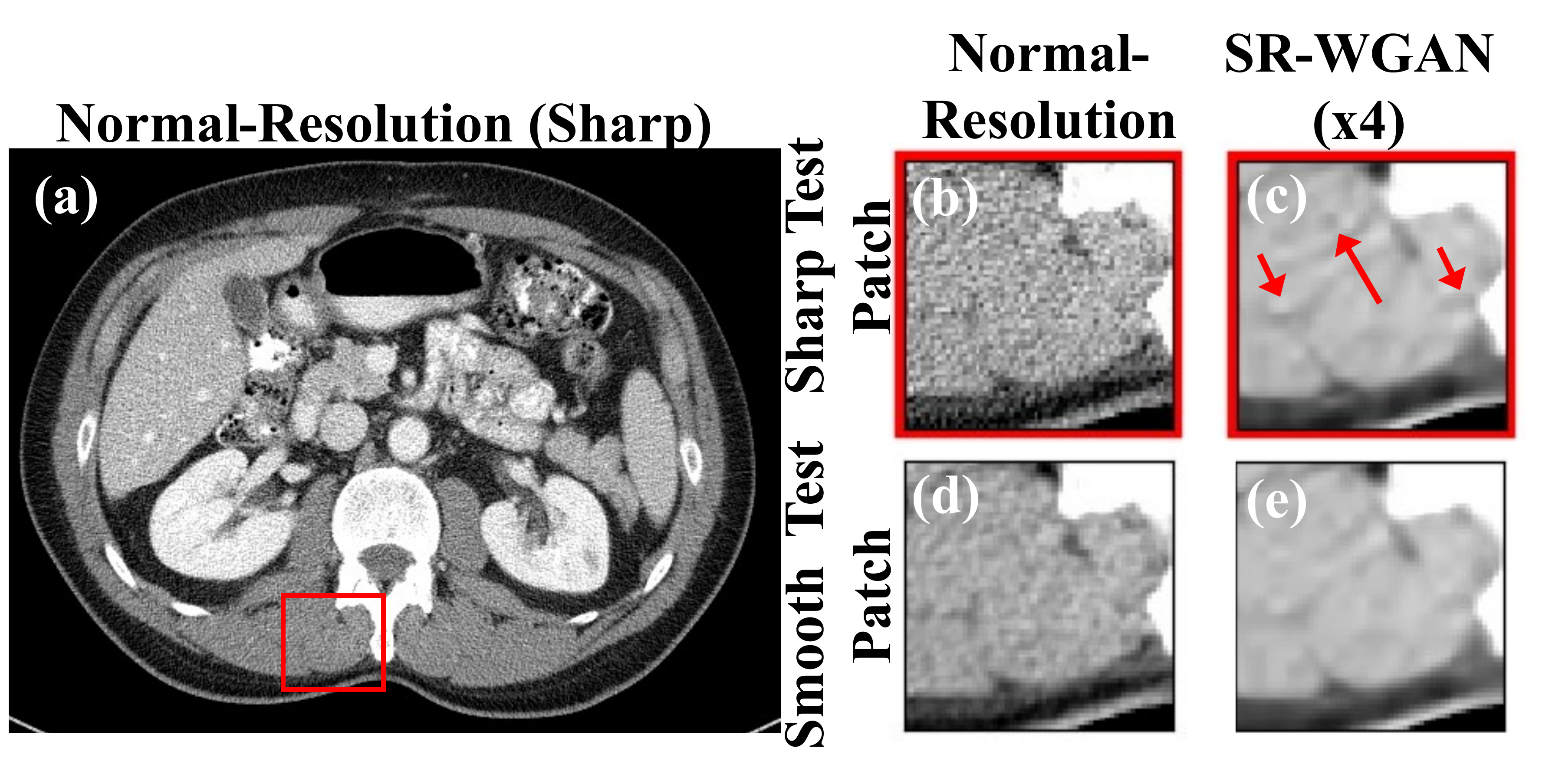}
\caption{\tip{An explicit illustration of the decay in performance of the SRWGAN super-resolution model—trained using smooth data—when analyzed using our sFRC approach against (b) sharp- and (d) smooth-kernel–based normal-resolution patches. The smoothly trained SRWGAN generates hallucinated indentations in (c), which are detected by our sFRC analysis. In contrast, the same SRWGAN upsamples (x$4$) smooth test data more faithfully in (e) and is not flagged by sFRC. The patches in (b–e) correspond to an ROI from an abdominal CT scan shown in (a). This plot is similar to that obtained from the SRGAN model shown in \tobereviewd{fig.\,7} of the main paper. Display window is (W:$400$ L:$50$).}}
\label{img:wgan_sh_sm_pic}
\end{figure*}

\begin{figure*}
\centering
\includegraphics[width=0.6\linewidth]{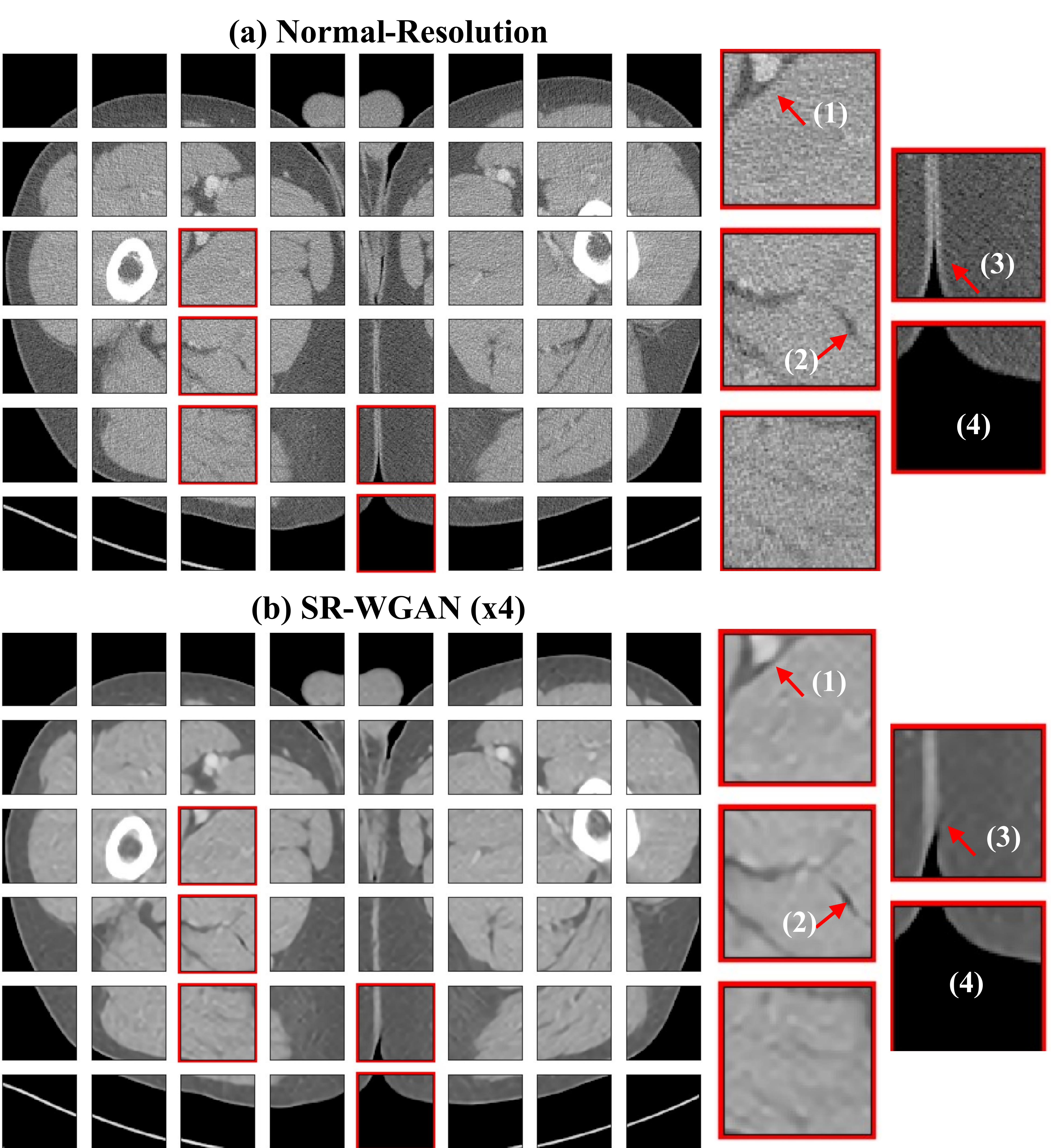}
\caption{\tip{Red bounding boxes indicate hallucinations identified by our sFRC analysis in an SRWGAN output. A plaque-like hallucination is incorrectly added along a bordering region (indicated by arrow (1)), and an ROI with fatty attenuation is underfitted as air (indicated by arrow (2)) in the SRWGAN-based output compared with its normal-resolution FBP counterpart. Radiologists are trained to identify even small amounts of air; therefore, such air-based hallucinations can mislead clinical decision-making. This plot is similar to that obtained from the SRGAN model shown in \tobereviewd{fig.\,8} of the main paper. However, compared with \tobereviewd{fig.\,8}, sFRC flags two additional ROIs in the SRWGAN results here, as shown in the third column. Red arrow (3) indicates a cut-like hallucination, while marking (4) appears to be a false positive. Display window is (W:$700$ L:$50$).}}
\label{img:srwgan_air}
\end{figure*}

\bibliographystyle{IEEEtran}
\bibliography{sreport}